%% file: main_arxiv.tex
\documentclass[11pt, a4paper, logo, copyright]{hail_lab}

\usepackage[utf8]{inputenc} 
\usepackage[T1]{fontenc}    
\usepackage{url}            
\usepackage{booktabs}       
\usepackage{amsfonts}       
\usepackage{nicefrac}       
\usepackage{microtype}      
\usepackage{xcolor}         

\input{math_commands.tex}

\usepackage{microtype}
\usepackage{graphicx}
\usepackage{subfig}
\usepackage{booktabs}
\usepackage{amsmath,bm}
\usepackage{amsthm}
\usepackage{cite}
\usepackage{natbib}
\usepackage{enumerate}
\usepackage{wrapfig}
\usepackage{color, soul}
\usepackage{psfrag}
\usepackage{adjustbox}
\usepackage{xspace}
\usepackage{amssymb}
\usepackage{multirow}
\usepackage{afterpage}
\usepackage{colortbl}
\usepackage{float}
\usepackage{textcomp}
\usepackage{siunitx}
\usepackage{mdframed}
\usepackage{makecell}
\usepackage[super]{nth}

\usepackage{enumitem}
\usepackage{bbding}
\usepackage{pifont}
\usepackage{bbm}
\usepackage{xcolor}
\usepackage{pdflscape}
\usepackage{theoremref}
\usepackage{pifont}
\usepackage{adjustbox}
\usepackage{xcolor}
\usepackage{mathtools}
\usepackage{subcaption}
\usepackage{pbox}
\usepackage{longtable}
\usepackage{listings}
\usepackage{lmodern}
\colorlet{darkgreen}{green!65!black}
\colorlet{darkblue}{blue!75!black}
\colorlet{darkred}{red!80!black}
\definecolor{statistical}{HTML}{8c564b}
\definecolor{structural}{HTML}{0070C0}
\definecolor{semantic}{HTML}{008080}
\definecolor{yellow}{HTML}{f7c600}
\definecolor{lightblue}{HTML}{0071bc}
\definecolor{lightgreen}{HTML}{39b54a}
\definecolor{deemph}{gray}{0.55}

\definecolor{baselinecolor}{gray}{.95}
\definecolor{graycolor}{gray}{.95}
\newcommand{\grayrow}{\rowcolor[gray]{.95}}

\usepackage[pagebackref=false, breaklinks=true, colorlinks, citecolor=lightblue, urlcolor=lightblue, bookmarks=false]{hyperref}

\newlength\savewidth
\newcolumntype{x}[1]{>{\centering\arraybackslash}p{#1pt}}
\newcolumntype{y}[1]{>{\raggedright\arraybackslash}p{#1pt}}
\newcolumntype{z}[1]{>{\raggedleft\arraybackslash}p{#1pt}}

\newcommand{\name}{\textsc{HeaRTS}\xspace}
\newcommand{\lightname}{\textsc{HeaRTS-Lightening}\xspace}
\newcommand{\nemotron}{Nemotron Nano 12B V2\xspace}
\newcommand{\llama}{Llama 4 Maverick\xspace}
\newcommand{\minimax}{MiniMax M2\xspace}
\newcommand{\geminiPro}{Gemini 2.5 Pro\xspace}
\newcommand{\geminiProthree}{Gemini 3.1 Pro\xspace}
\newcommand{\geminiFlash}{Gemini 2.5 Flash\xspace}
\newcommand{\gptFiveMini}{GPT 5 Mini\xspace}
\newcommand{\gptFourOneMini}{GPT 4.1 Mini\xspace}
\newcommand{\claudeHaiku}{Claude 4.5 Haiku\xspace}
\newcommand{\qwen}{Qwen3 235B\xspace}
\newcommand{\qwenCoder}{Qwen3 Coder 480B\xspace}
\newcommand{\deepseek}{DeepSeek V3.1\xspace}
\newcommand{\grok}{Grok 4.1 Fast\xspace}
\newcommand{\kimi}{Kimi K2 Thinking\xspace}
\newcommand{\glm}{GLM 4.7\xspace}
\newcommand{\glmfive}{GLM 5\xspace}

\definecolor{textgreen}{RGB}{57, 172, 57}
\definecolor{textred}{RGB}{200, 10, 10}
\definecolor{boxyellow}{HTML}{FAF5E6}
\definecolor{frameyellow}{HTML}{B7950B}
\definecolor{boxpurple}{HTML}{F4EFF6}
\definecolor{framepurple}{HTML}{6C3483}
\definecolor{boxblue}{HTML}{EEF4F8}
\definecolor{frameblue}{HTML}{2874A6}
\definecolor{boxgray}{HTML}{F0F2F3}
\definecolor{framegray}{HTML}{5D6D7E}
\definecolor{boxgreen}{HTML}{EAFaf1}
\definecolor{framegreen}{HTML}{196F3D}
\usepackage{pifont} 
\newcommand{\cmark}{\textcolor{textgreen}{\ding{51}}}
\newcommand{\xmark}{\textcolor{textred}{\ding{55}}}

\usepackage[most]{tcolorbox}
\tcbset{
    center title,
    left=0pt,
    right=0pt,
    top=0pt,
    bottom=0pt,
    colframe=black,
    colback=gray!10,
    width=\linewidth,
    enlarge left by=0mm,
    boxsep=5pt,
    boxrule=1pt,
    arc=0pt,outer arc=0pt,
}

\newtcolorbox{promptbox}[1][]{
    enhanced,
    colback=white,
    colframe=black,
    fonttitle=\bfseries,
    title=Prompt,
    attach boxed title to top left={xshift=10pt, yshift*=-\tcboxedtitleheight/2},
    boxed title style={colback=black},
    top=12pt, bottom=10pt, left=10pt, right=10pt,
    #1
}

\tcbset{
    fancybox/.style={
        enhanced,
        breakable,
        arc=1mm,
        outer arc=1mm,
        boxrule=0pt,
        leftrule=1mm,
        toptitle=0.5mm,
        bottomtitle=0.5mm,
        fonttitle=\bfseries\sffamily\small,
        fontupper=\scriptsize,
        coltitle=white,
        drop shadow
    }
}

\newtcolorbox{thoughtbox}{
    fancybox,
    colback=boxyellow,
    colframe=frameyellow,
    coltitle=black,
    title=Thought
}
\newtcolorbox{userbox}{
    fancybox,
    colback=boxpurple,
    colframe=framepurple,
    title=User
}
\newtcolorbox{agentbox}{
    fancybox,
    colback=boxblue,
    colframe=frameblue,
    title=Agent
}
\newtcolorbox{outputbox}{
    fancybox,
    colback=boxgray, 
    colframe=framegray,
    coltitle=black,
    title=Execution Output
}
\newtcolorbox{solutionbox}{
    fancybox,
    colback=boxgreen,
    colframe=framegreen,
    title=Solution
}

\definecolor{codegreen}{rgb}{0.0, 0.5, 0.0}
\definecolor{codegray}{rgb}{0.4, 0.4, 0.4}
\definecolor{codepurple}{rgb}{0.50, 0, 0.50}
\definecolor{backcolour}{rgb}{0.97, 0.97, 0.97}

\lstdefinestyle{mystyle}{
    backgroundcolor=\color{backcolour},
    commentstyle=\color{codegreen},
    keywordstyle=\color{magenta},
    stringstyle=\color{codepurple},
    basicstyle=\ttfamily\scriptsize, 
    breakatwhitespace=false,
    breaklines=true,
    captionpos=b,
    keepspaces=true,
    numbers=none,              
    showspaces=false,
    showstringspaces=false,
    showtabs=false,
    tabsize=2,
    frame=single,
    rulecolor=\color{black!10}, 
    frameround=fttt,            
    upquote=true
}
\title{
\raisebox{-0.25em}{\includegraphics[height=1.2em]{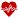}}
\name: Benchmarking LLM Reasoning \\on Health Time Series
}

\author[1$*$]{Sirui Li}
\author[1$*$]{Shuhan Xiao}
\author[1]{Mihir Joshi}
\author[2]{Ahmed Metwally}
\author[2]{Daniel McDuff}
\author[1]{Wei Wang}
\author[1$\dagger$]{Yuzhe Yang}

\affil[1]{University of California, Los Angeles}
\affil[2]{Google Research}

\correspondingauthor{$^*$Equal contribution. $^\dagger$Correspondence to: yuzhey@ucla.edu.}

\datasetlink{https://huggingface.co/datasets/yang-ai-lab/HEARTS}
\codelink{https://github.com/yang-ai-lab/HEARTS}
\projecturl{https://yang-ai-lab.github.io/HEARTS}

\input{chapters_arxiv/0_abstract_arxiv}

\begin{document}

\maketitle

\newenvironment{Itemize}{
    \begin{itemize}[leftmargin=*]
    \setlength{\itemsep}{0pt}
    \setlength{\topsep}{0pt}
    \setlength{\partopsep}{0pt}
    \setlength{\parskip}{1pt}}
{\end{itemize}}
\setlength{\leftmargini}{9pt}

\input{chapters_arxiv/1_intro_arxiv}
\input{chapters_arxiv/2_related_arxiv}
\input{chapters_arxiv/3_hearts_arxiv}
\input{chapters_arxiv/4_exp_and_anal_arxiv}
\input{chapters_arxiv/5_discussion_arxiv}

\section*{Acknowledgments}
We gratefully acknowledge the support by Amazon, UCLA DataX, Anthropic, OpenAI, and Google Cloud.
Any opinions, findings, conclusions, or recommendations expressed in this material are those of the author(s) and do not necessarily reflect the views of the funders.

\bibliography{ref}
\bibliographystyle{plain}

\newpage
\appendix
\input{chapters_arxiv/6_appendix_arxiv}

\end{document}

%% file: math_commands.tex
\usepackage{amsmath,amsfonts,bm}

\def\eqref#1{equation~\ref{#1}}

\def\1{\bm{1}}

\DeclareMathAlphabet{\mathsfit}{\encodingdefault}{\sfdefault}{m}{sl}
\SetMathAlphabet{\mathsfit}{bold}{\encodingdefault}{\sfdefault}{bx}{n}



%% file: chapters_arxiv/0_abstract_arxiv.tex
\begin{abstract}
The rise of large language models (LLMs) has shifted time series analysis from narrow analytics to general-purpose reasoning. Yet, existing benchmarks cover a limited set of \textit{health time series} modalities and tasks, failing to capture the diverse domains and temporal dependencies of real-world physiological modeling. To bridge this gap, we introduce \textbf{\name} (\textbf{Hea}lth \textbf{R}easoning over \textbf{T}ime \textbf{S}eries), a unified benchmark for evaluating hierarchical LLM reasoning over general health time series. \name integrates 16 datasets across 12 domains and 20 signal modalities, defining a taxonomy of 110 tasks grouped into four capabilities: \textit{Perception}, \textit{Inference}, \textit{Generation}, and \textit{Deduction}.
Evaluating 16 state-of-the-art LLMs on over 20K test samples reveals intriguing findings. 
First, LLMs substantially underperform specialized models, with performance only weakly related to general reasoning scores. 
Moreover, LLMs often rely on simple heuristics and struggle with multi-step temporal reasoning. 
Finally, performance declines with increasing temporal complexity, with similar failure modes within model families, indicating scaling alone is insufficient. 
By making these gaps measurable, \name provides a standardized testbed and living benchmark for developing next-generation LLM agents capable of reasoning over diverse health signals.
\end{abstract}

%% file: chapters_arxiv/1_intro_arxiv.tex
\section{Introduction}
\label{sec:intro}

\begin{figure*}[t!]
\vspace{-5pt}
\centering
\includegraphics[width=0.99\textwidth]{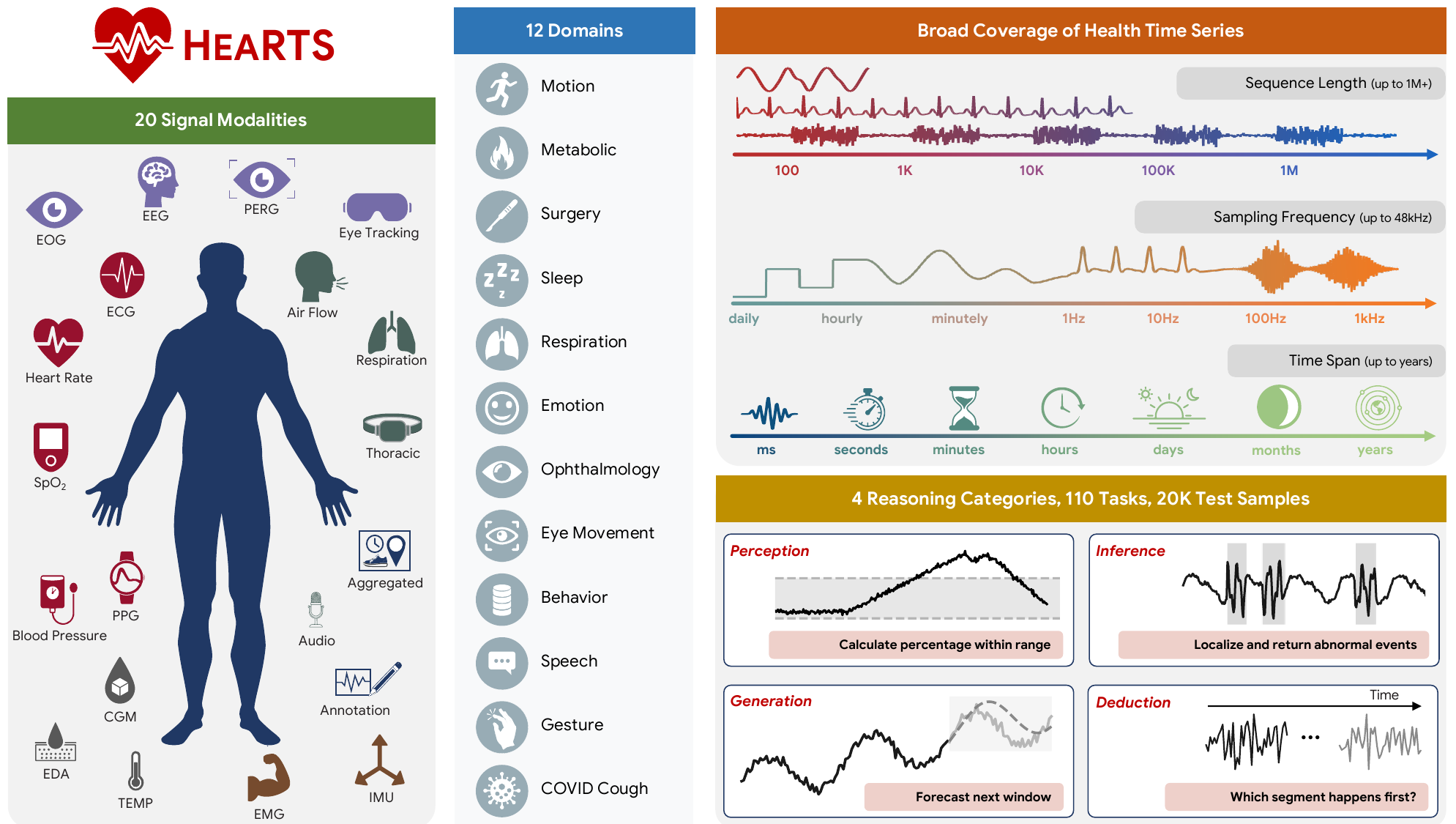}
\caption{
\small
\textbf{Overview of \name.} We present the first diverse benchmark for
health time-series reasoning, encompassing 20 signal modalities spanning 16 datasets and 12 health domains, with to date the broadest coverage of sequence length, frequency, and time span. It comprises over 20K test samples across 110 tasks, organized into four reasoning categories. More details are in Appendix~\ref{app_subsec:task_design_detail} and \ref{app_subsec:example_prompt}.
}
\label{fig:hero}
\vspace{-5pt}
\end{figure*}

Time series data serve as the backbone of decision-making in various domains, including energy \cite{alvarez2010energy}, finance \cite{lu2024trnn}, transportation \cite{guan2023spatial}, climate \cite{mudelsee2019trend}, political science \cite{beck2011modeling}, and healthcare \cite{zhang2025sensorlm, xu2026sleeplm}. Among these, healthcare is uniquely high-stakes: physiological time series encode multi-scale dynamics across time and frequency, where a brief anomaly, a slow drift, or a cross-channel interaction can change diagnosis and treatment \cite{shuai2026osf, yang2022artificial}. Consequently, the ability to \textit{reason} over these signals by linking observations to mechanisms and context is a key requirement for the next generation of healthcare artificial intelligence.

The rise of large language models (LLMs) has sparked a shift in this challenge. Because LLMs show strong reasoning over text, code, and mathematics \cite{wei2022chain-text, yang2025code, shao2024deepseekmath}, recent work has begun adapting them to time series, aiming to transfer general problem-solving skills to temporal data \cite{gruver2023largellm, jin2023timellm}. However, a critical question remains:
\vspace{-5pt}
\begin{center}
\emph{Do LLMs perform \textbf{genuine reasoning} over health time series,\\ or only exploit surface patterns and domain priors?}
\end{center}
\vspace{-5pt}
While LLMs are effective at semantic reasoning in language, their ability to carry out multi-step deduction, cross-channel correlation, and long-horizon synthesis on numerical physiological data remains unclear and is often limited in practice \cite{zhou2024can, jin2024position, feli2025llm}.

Despite the pressing need, evaluation has not kept pace with model development. 
On the one hand, existing time series reasoning benchmarks \cite{chen2025mtbench, sen2025bedtime, cai2024timeseriesexam} underrepresent health time series, failing to capture realistic temporal structures such as long-range dependencies and extreme variances in scale and frequency.
On the other hand, benchmarks that include health data either rely on synthetic signals that miss real clinical context and noise \cite{merrill2024tsandlanguage, guan2025timeomni}, or are narrowly focused on a single domain and a small set of modalities (e.g., ECG) \cite{ye2025tsaia, kong2025time-mqa, oh2023ecgqa}, limiting their ability to test diverse, human-level reasoning over heterogeneous health signals.

To bridge this gap, we present \name (\textbf{Hea}lth \textbf{R}easoning over \textbf{T}ime \textbf{S}eries), a unified living benchmark designed to evaluate hierarchical reasoning over realistic health time-series. \name integrates \textbf{16} real-world datasets across \textbf{12} health domains, covering \textbf{20} signal modalities with diverse temporal resolutions and frequency ranges (Fig.~\ref{fig:hero}). Crucially, we define a comprehensive taxonomy of \textbf{110} tasks organized into four capability families: \textit{Perception}, \textit{Inference}, \textit{Generation}, and \textit{Deduction}. This design supports controlled evaluation from lower-level signal understanding to higher-level reasoning that requires long-context integration, cross-channel consistency, and temporal directionality.

\name is designed to be easily extensible, supporting new models, datasets, and tasks as the field evolves.
It operates as a \textit{living ecosystem}, with standardized community submissions and rigorous quality checks to support continual expansion.
Using the initial release, we evaluate \textbf{16} state-of-the-art (SOTA) LLMs on more than \textbf{20K} test samples, and reveal findings that highlight key gaps and directions for future research.
Our contributions are as follows:
\begin{Itemize}
    \item We introduce \name, the first diverse benchmark for health time-series reasoning, spanning 12 health domains, 20 signal modalities, and 110 tasks, with to date the broadest coverage of sequence length, frequency, and time span.
    \vspace{-1pt}
    \item We propose a unified, hierarchical evaluation setting that goes beyond standard multiple-choice questions, with four cognitive levels that range from basic calculation to long-horizon synthesis and counterfactual deduction.
    \vspace{1pt}
    \item Experiments on 16 LLMs over more than 20K test samples show that SOTA models struggle with health time-series reasoning across tasks, and that performance is only weakly related to general reasoning scores.
    \vspace{1pt}
    \item We identify consistent scaling challenges with longer inputs, higher sampling frequencies, and longer time spans, revealing model-agnostic difficulty patterns and highlighting open directions for future research. 
\end{Itemize}

%% file: chapters_arxiv/2_related_arxiv.tex
\section{Related Work}
\label{sec:related-work}

\input{tables/table_related_work}

\textbf{Language Models for Time Series Analysis.}
Early work on adapting LLMs to time series focused on forecasting, either by bridging modality gaps through reprogramming and feature alignment \cite{jin2023timellm, li2024urbangpt} or by leveraging long-context prompting for zero-shot prediction \cite{liu2024autotimes, lu2024incontext}. Recent studies shift toward agentic and tool-augmented settings, using multi-agent designs for data understanding \cite{lee2025timecap, zhou2025merit} and combining analytical tools with hierarchical chain-of-thought for more complex reasoning workflows \cite{liu2025ts, guan2025timeomni}. Despite this progress, it remains unclear how well these methods support \emph{physiological} reasoning, including cross-signal correlation, long-horizon synthesis, and clinically meaningful inference.
In this work, we propose a dedicated benchmark for evaluating LLM reasoning over health time series.

\textbf{Time Series Reasoning Benchmarks.}
Healthcare remains underrepresented in existing time series reasoning benchmarks. Most frameworks target general-domain data, with health modalities absent or only lightly covered \cite{cai2024timeseriesexam, sen2025bedtime,chen2025mtbench}. Benchmarks that include health settings often sacrifice realism for breadth, or remain limited in domain and modality coverage. For example, recent benchmarks rely on synthetic health signals \cite{merrill2024tsandlanguage, guan2025timeomni}, which can miss real physiological semantics, noise, and clinical constraints. Conversely, benchmarks built on real-world medical data are typically confined to a few domains and a small set of modalities \cite{ye2025tsaia, kong2025time-mqa, oh2023ecgqa}.
As summarized in Table \ref{tab:benchmark_comparison}, current benchmarks therefore lack the diversity and fidelity required to assess human-level reasoning in real-world health contexts. In contrast, \name addresses these gaps by providing a unified, health-focused evaluation framework with diverse real-world datasets and a hierarchical task taxonomy that tests reasoning across disparate temporal scales and healthcare contexts.

%% file: tables/table_related_work.tex
\begin{table*}[!t]
\setlength{\tabcolsep}{2pt}
\renewcommand{\arraystretch}{1.0}
\centering
\small
\vspace{-5pt}
\caption{
\small
\textbf{Comparisons between \name and other time series reasoning benchmarks.}}
\label{tab:benchmark_comparison}
\small

\definecolor{barbg}{HTML}{E5E7EB}
\definecolor{seqcolor}{HTML}{009B77}
\definecolor{freqcolor}{HTML}{3B82F6}
\definecolor{timecolor}{HTML}{F97316}
\definecolor{rwred}{HTML}{DC2626}
\definecolor{rwgreen}{HTML}{16A34A}

\newcommand{\seqbar}[3]{
    \makebox[12mm][r]{\raisebox{0.2ex}{\tiny #3}}%
    \makebox[21mm][l]{%
        \rlap{\textcolor{barbg}{\rule{21mm}{1.5ex}}}%
        \hspace{#1mm}%
        \textcolor{seqcolor}{\rule{#2mm}{1.5ex}}%
    }%
}

\newcommand{\specbar}[3]{
    \makebox[18mm][r]{\raisebox{0.2ex}{\tiny #3}}%
    \makebox[21mm][l]{%
        \rlap{\textcolor{barbg}{\rule{21mm}{1.5ex}}}%
        \hspace{#1mm}%
        \textcolor{freqcolor}{\rule{#2mm}{1.5ex}}%
    }%
}

\newcommand{\timebar}[3]{
    \makebox[16mm][r]{\raisebox{0.2ex}{\tiny #3}}%
    \makebox[21mm][l]{%
        \rlap{\textcolor{barbg}{\rule{21mm}{1.5ex}}}%
        \hspace{#1mm}%
        \textcolor{timecolor}{\rule{#2mm}{1.5ex}}%
    }%
}

\adjustbox{max width=\textwidth}{
\begin{tabular}{lc ccc cccc @{\hspace{6pt}} cccc}
    \toprule[1.5pt]
    \multirow{2.5}{*}{\textbf{Benchmark}} & \multirow{2.5}{*}{\makecell{\textbf{Real} \\ \textbf{World?}}} & \multirow{2.5}{*}{\makecell{\hspace{10pt}\textbf{Sequence Length} \\ \hspace{10pt}\textbf{Range}}} & \multirow{2.5}{*}{\makecell{\textbf{Frequency Range}}} & \multirow{2.5}{*}{\textbf{Time Range}} & \multirow{2.5}{*}{\textbf{\# Domain}} & \multirow{2.5}{*}{\makecell{\textbf{\# TS} \\ \textbf{Modality$^\dagger$}}} & \multirow{2.5}{*}{\makecell{\textbf{\# Test} \\ \textbf{Sample{$^\dagger$}}}} & \multirow{2.5}{*}{\textbf{\# Task}} & \multicolumn{4}{c}{\textbf{Task Type}} \\
    \cmidrule(lr){10-13}
     & & & & & & & & & Per. & Inf. & Gen. & Ded. \\
    \midrule\midrule

    \textit{TimeSeriesExam}~\cite{cai2024timeseriesexam}  & \textcolor{rwred}{\xmark} & \seqbar{10}{1}{1K}             & -                              & -                              & 1  & 0  & 0       & 100 & \cmark & \cmark & \xmark & \xmark \\
    \textit{BEDTime}~\cite{sen2025bedtime}                & \textcolor{rwred}{\xmark} & \seqbar{2}{15}{36--25K}         & -                              & -                              & 2  & 0  & 0       & 12  & \cmark & \cmark & \xmark & \xmark \\
    \textit{TSR-SUITE}~\cite{guan2025timeomni}            & \textcolor{rwred}{\xmark} & \seqbar{2}{9}{48--0.8K}         & \specbar{2}{2}{daily--hourly}& \timebar{4}{6}{hours--days}   & 10 & 15 & - & 36  & \xmark & \cmark & \cmark & \xmark \\
    \textit{TS Reasoning}~\cite{merrill2024tsandlanguage} & \textcolor{rwred}{\xmark} & \seqbar{1}{12}{10--1.5K}        & \specbar{0}{4}{annual--hourly} & \timebar{4}{17}{hours--years}& 10 & 15 & 14K     & 50  & \xmark & \cmark & \cmark & \xmark \\

    \midrule

    \textit{MTBench}~\cite{chen2025mtbench}               & \textcolor{rwgreen}{\cmark} & \seqbar{8}{5}{168--1.3K}        & \specbar{4}{3}{hourly-minutely}& \timebar{10}{6}{days--months} & 2  & 0  & 0       & 12  & \cmark & \cmark & \xmark & \xmark \\
    \textit{TSAIA}~\cite{ye2025tsaia}                     & \textcolor{rwgreen}{\cmark} & \seqbar{1}{9}{10--0.9K}         & \specbar{2}{12}{daily--128Hz}  & \timebar{0}{21}{secs--years} & 4  & 1  & 0.07K   & 12  & \xmark & \cmark & \cmark & \xmark \\
    \textit{ECG-QA}~\cite{oh2023ecgqa}                    & \textcolor{rwgreen}{\cmark} & \seqbar{13}{1}{5K}             & \specbar{16}{1}{500Hz}         & \timebar{0}{2}{seconds}       & 1  & 1  & 8K      & 70  & \xmark & \cmark & \xmark & \xmark \\
    \textit{Time-MQA}~\cite{kong2025time-mqa}             & \textcolor{rwgreen}{\cmark} & \seqbar{1}{5}{8--0.2K}          & \specbar{2}{13}{daily--256Hz}& \timebar{0}{10}{secs--days}   & 12 & 3  & 56K     & 60  & \xmark & \cmark & \cmark & \xmark \\

    \midrule

    \rowcolor[gray]{.95}
    \textbf{\name (ours)} & \textbf{\textcolor{rwgreen}{\cmark}} & \seqbar{2.5}{18.5}{\textbf{60--1M}} & \specbar{2}{19}{\textbf{daily--48kHz}} & \timebar{0}{21}{\textbf{secs--years}} & \textbf{12} & \textbf{20} & 20K & \multicolumn{1}{c@{\color[gray]{.95}\vrule width 6pt}}{\textbf{110}} & \textbf{\cmark} & \textbf{\cmark} & \textbf{\cmark} & \textbf{\cmark} \\
    \bottomrule[1.5pt]
\end{tabular}
}
\vspace{5pt}
\newline
\tiny
Per.: Perception,~~~Inf.: Inference,~~~Gen.: Generation,~~~Ded.: Deduction.~~~\textcolor{seqcolor}{\rule{2mm}{1.5ex}} Sequence Length Coverage~~~\textcolor{freqcolor}{\rule{2mm}{1.5ex}} Frequency Coverage~~~\textcolor{timecolor}{\rule{2mm}{1.5ex}} Time Coverage~~~$^\dagger$Only health time series considered.\\[4pt]
\end{table*}

%% file: chapters_arxiv/3_hearts_arxiv.tex
\section{\name}
\label{sec:ours}

To address the evaluation gap, we introduce \name, the first unified benchmark for health time-series reasoning. Departing from the conventional paradigm that reduces diverse objectives to a single QA format, \name establishes a taxonomy anchored in four categories: Perception, Inference, Generation, and Deduction. This allows for evaluation in intrinsic task formats rather than relying solely on MCQs. \name integrates 16 datasets across 12 domains and 20 signal modalities, encompassing 110 tasks and 20,226 test cases to ensure robust and comprehensive evaluation.

\input{tables/table_main_table}

\subsection{Data Overview}
\label{subsec:data_overview}

\textbf{Dataset Diversity and Domain Coverage.} \name curates 16 datasets across 12 domains to facilitate rigorous benchmarking. This encompasses a broad spectrum of health contexts (complete statistics in Appendix \ref{app_subsec:data_statistics}):
\vspace{-10pt}
\begin{Itemize}
    \item \textit{\textbf{Motion:}} Human activity recognition (HAR) and intensity estimation via wearable IMU data from Capture24 \cite{chan2024capture24} and PAMAP2 \cite{bleser2015pamap2}.
    \item \textit{\textbf{Metabolic:}} Longitudinal glucose variability and dietary response tracking via Shanghai Diabetes \cite{zhao2023shanghai} and CGMacros \cite{das2025cgmacros}.
    \item \textit{\textbf{Surgery:}} Intra-operative vital sign monitoring and anesthesia management using VitalDB \cite{lee2022vitaldb}.
    \item \textit{\textbf{Sleep:}} Polysomnography (PSG) based sleep physiology analysis from SHHS \cite{quan1997shhs}.
    \item \textit{\textbf{Respiration:}} Respiratory mechanics and breathing pattern analysis utilizing Harespod \cite{zhang2024harespod}.
    \item \textit{\textbf{Emotion:}} Affective state recognition derived from multimodal biosignals in PhyMER \cite{pant2023phymer}.
    \item \textit{\textbf{Ophthalmology:}} Retinal function assessment via pattern electroretinography (PERG) signals in PERG-IOBA \cite{fernandez2024perg}.
    \item \textit{\textbf{Eye Movement:}} Oculomotor trajectory analysis using GazeBase \cite{griffith2021gazebase}.
    \item \textit{\textbf{Behavior:}} Longitudinal behavior and mental health monitoring using GLOBEM \cite{xu2023globem}.
    \item \textit{\textbf{Speech:}} Voice biomarker identification and pathological speech analysis using Bridge2AI-voice \cite{bensoussan2025bridge2ai} and VCTK \cite{yamagishi2019vctk}.
    \item \textit{\textbf{Gesture:}} Electromyography (EMG) based hand gesture classification leveraging GrabMyo \cite{pradhan2022grabmyo}.
    \item \textit{\textbf{COVID Cough:}} Respiratory symptom screening via cough audio recordings in CoughVID \cite{orlandic2021coughvid} and Coswara \cite{sharma2020coswara}.
\end{Itemize}
\vspace{-10pt}

\begin{wrapfigure}{r}{0.5\textwidth}
    \vspace{-5pt}
    \centering
    \includegraphics[width=0.9\linewidth]{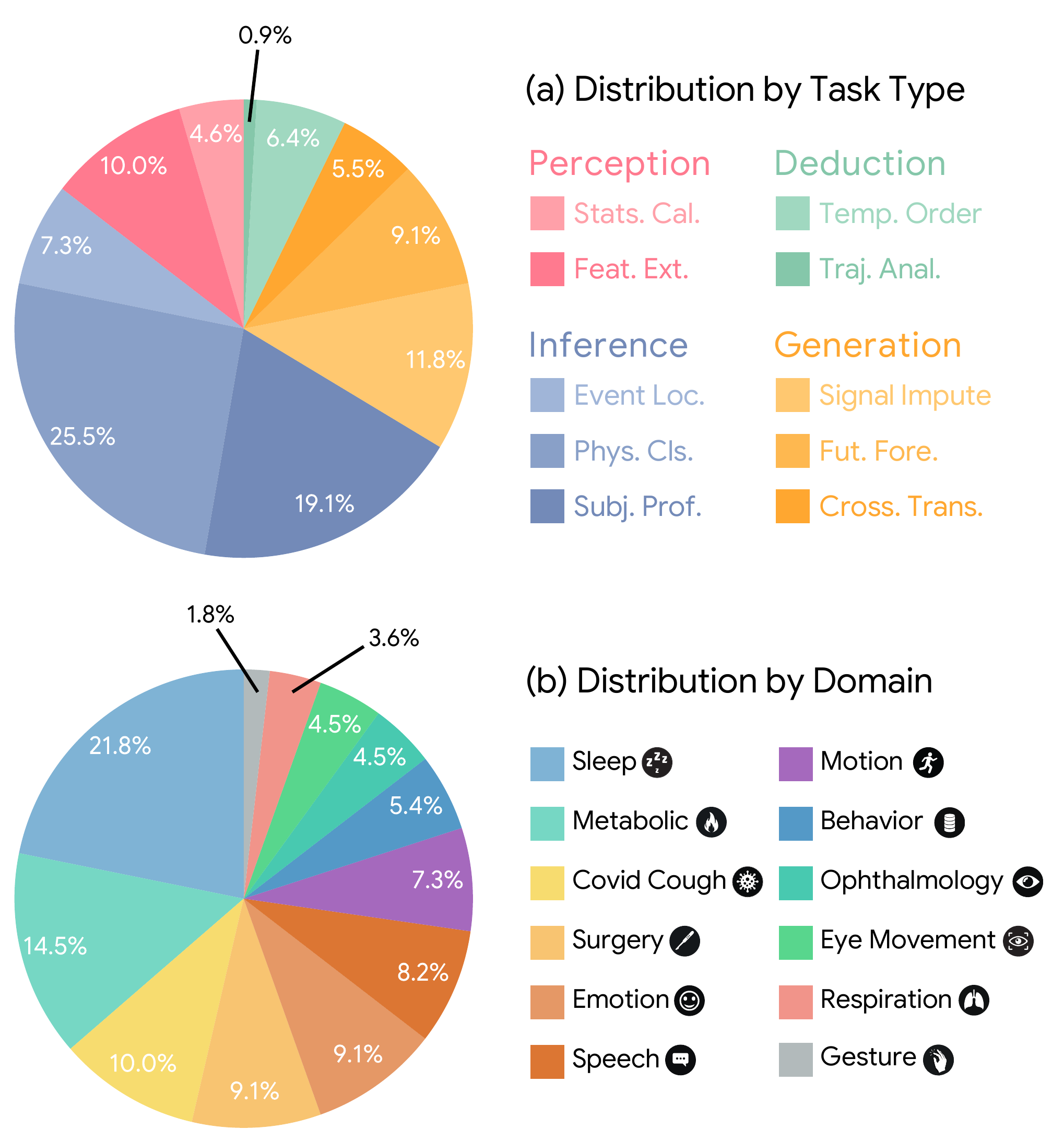}
    \caption{\small
    \textbf{Task and domain distributions in \name.}
    }
    \label{fig:task_distribution}
    \vspace{-5pt}
\end{wrapfigure}

\textbf{Multi-Modal Signal Complexity.} 
\name highlights broad modality coverage: it includes 20 distinct signal channels, spanning standard bioelectrical recordings to high-dimensional acoustic streams and daily aggregated measurements (Fig.~\ref{fig:hero}). This heterogeneity goes beyond prior limited-modality benchmarks and challenges models to generalize across signals with fundamentally different physical origins, sampling regimes, and physiological semantics.

\textbf{Temporal Scale and Resolution.} \name captures wide variation in temporal scale: \textit{input lengths} range from 60 to over 1M steps, and \textit{sampling frequencies} span from daily metrics to 48kHz. Consequently, observation windows vary from seconds to years. This setting requires models to handle both high-frequency structure and long-range dependencies across highly different resolutions.

\vspace{-2pt}
\subsection{Task Overview}
\label{subsec:task_overview}

\textbf{Design Principles and Formulation.}
Task construction in \name follows 3 principles that ensure practical relevance, label validity and well-defined reasoning structure: 
\ding{182} \textit{Evidence-Grounded Significance}: each task is anchored in real world needs by aligning its definition with authoritative medical guidelines or prior peer-reviewed studies of the same (or closely related) problem. 
\ding{183} \textit{Context-Grounded Label Validity}: labels are directly derived directly from the recorded signals and metadata, avoiding external assumptions that cannot be checked within the dataset. 
\ding{184} \textit{Data-Grounded Multi-Step Reasoning}: solving a task must require non-trivial, multi-step reasoning over the data instead of single-shot pattern matching. We formalize the reasoning process as a Markov Decision Process (MDP) trajectory $\tau = \{(s_i, a_i, r_i)\}_{i=0}^n$, where states $s_i \in S$ encode the data context, actions $a_i \in A$ denote intermediate decisions, and rewards $r_i \in R$ evaluate the validity of the reasoning path. This formulation enforces that successful solutions proceed through a sequence of data-grounded decisions rather than a single shortcut step.

\textbf{Hierarchical Reasoning Taxonomy.} Guided by these principles, we curate 110 tasks grouped into four hierarchical reasoning categories. Fig. \ref{fig:task_distribution} summarizes their distribution across task types and domains. A key feature of \name is that it moves beyond the multiple-choice paradigm, emphasizing exact numerical answers and open-ended generation. Appendix \ref{app_subsec:task_design_detail} and \ref{app_subsec:example_prompt} provide detailed design examples.

\vspace{-10pt}
\begin{itemize}[leftmargin=*]
\setlength{\itemsep}{-1pt}
    \item \textit{\textbf{Perception.}} This category focuses on identifying physiological markers that serve as prerequisites for downstream analysis. Tasks include:
    \begin{itemize}[nosep]
        \item \textit{Statistical Calculation:} Computing descriptive statistics of raw signals (e.g. the percentage of time a CGM signal remains within normal range).
        \item \textit{Feature Extraction:} Derive informative representations or biomarkers from signals (e.g. isolating spectral bandpower from EEG data).
    \end{itemize}

    \item \textit{\textbf{Inference.}} This category advances to analysis at varying granularities, ranging from precise temporal grounding to holistic subject characterization. It includes:
    \begin{itemize}[nosep]
        \item \textit{Event Localization:} Identify the timestamp or temporal boundaries of specific events within a signal.
        \item \textit{Physiological Classification:} Categorizing physiological state of a signal segment (e.g., sleep staging).
        \item \textit{Subject Profiling:} Inferring holistic, subject-level attributes or conditions (e.g., Parkinson's prediction).
    \end{itemize}

    \item \textit{\textbf{Generation.}} This category evaluates temporal understanding by requiring point-by-point synthesis of complete sequences. Three types of task are included:
    \begin{itemize}[nosep]
        \item \textit{Future Forecasting:} Predict future segments of a biosignal based on observed historical context.
        \item \textit{Signal Imputation:} Reconstructing missing or corrupted portions within a temporal sequence to restore signal integrity.
        \item \textit{Cross-modal Translation:} Synthesizing a target signal modality from a source channel based on their underlying physiological correlations.
    \end{itemize}

    \item \textit{\textbf{Deduction.}} Finally, this category evaluates arrow-of-time reasoning and the ability to derive insights from longitudinal dependencies and multi-session interactions:
    \begin{itemize}[nosep]
        \item \textit{Temporal Ordering:} Determine the relative temporal order of signals.
        \item \textit{Trajectory Analysis:} Analyze long-term health trajectories across multiple visits or sessions.
    \end{itemize}
\end{itemize}
\vspace{-10pt}

\subsection{\name as a Living Ecosystem}
\label{subsec:living_ecosys}

Unlike traditional benchmarks that become static artifacts, we position \name as a dynamic, community-driven ecosystem that can evolve with rapid progress in AI for healthcare. We provide standardized contribution protocols on our project website, allowing the broader research community to submit new \textit{datasets}, \textit{clinical tasks} (task APIs in Appendix \ref{app_subsec:task_base_code}), and new \textit{agent architecture} (agent APIs in Appendix \ref{app_subsec:agent_base_code}). 
In addition, any LLM can be evaluated in \name through a simple API endpoint interface. To maintain benchmark integrity, we use a human-in-the-loop maintenance process: a dedicated team reviews submissions for correctness and compliance with privacy requirements before merging them into the core repository. This continuous integration setup keeps the leaderboard up to date (see Fig. \ref{fig:llm_performance}), supporting a collaborative benchmark that grows in coverage and difficulty through community effort.

%% file: tables/table_main_table.tex
\renewcommand{\ul}[1]{\underline{#1}}

\begin{table*}[t]
\centering
\small
\setlength{\tabcolsep}{0pt}
\vspace{-5pt}
\caption{\textbf{Breakdown of LLM performance across each reasoning category.} The best results across all models are shown in \textbf{bold}, and the second-best results are \underline{underlined}. Overall scores are computed as the macro-average of the four category scores. SOTA ML performance is reported only on the 32-task subset, as detailed in Sec. \ref{subsec:over_performance}.}
\vspace{-5pt}
\begin{tabular*}{0.99\textwidth}{@{\extracolsep{\fill}}l cc ccc ccc cc c}
\toprule[1.5pt]
& \multicolumn{2}{c}{\textbf{Perception}} & \multicolumn{3}{c}{\textbf{Inference}} & \multicolumn{3}{c}{\textbf{Generation}} & \multicolumn{2}{c}{\textbf{Deduction}} & \\
\cmidrule(lr){2-3} \cmidrule(lr){4-6} \cmidrule(lr){7-9} \cmidrule(lr){10-11}
\textbf{Model} &
\multicolumn{1}{c}{Stat.} & \multicolumn{1}{c}{Feat.} & \multicolumn{1}{c}{Event} &
\multicolumn{1}{c}{Phys.} & \multicolumn{1}{c}{Subj.} &
\multicolumn{1}{c}{Sig.} & \multicolumn{1}{c}{Fut.} & \multicolumn{1}{c}{Cross.} &
\multicolumn{1}{c}{Temp.} & \multicolumn{1}{c}{Traj.} &
\textbf{Overall} \\
&
\multicolumn{1}{c}{Calc.} & \multicolumn{1}{c}{Ext.} & \multicolumn{1}{c}{Loc.} &
\multicolumn{1}{c}{Cls.} & \multicolumn{1}{c}{Prof.} &
\multicolumn{1}{c}{Imp.} & \multicolumn{1}{c}{Fore.} & \multicolumn{1}{c}{Trans.} &
\multicolumn{1}{c}{Ord.} & \multicolumn{1}{c}{Anal.} &
\textbf{Score} \\

\midrule\midrule

Naive Baseline & 1.00 & 1.00 & 0.21 & 0.33 & 0.45 & 0.68 & 0.59 & 0.59 & 0.45 & 0.50 & 0.61 \\
SOTA ML$^*$ & - & - & 0.78 & 0.85 & 0.86 & - & - & - & - & 0.74 & - \\

\midrule

\grayrow
\multicolumn{12}{l}{\textit{\textbf{Non-Reasoning Models:}}} \\
\raisebox{-0.25em}{\includegraphics[height=1.2em]{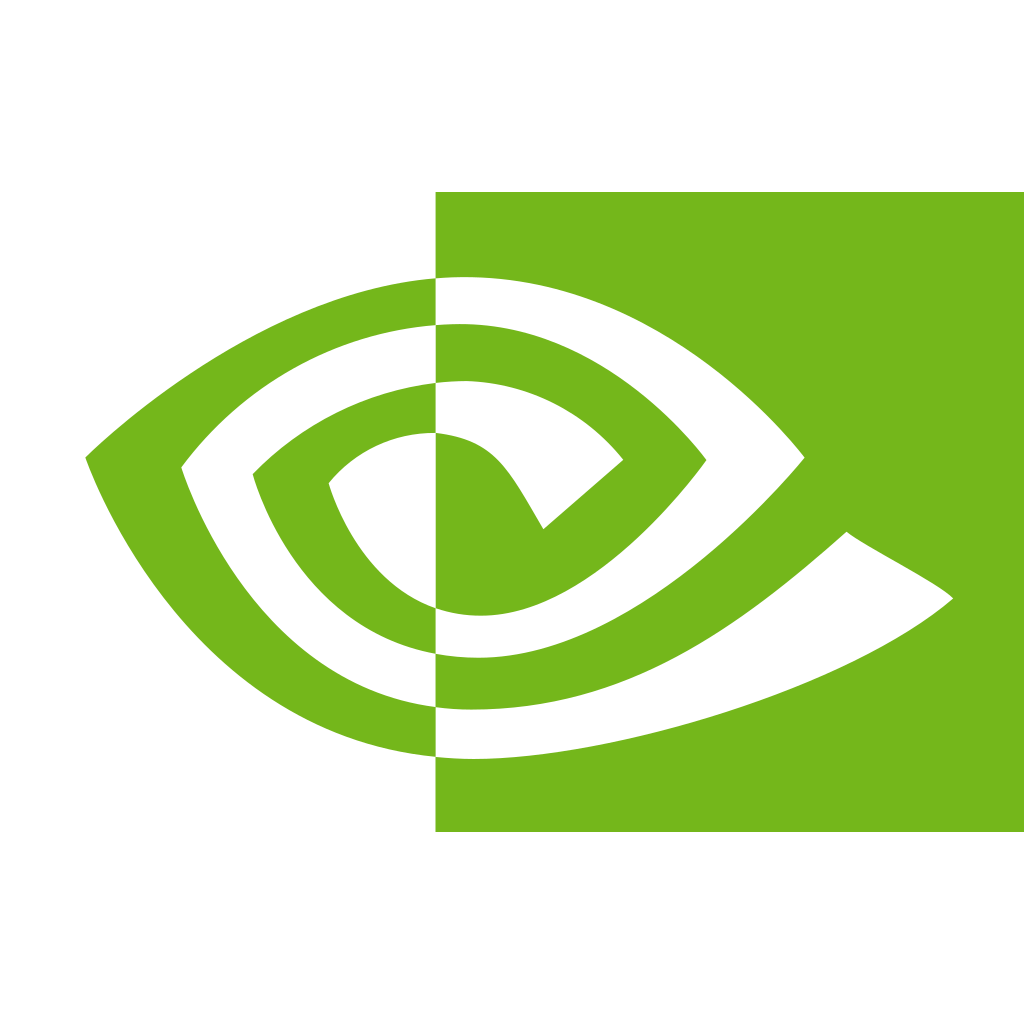}} \nemotron & 0.58 & 0.49 & 0.22 & 0.34 & 0.50 & 0.61 & 0.49 & 0.44 & 0.45 & \ul{0.54} & 0.47 \\
\raisebox{-0.25em}{\includegraphics[height=1.2em]{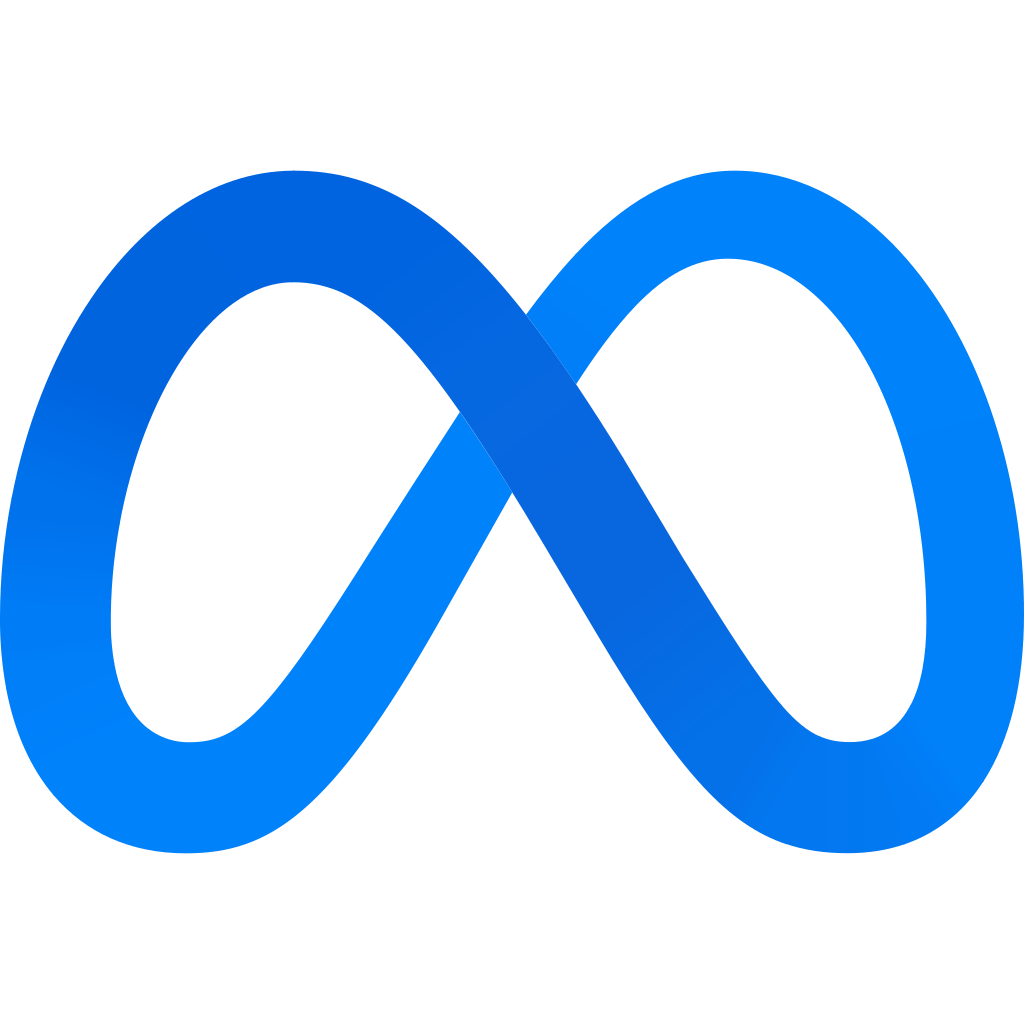}} \llama & 0.75 & 0.71 & 0.30 & 0.40 & \ul{0.56} & 0.75 & 0.59 & 0.54 & 0.50 & 0.49 & 0.58 \\
\raisebox{-0.25em}{\includegraphics[height=1.2em]{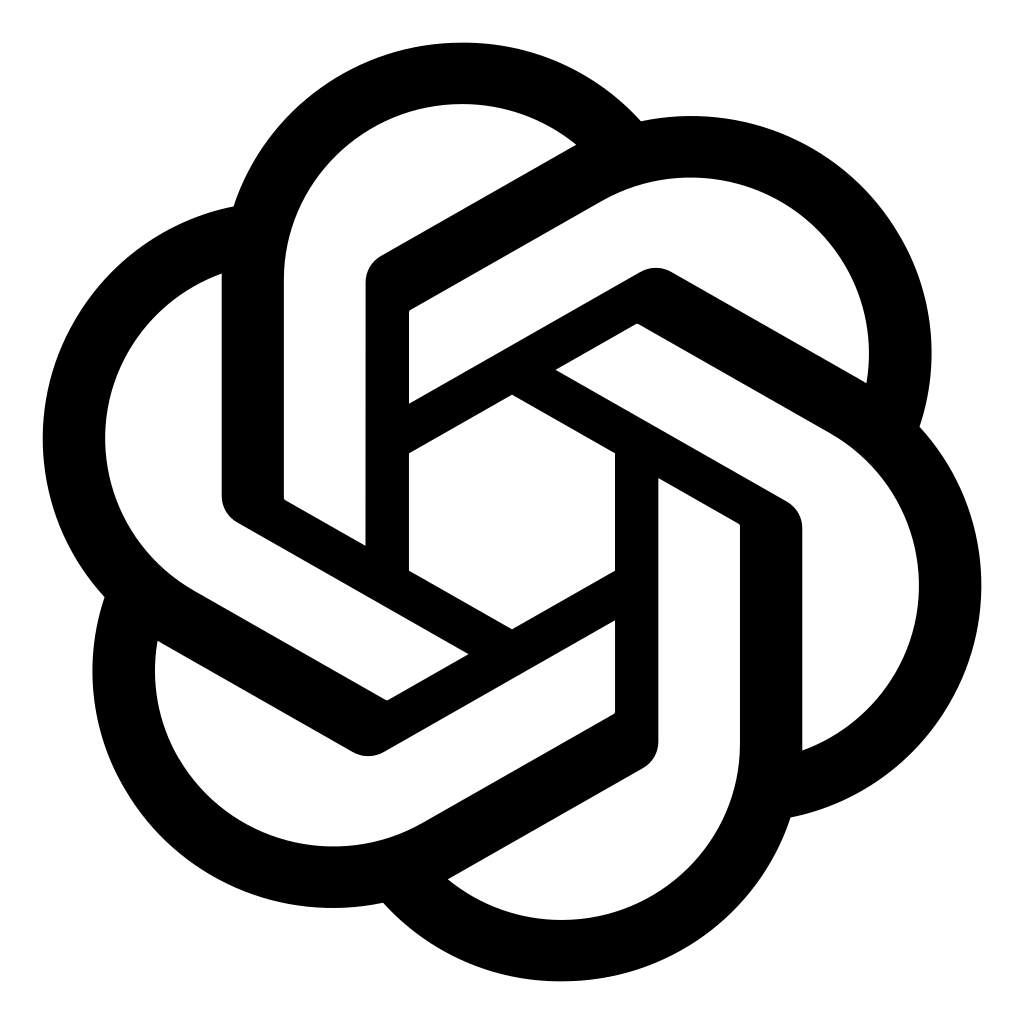}} \gptFourOneMini & 0.89 & 0.77 & 0.29 & 0.43 & \ul{0.56} & 0.79 & 0.63 & 0.57 & 0.56 & \textbf{0.55} & 0.63 \\
\raisebox{-0.25em}{\includegraphics[height=1.2em]{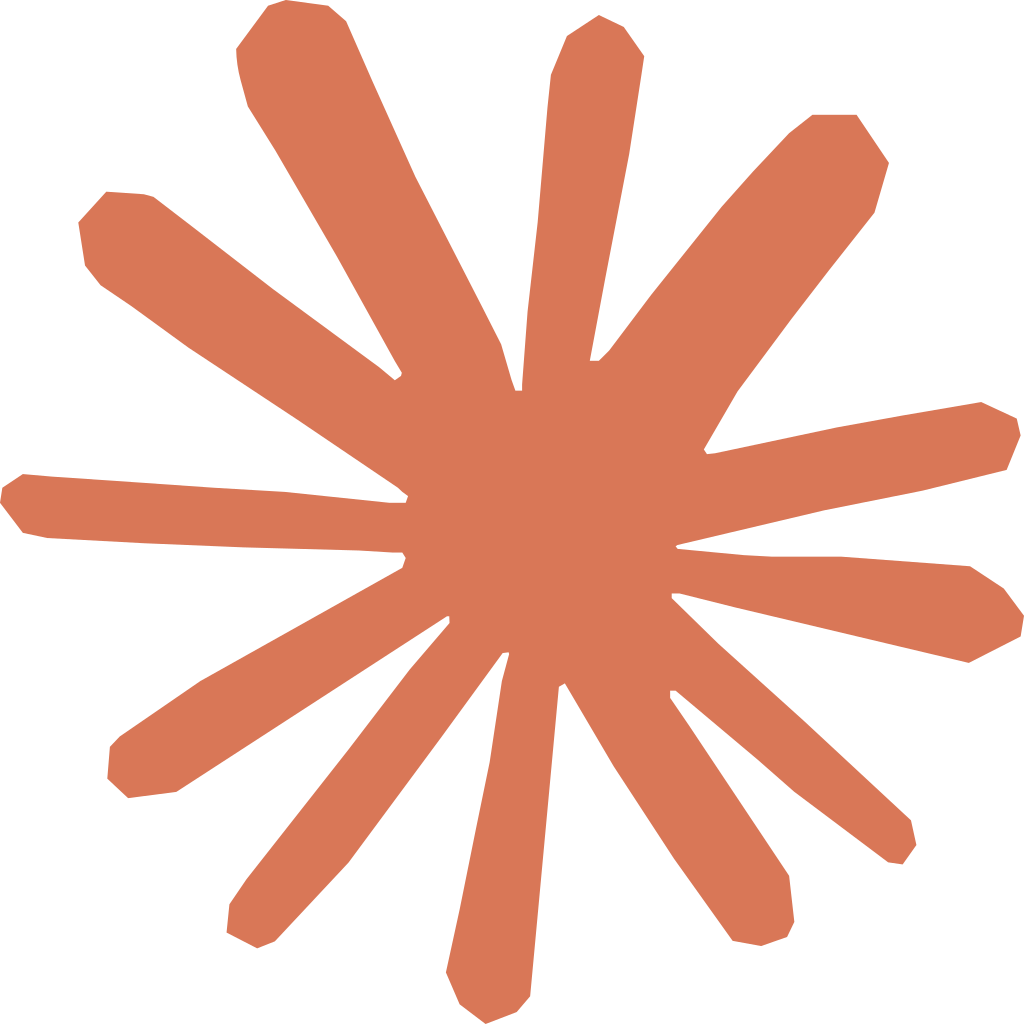}} \claudeHaiku & \ul{0.93} & 0.79 & 0.32 & 0.43 & 0.52 & 0.80 & 0.65 & 0.60 & 0.55 & 0.50 & 0.63 \\
\raisebox{-0.25em}{\includegraphics[height=1.2em]{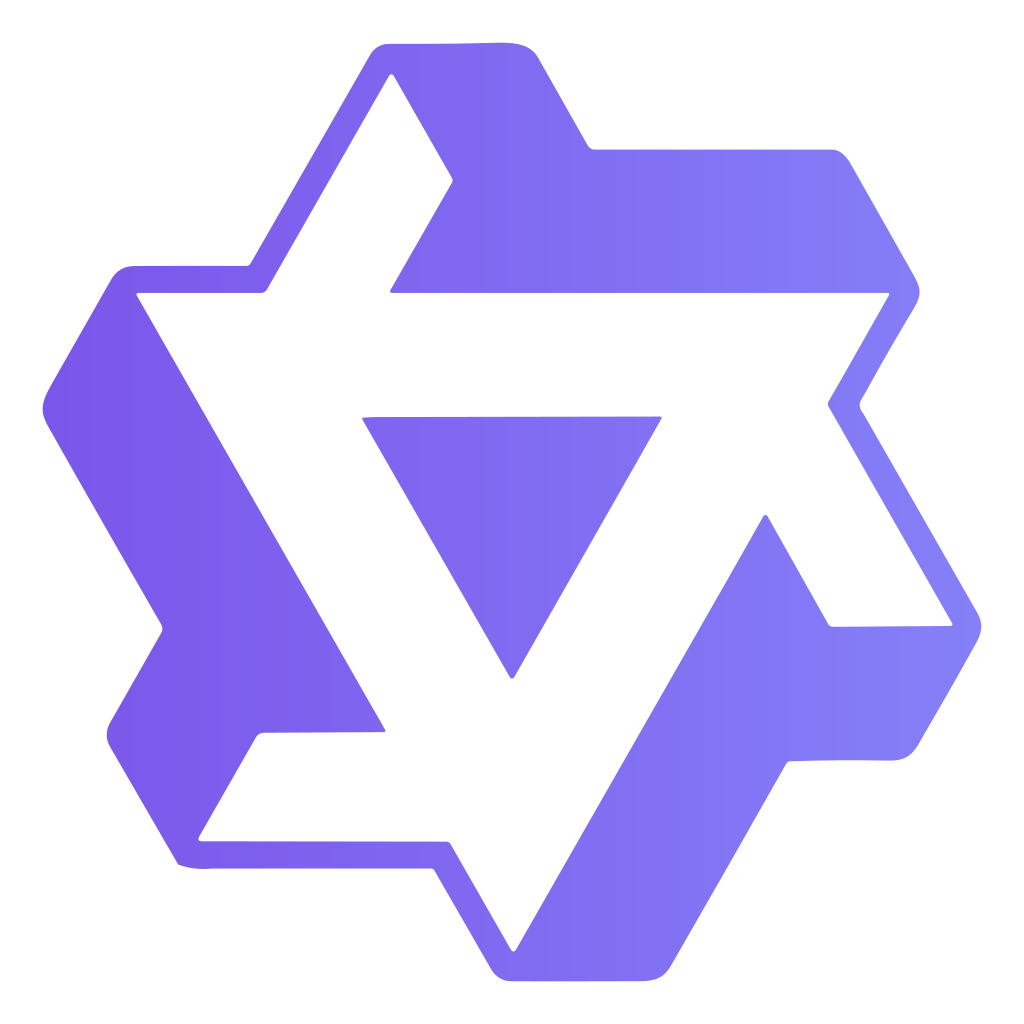}} \qwen & 0.88 & 0.76 & 0.29 & 0.40 & \ul{0.56} & 0.78 & 0.64 & 0.59 & 0.62 & 0.49 & 0.63 \\
\raisebox{-0.25em}{\includegraphics[height=1.2em]{icons/qwen-color.png}} \qwenCoder & 0.92 & 0.80 & 0.32 & 0.43 & 0.55 & 0.82 & 0.66 & \ul{0.66} & 0.51 & 0.47 & 0.63 \\
\raisebox{-0.25em}{\includegraphics[height=1.2em]{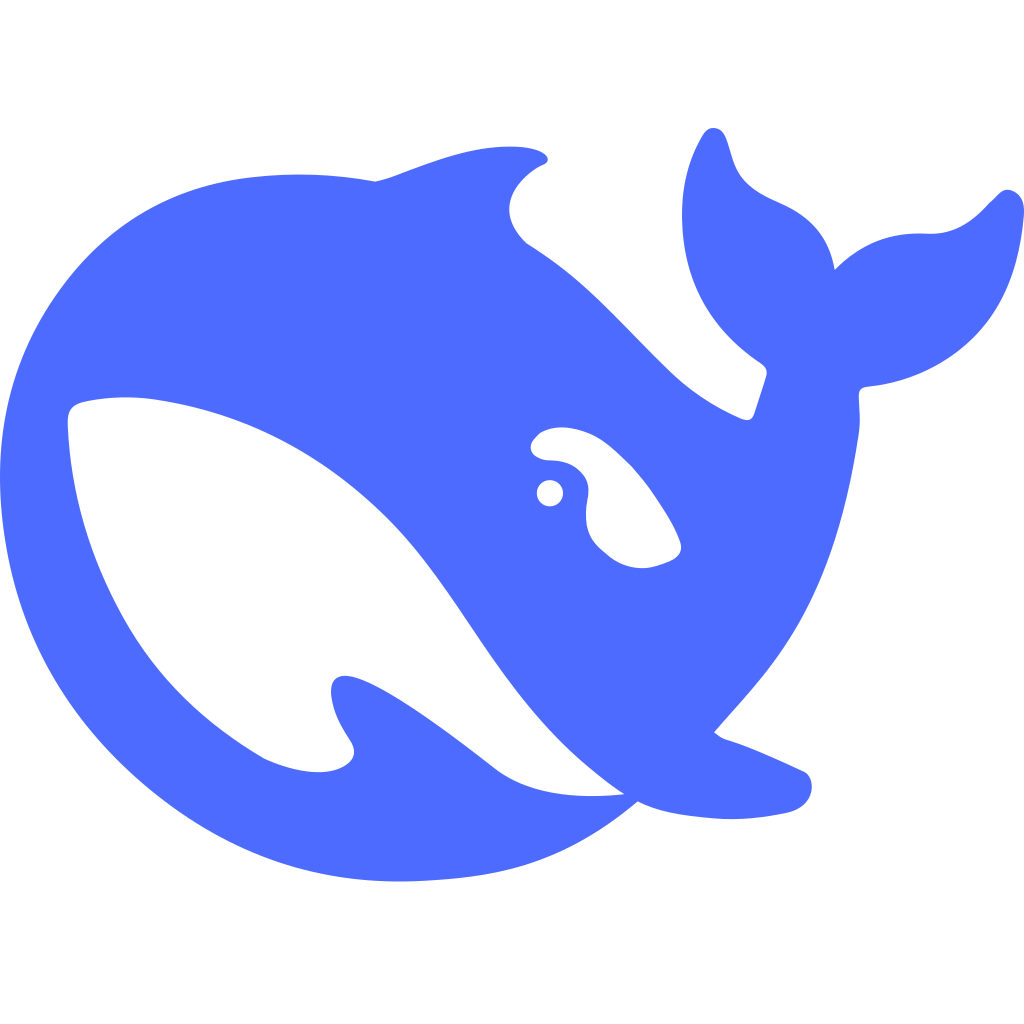}} \deepseek & 0.91 & 0.79 & 0.29 & 0.43 & 0.55 & 0.77 & 0.63 & 0.60 & 0.62 & 0.52 & 0.64 \\

\midrule

\grayrow
\multicolumn{12}{l}{\textit{\textbf{Reasoning Models:}}} \\
\raisebox{-0.25em}{\includegraphics[height=1.2em]{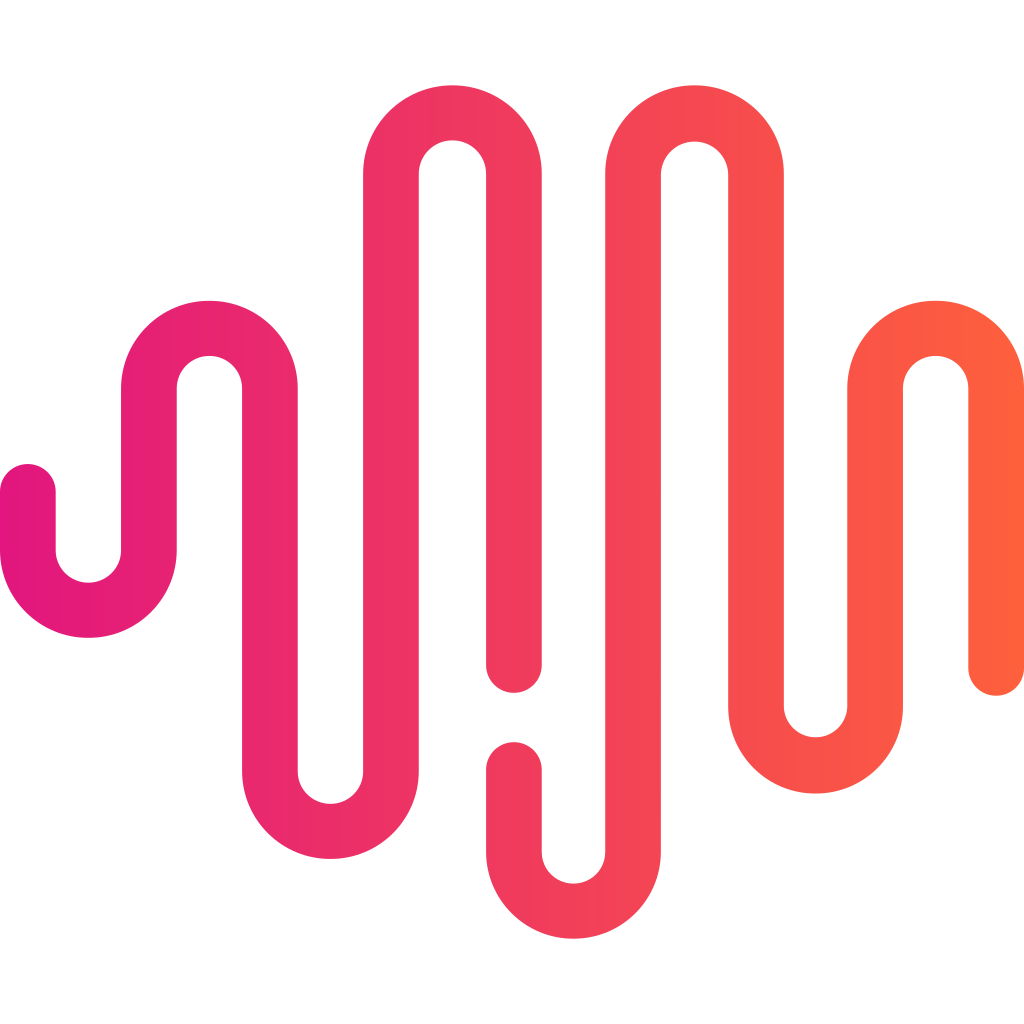}} \minimax & 0.88 & 0.68 & 0.31 & \textbf{0.48} & 0.52 & 0.81 & 0.64 & 0.64 & 0.50 & \textbf{0.55} & 0.61 \\
\raisebox{-0.25em}{\includegraphics[height=1.2em]{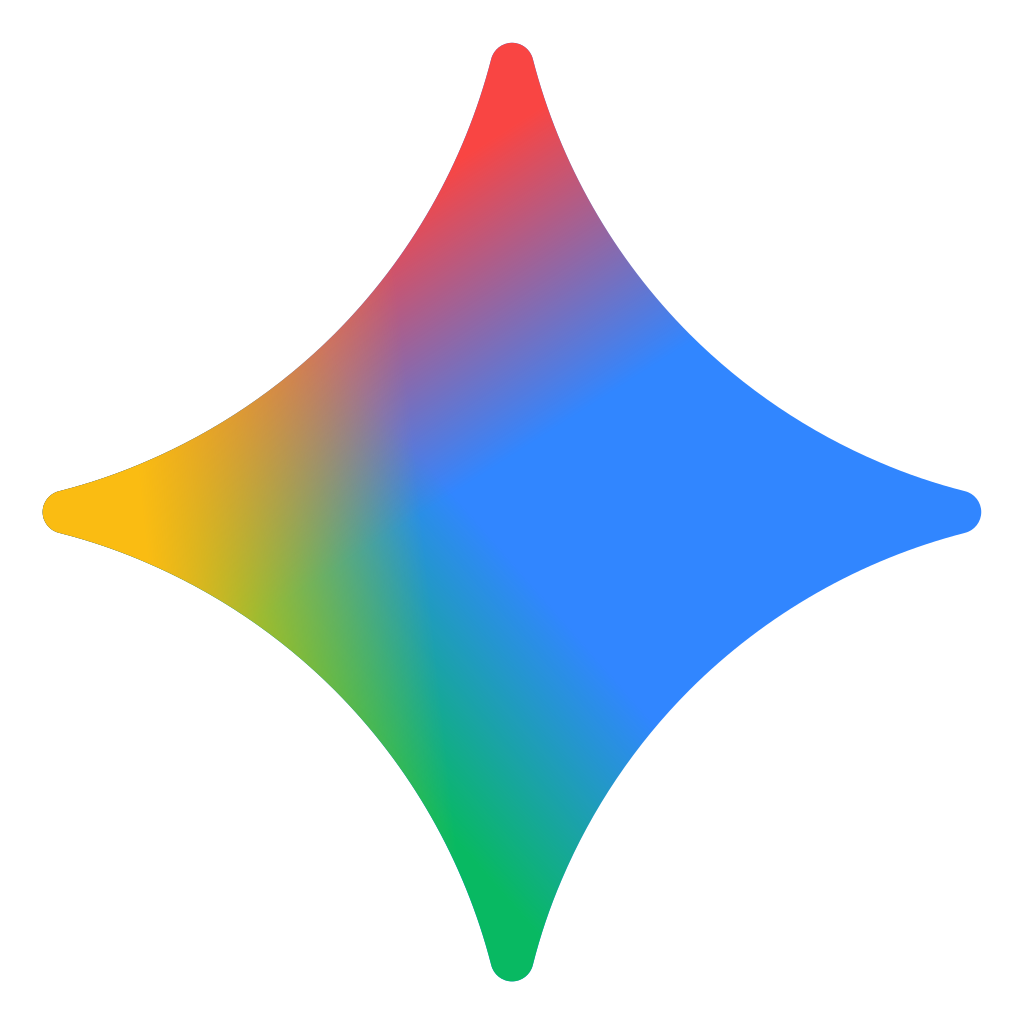}} \geminiPro & 0.82 & 0.74 & 0.32 & 0.40 & 0.51 & 0.79 & 0.65 & 0.65 & 0.59 & 0.45 & 0.62 \\
\raisebox{-0.25em}{\includegraphics[height=1.2em]{icons/gemini-color.png}} \geminiFlash & 0.88 & 0.74 & 0.27 & 0.41 & 0.54 & 0.77 & 0.60 & 0.61 & 0.60 & 0.46 & 0.62 \\
\raisebox{-0.25em}{\includegraphics[height=1.2em]{icons/openai.png}} \gptFiveMini & 0.88 & 0.74 & 0.23 & 0.42 & 0.55 & 0.79 & 0.63 & 0.64 & 0.57 & 0.48 & 0.62 \\
\raisebox{-0.25em}{\includegraphics[height=1.2em]{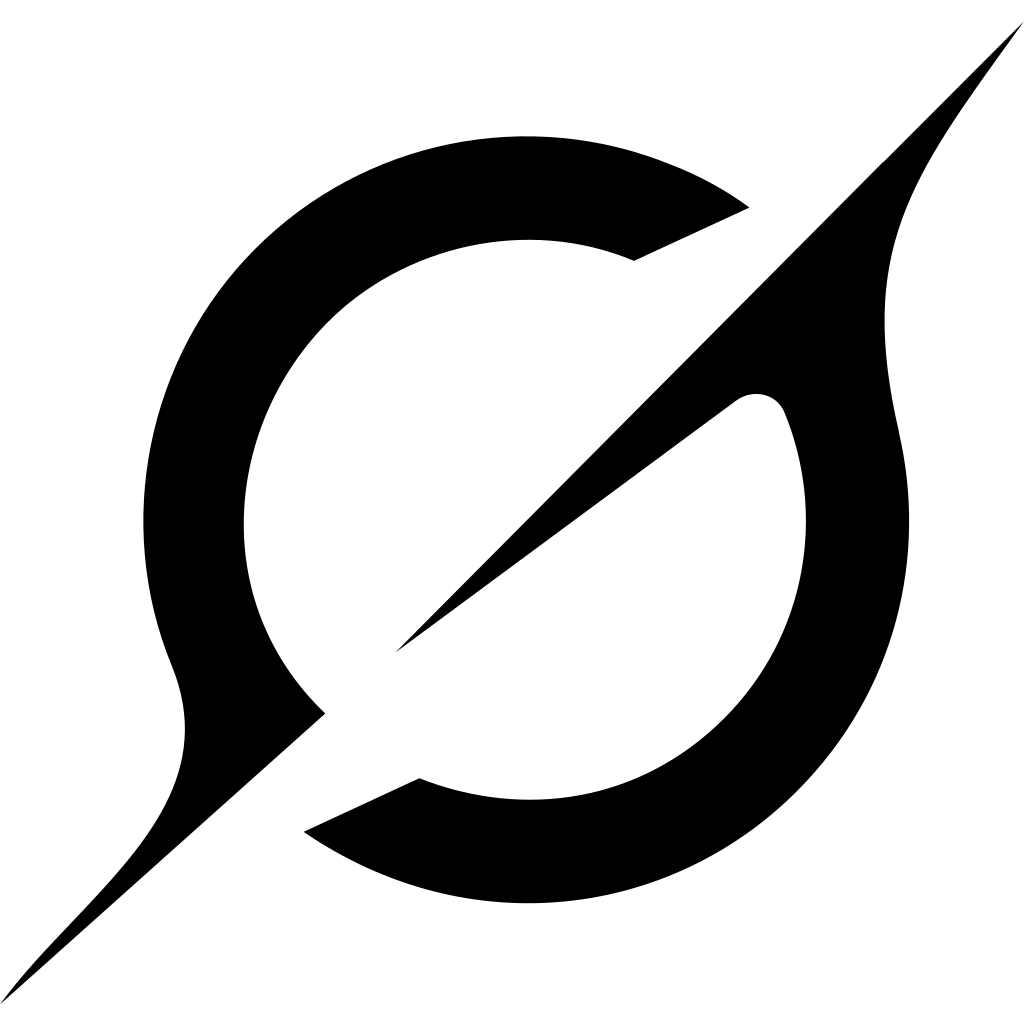}} \grok & \ul{0.93} & 0.78 & 0.34 & 0.44 & 0.55 & 0.83 & 0.66 & 0.63 & 0.60 & 0.45 & 0.65 \\
\raisebox{-0.25em}{\includegraphics[height=1.2em]{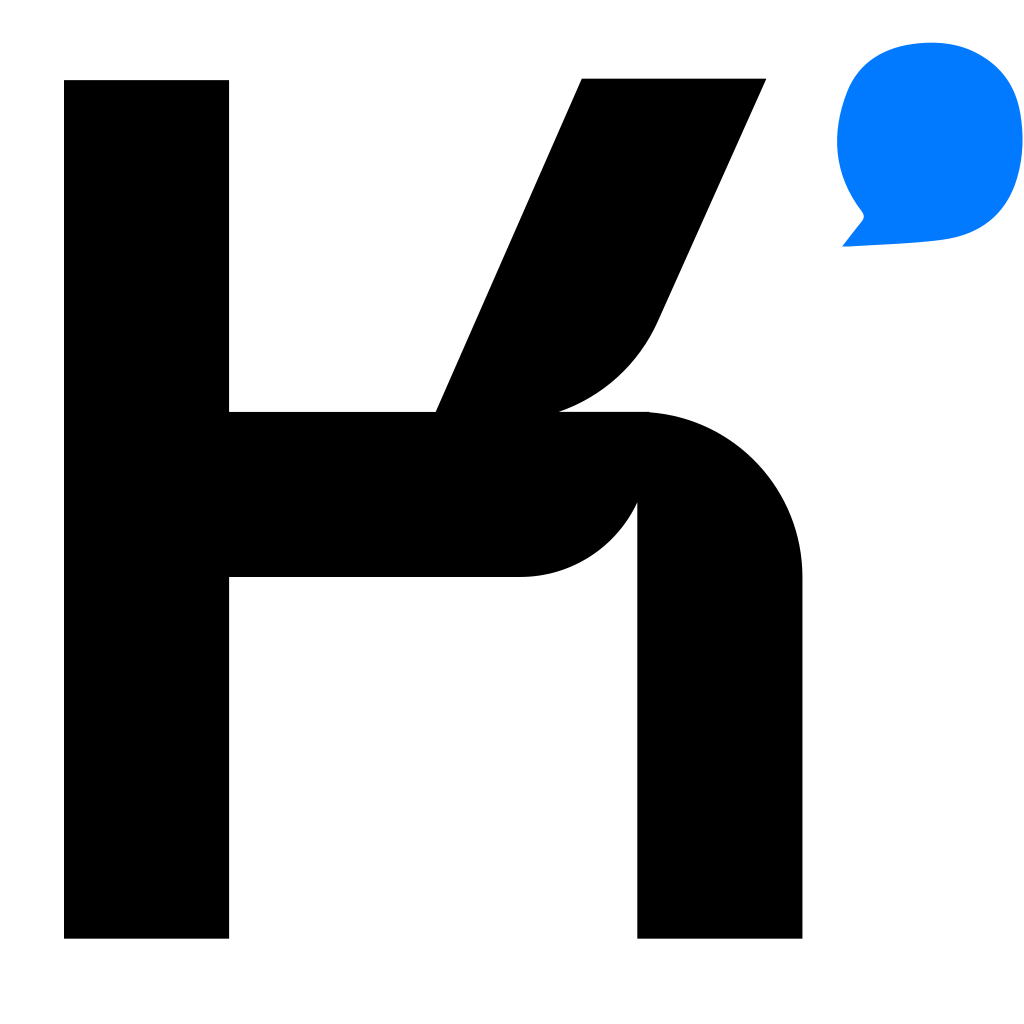}} \kimi & 0.91 & 0.80 & 0.36 & 0.44 & \ul{0.56} & 0.82 & 0.67 & 0.63 & 0.60 & 0.43 & 0.65 \\
\raisebox{-0.25em}{\includegraphics[height=1.2em]{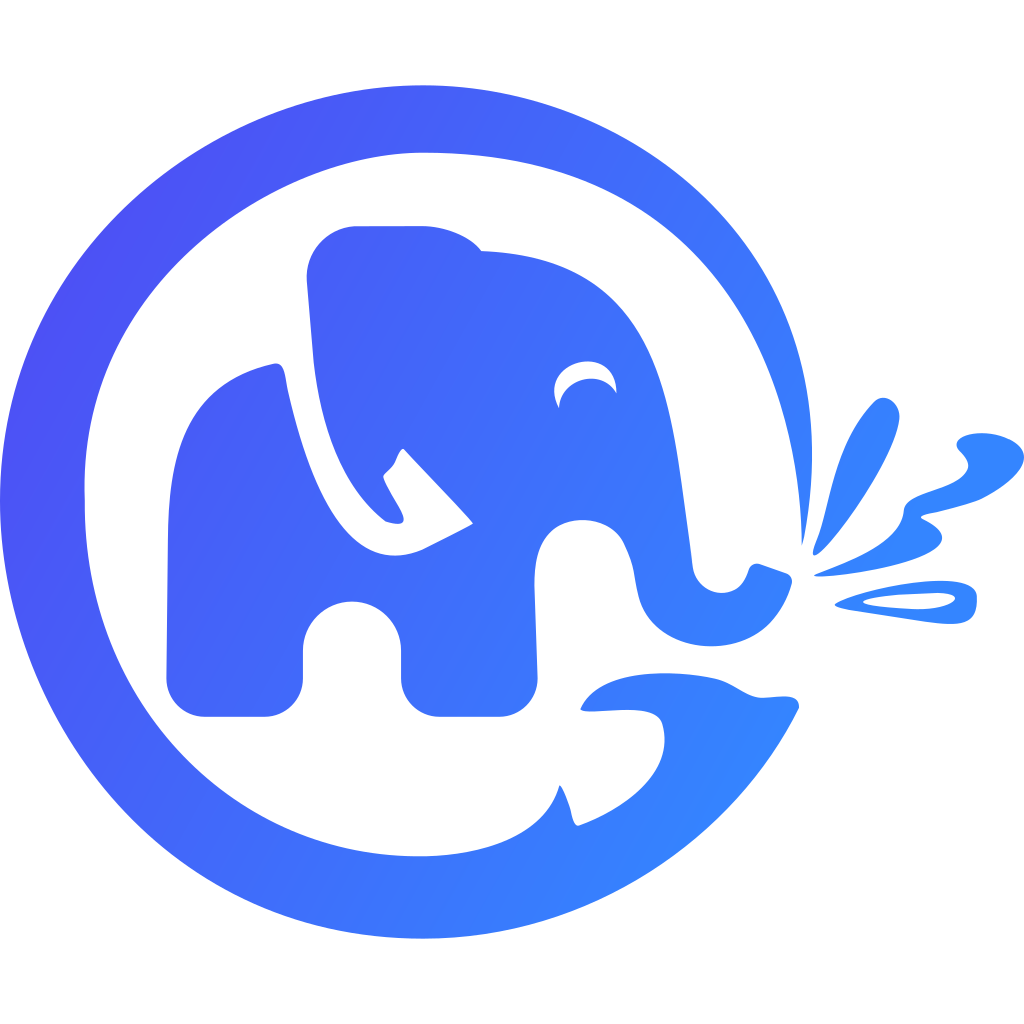}} \glm & 0.92 & 0.80 & 0.38 & \ul{0.45} & 0.55 & 0.82 & \ul{0.68} & 0.64 & 0.60 & 0.47 & 0.66 \\
\raisebox{-0.25em}{\includegraphics[height=1.2em]{icons/chatglm-color.png}} \glmfive & \textbf{0.94} & \ul{0.81} & \ul{0.40} & 0.44 & \ul{0.56} & \ul{0.84} & 0.67 & 0.65 & \ul{0.68} & 0.52 & \ul{0.68} \\
\raisebox{-0.25em}{\includegraphics[height=1.2em]{icons/gemini-color.png}} \geminiProthree & \textbf{0.94} & \textbf{0.82} & \textbf{0.50} & 0.43 & \textbf{0.57} & \textbf{0.85} & \textbf{0.70} & \textbf{0.69} & \textbf{0.70} & 0.39 & \textbf{0.69} \\
\bottomrule[1.5pt]
\end{tabular*}
\label{tab:main_results}
\vspace{-5pt}
\end{table*}

%% file: chapters_arxiv/4_exp_and_anal_arxiv.tex
\vspace{-1pt}
\section{Experiments and Analysis}
\label{sec:exp_and_analysis}

\textbf{Standards and Baselines.} We standardize evaluation by using 200 test cases per task, unless limited by data availability, yielding 20,226 test cases in total. Each task is scored with a metric that matches its output type: Accuracy, IoU, or sMAPE. For the sMAPE-based tasks, we first apply min-max normalization to predictions and ground truth, and report $1 - \frac{1}{2}\text{sMAPE}$ so that scores are strictly bounded in $[0,1]$ (higher is better). We also include a naive baseline to contextualize performance across tasks and metrics. The naive baseline details are provided in Appendix \ref{app_subsec:naive_baseline}.

\textbf{Model Selection.} To address context-window limits when working with long time-series inputs, we standardize evaluation with the CodeAct framework \cite{wangExecutableCodeActions2024}. CodeAct allows an LLM to reason by writing Python code that reads and analyzes data from files, so the model does not need to ingest the full raw sequence in its prompt. To evaluate the model's intrinsic reasoning capabilities, rather than its ability to call advanced tools, we intentionally restrict the environment to minimal Python packages necessary for health time series processing (e.g., \textit{neurokit2}). We evaluate 16 state-of-the-art models spanning 11 families, covering both lightweight open-source models and large proprietary reasoning systems. Specifically, we evaluate: GPT-4.1 and GPT-5 mini \cite{openai2025gpt41mini, singh2025openaigpt5}, \claudeHaiku \cite{anthropic2025haiku}, Gemini 2.5 Pro and 2.5 Flash \cite{comanici2025gemini}, \geminiProthree \cite{googledeepmind2026gemini3.1}, \qwen and \qwenCoder \cite{yang2025qwen3}, \deepseek \cite{liu2024deepseek}, \llama \cite{meta2025llama4maverick}, \grok \cite{xai2025grok}, \minimax \cite{minimax2025m2}, \glm \cite{zhipuai2025glm47} and \glmfive \cite{glm5team2026glm5vibecodingagentic}, \nemotron \cite{basant2025nvidianemotron}, and \kimi \cite{team2025kimi}.

\begin{figure*}[t]
\vspace{-5pt}
\centering
\begin{minipage}[b]{0.48\textwidth}
    \centering
    \includegraphics[width=0.95\linewidth]{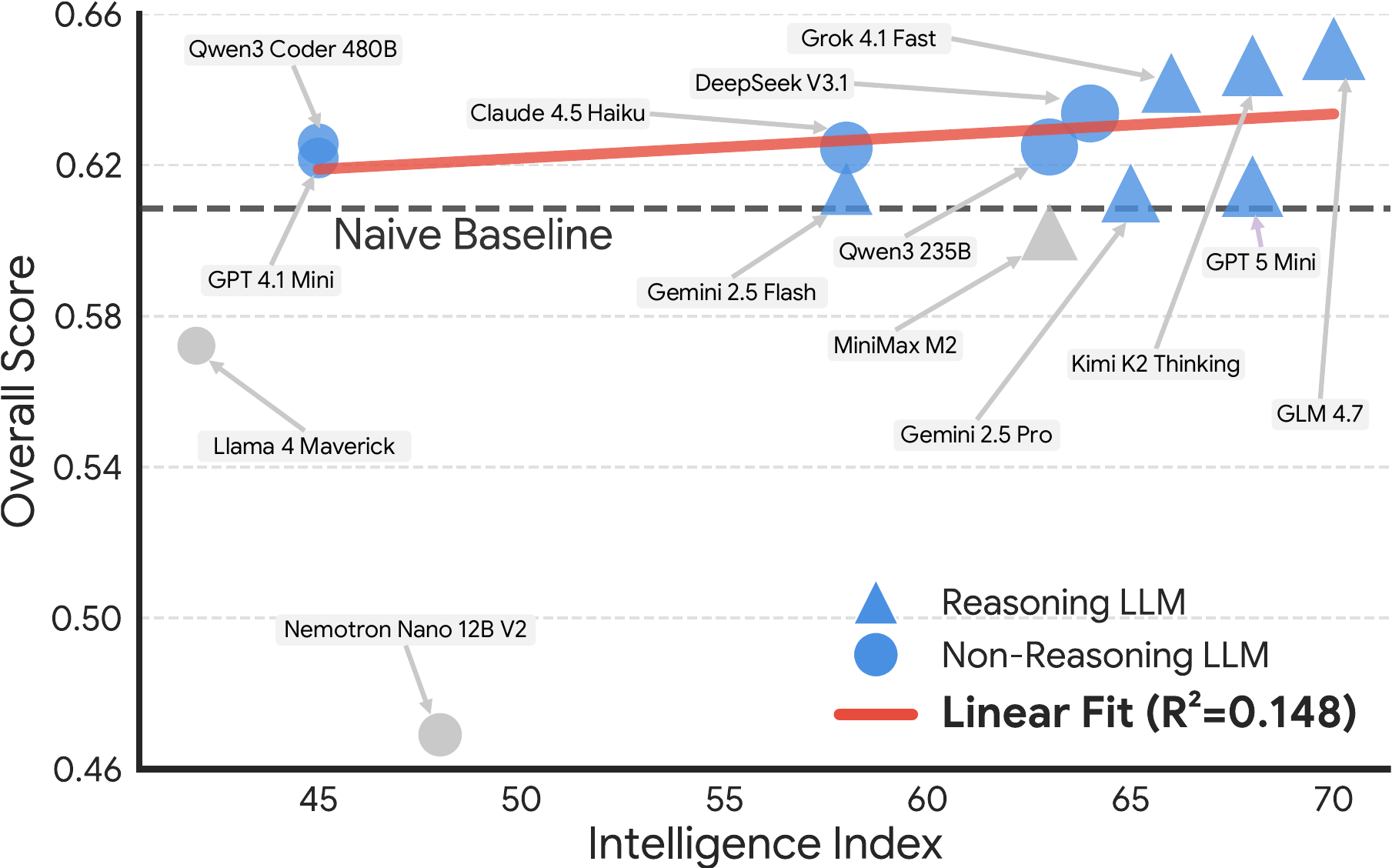}
    \caption{
    \small
    \textbf{Regression analysis of intelligence index vs. performance on \name.} Models performing below the naive baseline are excluded as outliers.}
    \label{fig:intel_score}
\end{minipage}
\hfill
\begin{minipage}[b]{0.48\textwidth}
    \centering
    \includegraphics[width=0.95\linewidth]{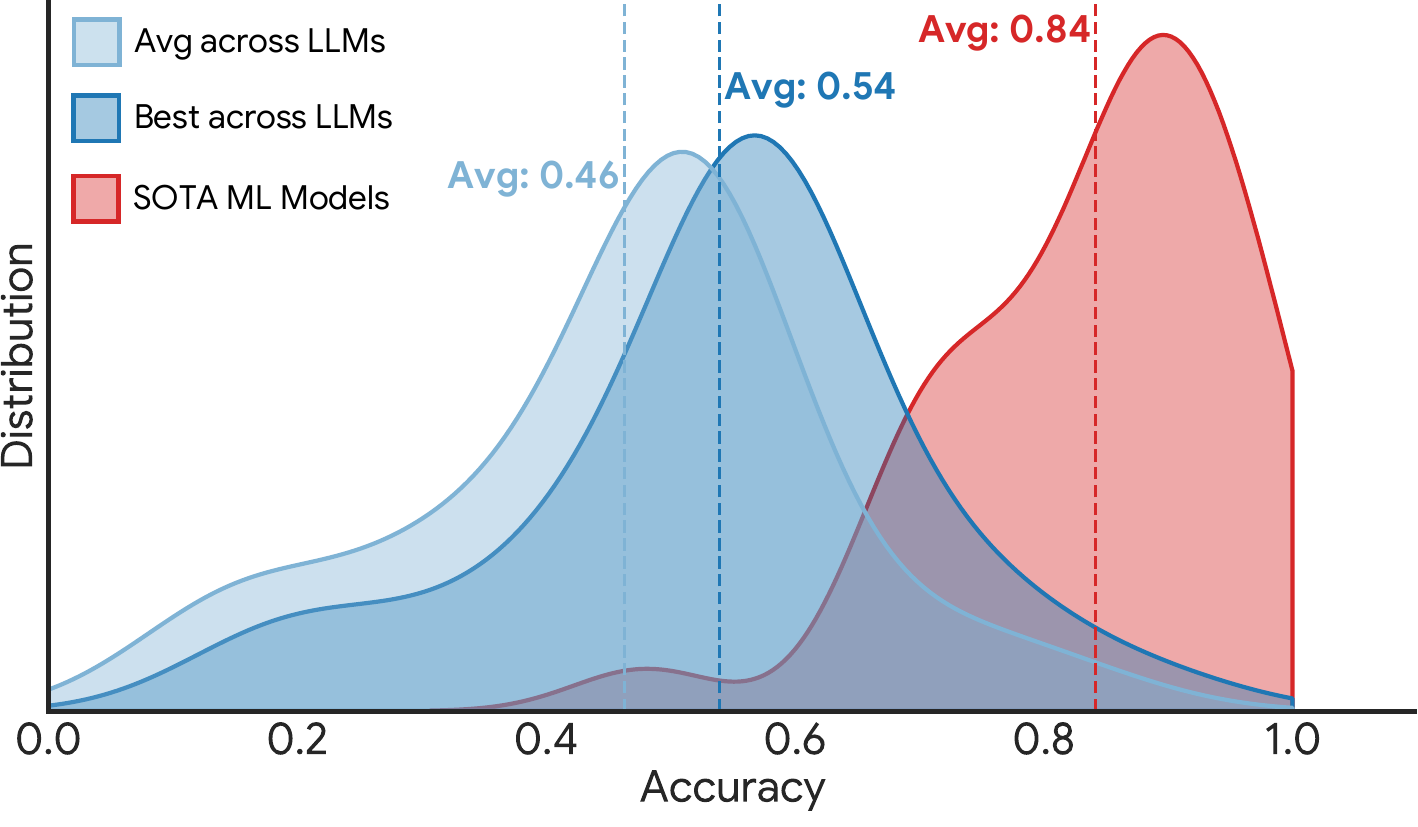}
    \caption{
    \small
    \textbf{Performance comparison of state-of-the-art ML methods and LLMs.} Density reflects the concentration of tasks at each performance level.}
    \label{fig:sota_vs_llm}
\end{minipage}
\vspace{-2pt}
\end{figure*}

\subsection{Are LLMs Good at Health Time-Series Reasoning?}
\label{subsec:over_performance}

\textbf{LLMs make only small gains over a naive baseline, and performance is weakly related to general intelligence indexes.}
Table \ref{tab:main_results} shows that most models outperform the naive baseline, but by modest margins, suggesting that current LLMs still lack strong reasoning ability in the health time-series setting. We further test whether success on \name tracks general reasoning capability using the intelligence index from \cite{kim2025towards} (Appendix~\ref{app_subsec:intel_index}). After removing clear outliers that do not exceed the naive baseline, Fig. \ref{fig:intel_score} shows no meaningful linear correlation between intelligence index and \name performance. This indicates that health time-series reasoning is not well predicted by performance on broad reasoning benchmarks, and likely depends on domain- and data-specific skills that are not captured by general scores.


\textbf{LLMs lag far behind specialized time-series models.}
To quantify this gap, we compare our evaluated LLMs (reporting both average and best performance across 16 models) against published task-specific SOTA baselines on a curated subset of 32 \textit{Inference} and \textit{Deduction} tasks (Fig.~\ref{fig:sota_vs_llm}). We observe a clear separation: specialized models concentrate in a high-performance regime, whereas LLMs are shifted substantially lower, even under best-case selection. Despite minor experimental detail differences, the non-overlapping distributions indicate that general-purpose LLM agents still fall short of the precision achieved by domain-specific methods on challenging physiological reasoning tasks.

Beyond standard ML baselines, we also evaluate OpenTSLM \cite{langer2025opentslm}, a family of time-series language models designed for reasoning over temporal data. Following the original experimental settings, we benchmark OpenTSLM on three representative \name tasks. We find that OpenTSLM underperforms relative to CodeAct-enabled LLMs (complete results are in Appendix \ref{app_subsec:opentslm_exp}).

\begin{tcolorbox}
    \textit{\textbf{Finding 1:}} LLMs lag substantially behind specialized models, and the performance on health time series is only weakly related to their general reasoning capabilities.
\end{tcolorbox}


\begin{figure*}[!t]
\centering
\vspace*{-10pt}
\begin{minipage}[b]{0.48\textwidth}
    \centering
    \includegraphics[width=0.95\linewidth]{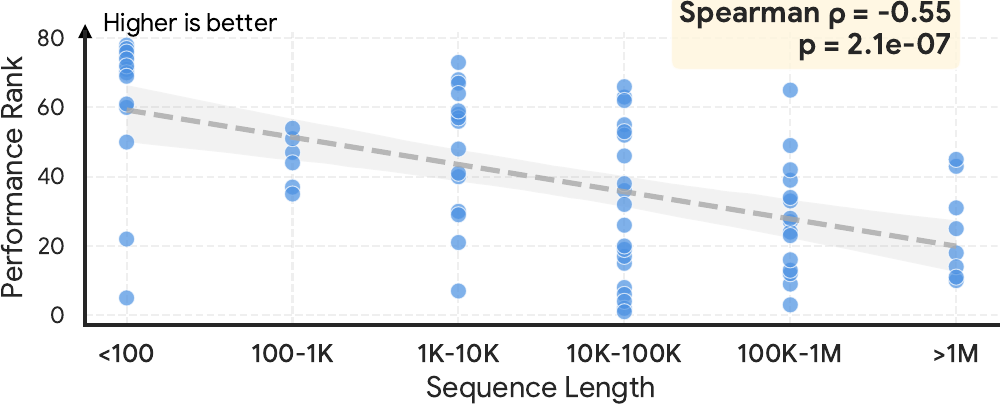}
    \includegraphics[width=0.95\linewidth]{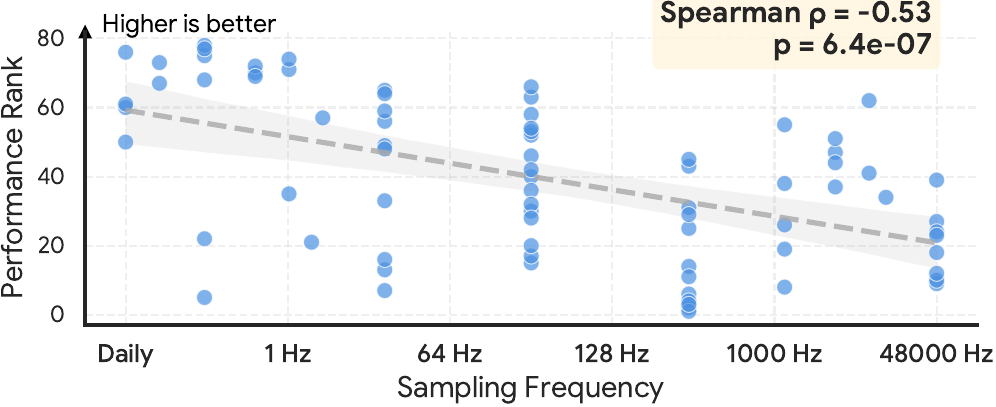}
    \caption{
    \small
    \textbf{Regression analysis of performance gain vs. data complexity.} Each point represents the 16-model average Kappa.}
    \label{fig:seq_freq}
\end{minipage}
\hfill
\begin{minipage}[b]{0.48\textwidth}
    \centering
    \includegraphics[width=\linewidth]{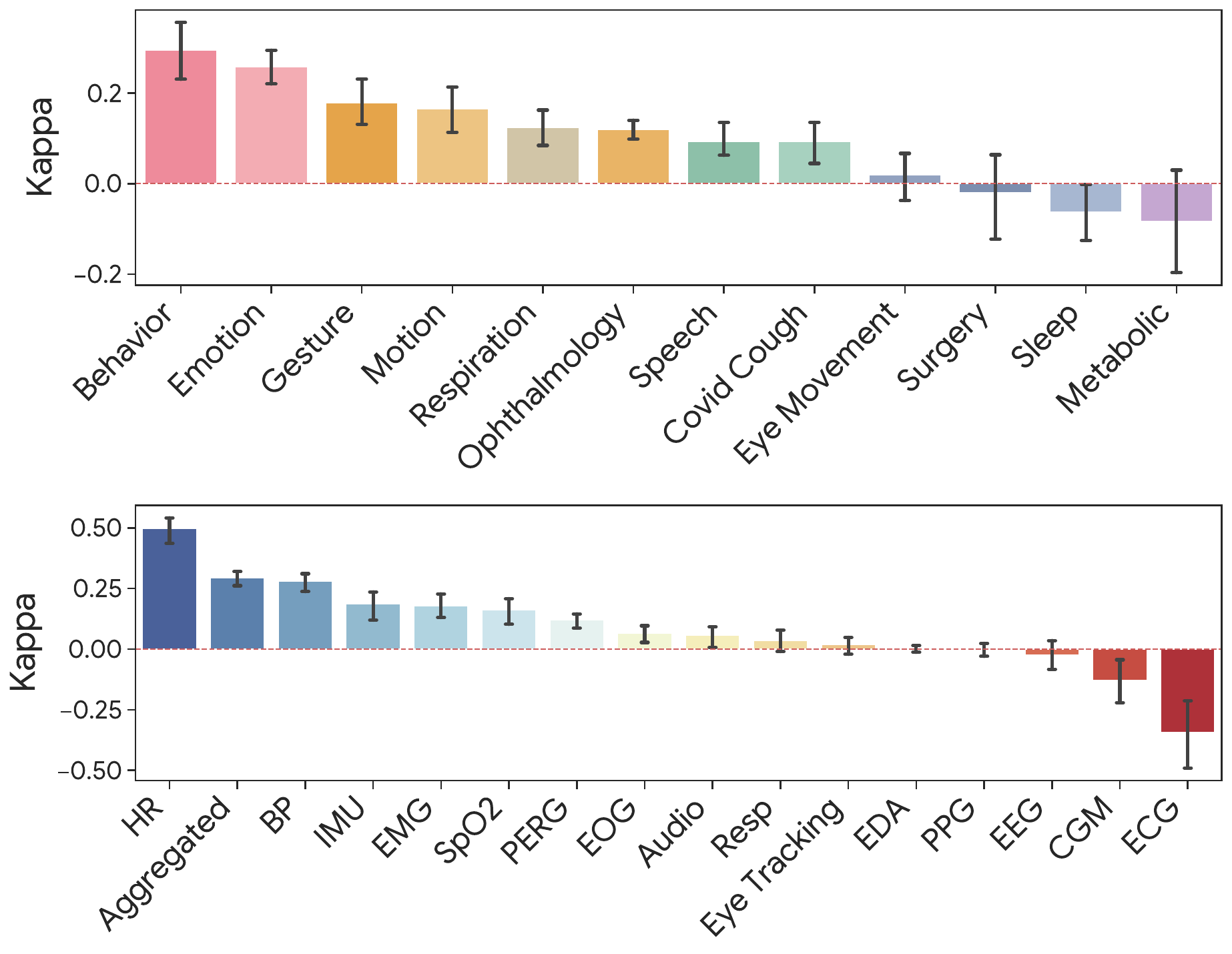}
    \caption{
    \small
    \textbf{Performance gains across different input modalities and domains.}}
    \label{fig:domain_input_bar}
\end{minipage}
\vspace{-5pt}
\end{figure*}

\subsection{Category-Level Behavioral Consistency in LLMs}
\label{subsec:task_category}

\textbf{Perception and Inference tasks depend on explicit rules and data separability.}
In the Perception category, while aggregate performance remains robust, LLMs demonstrate heightened proficiency specifically on tasks with well-defined algorithms or invokable functions (see Appendix~\ref{app_subsec:supp_analysis_task_category}). Inference tasks reveal a polarized performance landscape, hinging largely on the explicitness of task rules and the inherent separability of the data. Models outperform random guessing only on tasks with clear rule-based thresholds or high intrinsic discriminability aligned with their pre-trained knowledge. For instance, agents succeed in scenarios with explicit quantitative thresholds (e.g., hypotension at $65$) or clear common-sense patterns (e.g., reduced mobility during COVID). Conversely, performance deteriorates significantly in scenarios requiring fine-grained distinctions without reference, or specialized domain feature extraction (e.g., raw EMG interpretation). Notably, Subject Profiling degrades to near-random guessing without explicit medical thresholds (e.g., diabetes CV), highlighting the model's inability to capture transient events or derive holistic insights without priors.

\textbf{Generation and Deduction tasks are governed by low-complexity heuristics.} Performance on Generation tasks is characterized by a reliance on low-complexity heuristics rather than deep temporal reasoning. Across Signal Imputation and Future Forecasting tasks, LLM strategies are remarkably simplistic, predominantly defaulting to copy-pasting with noise injection, linear interpolation, basic statistical averaging, or regression fitting. Even in the rare instances where LLMs deploy auto-regressive models, the resulting generations remain no more sophisticated. A similar trend is observed in Cross-modal Translation tasks, where behavior is highly consistent yet formulaic. Instead of employing novel generative synthesis, models revert to standard deterministic signal processing routines, such as peak finding for heart rate estimation, thereby revealing a tendency to treat generation as a mechanical transformation rather than a semantic reconstruction. Performance on Deduction tasks parallels the limitations observed in inference, suggesting a deficiency in modeling temporal causality and physical signal continuity. Successful predictions appear to depend heavily on matching salient signal features to generalized priors, such as associating delta waves with early sleep stages, rather than capturing underlying dynamics. A comprehensive breakdown of experimental results across all categories is presented in Table~\ref{tab:main_results}.

\begin{tcolorbox}
    \textit{\textbf{Finding 2:}} LLMs rely heavily on low-complexity heuristics and explicit priors, and show limited deep temporal reasoning across task categories.
\end{tcolorbox}

\subsection{Impact of Temporal Resolution and Data Properties}
\label{subsec:temp_reso_data_property}
\textbf{Longer input sequences and higher sampling frequencies correlate with worse performance.}
Our statistical analysis regarding input sequence length and sampling frequency validates that increased temporal resolution negatively impacts reasoning efficacy (Fig.~\ref{fig:seq_freq}). To rigorously isolate the effect of temporal resolution, we exclude six outlier tasks whose performance is confounded by intrinsic task characteristics, namely low-frequency tasks dominated by complex biological reasoning rather than data scale, and high-frequency tasks that are solvable through exploitable statistical patterns or low-variability outputs. Within the filtered set, we identify a significant inverse correlation between performance and input sequence length, alongside a parallel negative trend regarding sampling frequency. These results quantify the "cognitive burden" of high-resolution temporal data, confirming that current LLMs struggle to maintain performance as temporal horizons expand and signal density increases.

\begin{wrapfigure}{r}{0.55\textwidth}
    \vspace{-5pt}
    \centering
    \includegraphics[width=0.99\linewidth]{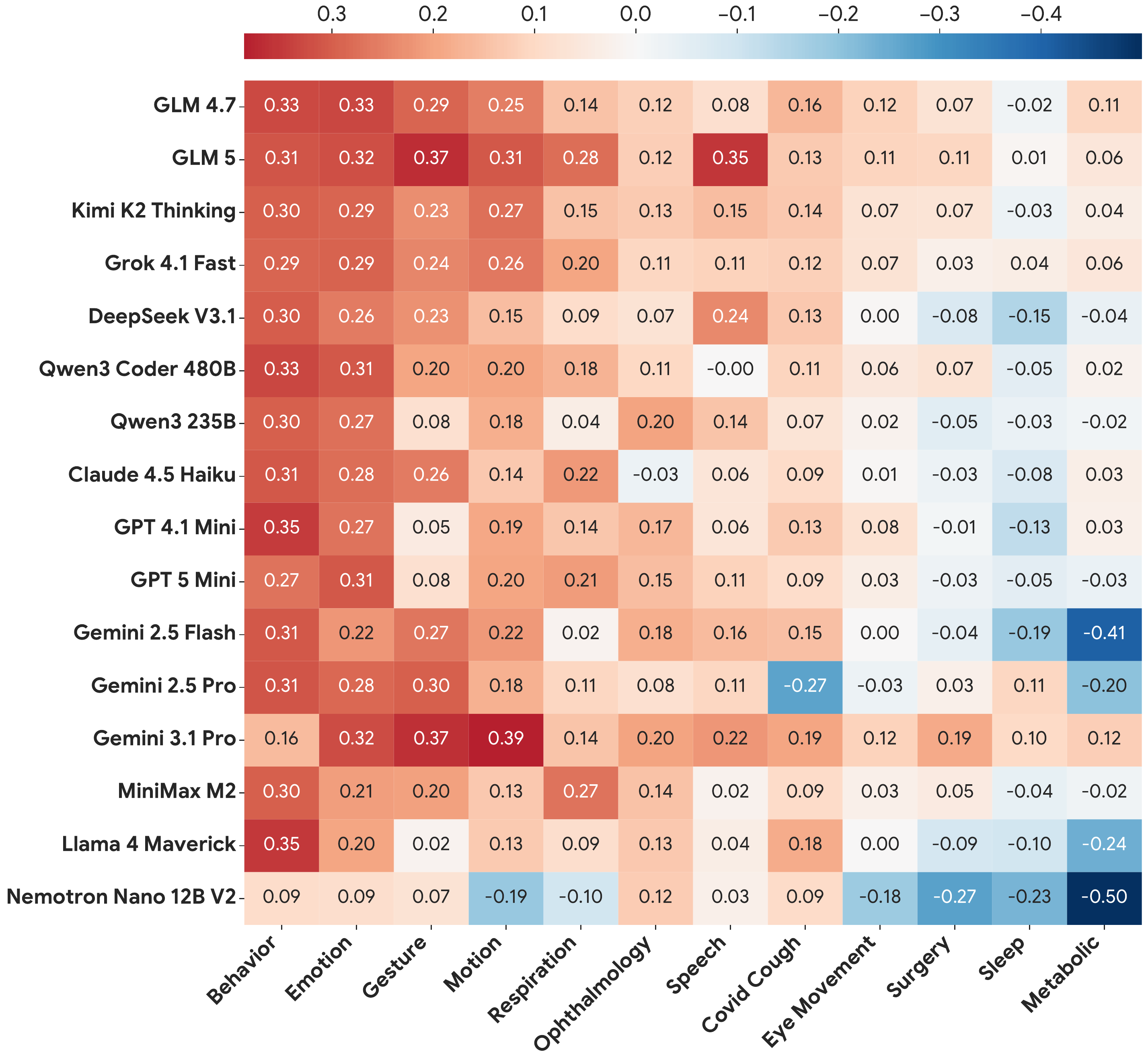}
    \caption{\small\textbf{Model performance heatmap across domains} The heatmap illustrates the Kappa scores across 12 domains. Heatmap for different input modalities is provided in Appendix~\ref{app_subsec:modality_heatmap}.}
    \label{fig:llm_domain_heatmap}
\end{wrapfigure}

\textbf{LLMs show large performance differences across signal modalities and domains.} To quantify these variations, we employ the Kappa coefficient, providing a normalized view of improvement over the naive baseline across all domains and input modalities. Specifically, to eliminate confounders arising from channel multiplicity, our modality-wise analysis is restricted to a subset of 81 single-channel tasks covering 16 distinct modalities. As illustrated in Fig.~\ref{fig:domain_input_bar}, the 95\% confidence intervals reveal significant performance stratification, confirming that model proficiency varies across input modality and domain. Furthermore, the performance heatmaps across 16 models (Fig.~\ref{fig:llm_domain_heatmap}) uncover a striking inter-model consistency: LLMs exhibit a synchronized competence profile, where domains accessible to one model are generally tractable for the entire cohort. This suggests a universal hierarchy of difficulty intrinsic to the health time-series domains themselves, rather than variance driven by specific model architectures.

\begin{tcolorbox}
    \textit{\textbf{Finding 3:}} Performance degrades with temporal complexity, with a shared, model-agnostic difficulty ordering across domains and input modalities.
\end{tcolorbox}

\begin{figure}[h!]
    \vspace{-5pt}
    \centering
    \includegraphics[width=0.9\linewidth]{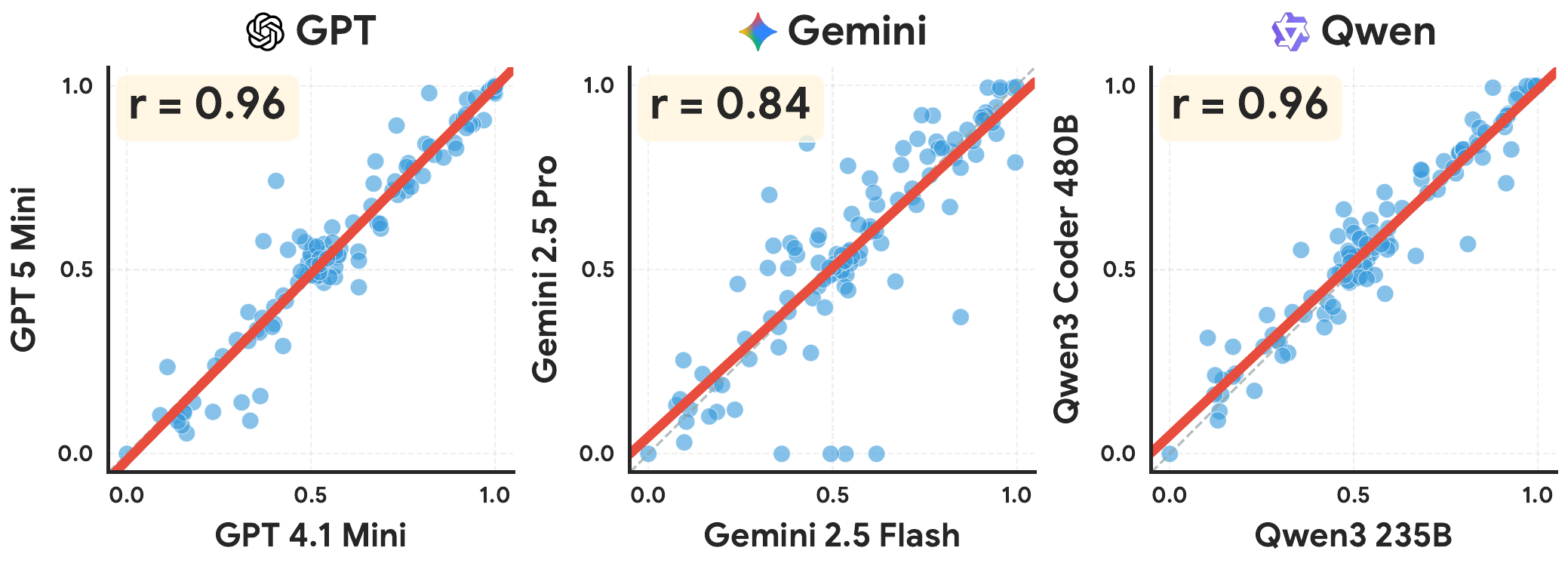}
    \caption{\small\textbf{Analysis of behavioral consistency across LLM families.} Each dot denotes a task, with coordinates indicating the performance of the two compared models.}
    \label{fig:llm_family}
    \vspace{-10pt}
\end{figure}

\subsection{Performance Across LLM Families}
\label{subsec:llm_family}

\textbf{Models within the same family show strong consistency in task performance.} We quantify the alignment of reasoning behaviors within model families by conducting a Pearson correlation analysis on absolute task scores for the GPT, Qwen, and Gemini series. Results in Fig.~\ref{fig:llm_family} present a remarkable intra-family consistency: all three families demonstrate strong linear correlations. Models sharing an architectural lineage possess a synchronized competence profile, where tasks that are challenging for one variant prove consistently difficult for its counterparts. Interestingly, this behavioral alignment coexists with limited gains in absolute performance, as the regression lines closely follow $y=x$. This suggests that while architectural design governs the distribution of task difficulty, increased model scale or generation recency does not strictly guarantee superior reasoning capabilities in the health time-series domain.

\begin{tcolorbox}
    \textit{\textbf{Finding 4:}} Models within the same architectural family exhibit highly consistent behavioral patterns, while scaling does not necessarily yield improved proficiency.
\end{tcolorbox}

\subsection{Does Input Format Affect Performance?}
\label{subsec:input_format}

\begin{wrapfigure}{r}{0.6\textwidth}
    \vspace{-5pt}
    \centering
    \includegraphics[width=\linewidth]{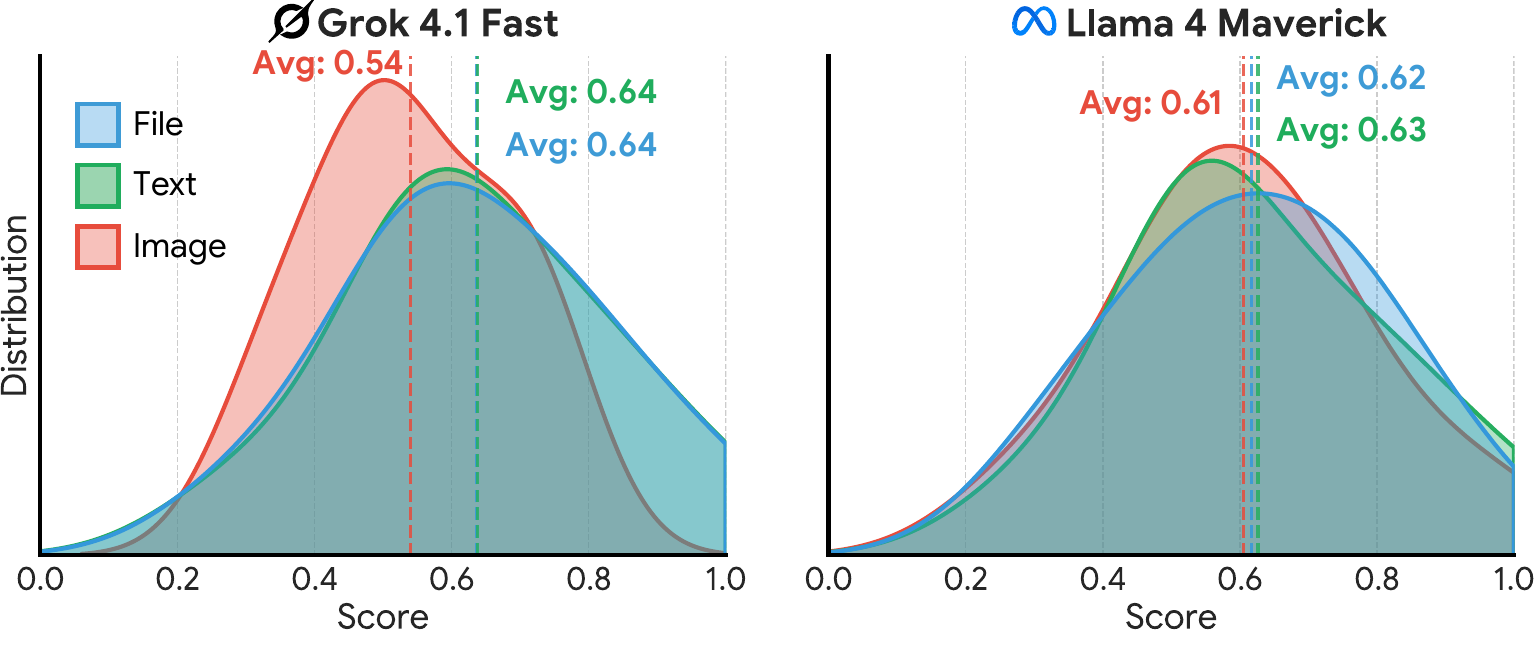}
    \caption{\small\textbf{Performance comparison of Llama 4 Maverick and Grok 4.1 Fast across different input formats.}}
    \label{fig:input_format}
    \vspace{-10pt}
\end{wrapfigure}

\textbf{Input format has only a moderate effect, with visual ingestion underperforming text-based formats.}
We extend beyond CodeAct’s file-access mechanism to assess the impact of alternative input formats, including textual and visual representations, on reasoning efficacy. To isolate the impact of format without confounding factors like context window exhaustion, we conduct this ablation on a curated subset of 10 short-sequence tasks ($L < 1000$, see Appendix~\ref{app_subsec:input_format}). Experiments using \llama and \grok (Fig.~\ref{fig:input_format}) reveal a modality-dependent performance hierarchy. While direct text input and file access achieve near-parity with negligible score differentials, visual ingestion causes notable performance attenuation (e.g., Grok drops to 0.54). Despite this, the distributional morphology of the scores remains structurally analogous across all formats. This suggests that while visual encoding may lose fine-grained fidelity, the fundamental semantic processing of time-series data remains consistent regardless of the input modality.

\begin{tcolorbox}
    \textit{\textbf{Finding 5:}} Input format mainly shifts absolute performance, while relative task difficulty remains consistent across formats.
\end{tcolorbox}

\subsection{Additional Studies}
\label{subsec:additional_studies}
\textbf{More input information does not necessarily improve performance.} We examine a curated subset of 13 tasks (organized into 6 pairs) selected from our 110-task benchmark, spanning the sleep, surgery, metabolic, and emotion domains. Each pair shares identical settings, differing solely in whether auxiliary signals are provided to aid inference. Contrary to expectations, adding these signals often fails to improve and can even degrade performance relative to single-channel baselines. We identify two primary failure modes: contextual neglect, where agents disregard auxiliary data to rely exclusively on the target channel, and informational distraction, where complex multimodal context is misinterpreted as noise or induces redundant reasoning steps. Detailed breakdowns are available in Appendix~\ref{app_subsec:supp_casestudy}.

\begin{wrapfigure}{r}{0.6\textwidth}
    \vspace{-5pt}
    \centering
    \includegraphics[width=\linewidth]{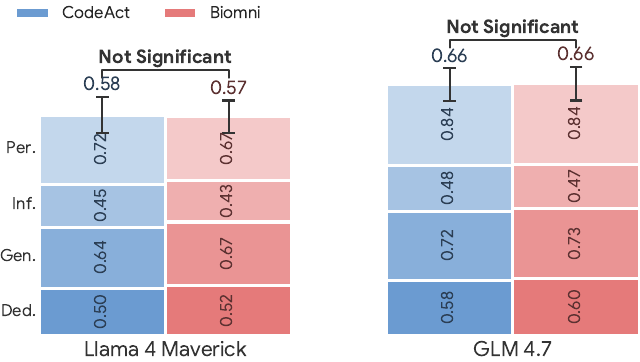}
    \caption{\small\textbf{Comparisons between Biomni and CodeAct.}}
    \label{fig:biomni}
    \vspace{-10pt}
\end{wrapfigure}

\textbf{Domain-specialist agents do not necessarily outperform generalist agents.} We investigate whether adopting a domain-specific agentic framework enhances performance by benchmarking against Biomni \cite{huang2025biomni}, a general-purpose biomedical agent. To ensure a rigorous comparison across disparate capability levels, we implemented Biomni using both \glm (representing high-performing models) and \llama (representing lower-tier models) as backbones. Counter-intuitively, we observe that the specialized Biomni architecture fails to yield statistically significant (Paired T-test) improvements over the standard CodeAct agent (Fig.~\ref{fig:biomni}). A granular analysis of agent trajectories (Appendix~\ref{app_subsec:supp_casestudy}) reveals the underlying cause: Biomni's retrieved action space, while comprehensive for tasks like gene prioritization or drug repurposing, suffers from toolset sparsity regarding health time-series analysis. Consequently, the agent rarely invokes its specialized library, effectively reverting to generic reasoning. This highlights a critical insight: broad biomedical domain expertise cannot compensate for the absence of modality-specific computational primitives in time-series processing.

%% file: chapters_arxiv/5_discussion_arxiv.tex
\section{Discussion}
\label{sec:discussion}

\textbf{Limitations.} A limitation of \name is that we do not yet provide a systematic, task-agnostic evaluation of agent reasoning trajectories. While \name measures final-task performance reliably, assessing the quality and faithfulness of intermediate steps remains difficult, and current approaches (e.g., LLM-as-a-judge scoring) are not reliable enough for absolute evaluation. In addition, many failures suggest that current agents lack domain prior knowledge needed for robust decisions on physiological signals; designing principled ways to incorporate such priors (e.g., via tools, constraints, or verified clinical rules) is an important direction.

\textbf{Conclusion.}
We present \name, a unified benchmark that evaluates LLMs on health time series beyond narrow analytics and toward hierarchical reasoning. \name spans 110 tasks across 12 healthcare domains and 20 signal modalities, enabling controlled evaluation of perception, inference, generation, and deduction. Our results show a consistent gap: models perform reasonably on basic signal perception but struggle on higher-level reasoning that requires long-horizon integration, causal thinking, and temporal directionality. Finally, \name is released as a community-driven, living benchmark so the evaluation can expand with new datasets, tasks, and agents as the field progresses.

%% file: chapters_arxiv/6_appendix_arxiv.tex
\newpage
\appendix
\onecolumn

\section{Details of \name Design}
\label{app_sec:hearts_Design}

\subsection{Data Statistics}
\label{app_subsec:data_statistics}
We provide a statistical breakdown of the datasets curated in \name in Table~\ref{tab:data_stats}. To characterize the heterogeneity of the benchmark, we detail the specific signal modalities and their corresponding sampling frequencies for each dataset, alongside the number of samples utilized in \name. To ensure transparency and support community-driven reproducibility, we detail the access protocols for each dataset.

\input{tables/table_Appendix_data_statistics}

\subsection{Task Design Detail}
\label{app_subsec:task_design_detail}
Below are the specifications of the task design for all 110 tasks included in \name. For each task, we report a unified schema including the task name, input and output formats, evaluation metrics, and the temporal granularity. For tasks with multiple input channels, the frequency and sequence length are determined by the maximum value across all channels.

\subsubsection{Capture24}
\textbf{Perception - Feature Extraction}
\begin{description}[itemsep=0pt, topsep=1pt, font=\normalfont]
    \item[$\bullet$ Step Count Calculation] \hfill \\
        Inputs: 120 sec IMU signal recorded during walking $\mid$ Output: Step count during walking \\
        Sequence Length: 10K-100K $\mid$ Frequency: 100 Hz \\
        Metric: Accuracy
\end{description}

\textbf{Inference - Physiological Classification}
\begin{description}[itemsep=0pt, topsep=1pt, font=\normalfont]
    \item[$\bullet$ Activity Classification] \hfill \\
        Inputs: 150 sec 3-axis signal $\mid$ Output: Activity type \\
        Sequence Length: 10K-100K $\mid$ Frequency: 100 Hz \\
        Metric: Accuracy
    
    \item[$\bullet$ Activity Transition Recognition] \hfill \\
        Inputs: 150 sec 3-axis signal $\mid$ Output: Activity transition sequence \\
        Sequence Length: 10K-100K $\mid$ Frequency: 100 Hz \\
        Metric: Accuracy
\end{description}

\textbf{Generation - Signal Imputation}
\begin{description}[itemsep=0pt, topsep=1pt, font=\normalfont]
    \item[$\bullet$ 1-axis Signal Imputation] \hfill \\
        Inputs: 150 sec 3-axis IMU signal with 30 sec missing on z axis $\mid$ Output: 30 sec missing-axis signal \\
        Sequence Length: 10K-100K $\mid$ Frequency: 100 Hz \\
        Metric: sMAPE score
    
    \item[$\bullet$ 3-axis Signal Imputation] \hfill \\
        Inputs: 150 sec 3-axis IMU signal with 30 sec all three axis missing $\mid$ Output: 30 sec missing 3-axis signal \\
        Sequence Length: 10K-100K $\mid$ Frequency: 100 Hz \\
        Metric: sMAPE score
\end{description}

\textbf{Generation - Future Forecasting}
\begin{description}[itemsep=0pt, topsep=1pt, font=\normalfont]
    \item[$\bullet$ 3-axis Signal Forecasting] \hfill \\
        Inputs: 120 sec 3-axis IMU signal $\mid$ Output: next 30 sec signal \\
        Sequence Length: 10K-100K $\mid$ Frequency: 100 Hz \\
        Metric: sMAPE score
\end{description}

\textbf{Deduction - Temporal Ordering}
\begin{description}[itemsep=0pt, topsep=1pt, font=\normalfont]
    \item[$\bullet$ Day and Night Signal Ordering] \hfill \\
        Inputs: 3 hour night data \& 3 hour day data $\mid$ Output: Which one is night data? \\
        Sequence Length: $>$1M $\mid$ Frequency: 100 Hz \\
        Metric: Accuracy
\end{description}

\subsubsection{PAMAP2}
\textbf{Inference - Event Localization}
\begin{description}[itemsep=0pt, topsep=1pt, font=\normalfont]
    \item[$\bullet$ Activity Localization] \hfill \\
        Inputs: 5x activity length 3-axis IMU signal $\mid$ Output: Activity's start time and end time \\
        Sequence Length: 100K-1M $\mid$ Frequency: 100 Hz \\
        Metric: IoU
\end{description}

\subsubsection{Shanghai Diabetes}
\textbf{Inference - Subject Profiling}
\begin{description}[itemsep=0pt, topsep=1pt, font=\normalfont]
    \item[$\bullet$ Cross-Subject Diabetes Type Comparison] \hfill \\
        Inputs: Full length CGM from two different subjects $\mid$ Output: Which subject has type 1 diabetes and which has type 2 diabetes \\
        Sequence Length: 1K-10K $\mid$ Frequency: Per minute \\
        Metric: Accuracy
    
    \item[$\bullet$ Diabetes Type Classification] \hfill \\
        Inputs: Full length CGM data from a diabetic subject $\mid$ Output: This subject has type 1 diabetes or type 2 diabetes \\
        Sequence Length: 1K-10K $\mid$ Frequency: Per minute \\
        Metric: Accuracy
\end{description}

\subsubsection{CGMacros}
\textbf{Perception - Statistical Calculation}
\begin{description}[itemsep=0pt, topsep=1pt, font=\normalfont]
     \item[$\bullet$ CGM Time in Range Calculation] \hfill \\
        Inputs: entire CGM \& normal CGM range \\
        Output: percentage of time below normal CGM range \& percentage of time above normal CGMrange \\
        Sequence Length: 1K-10K $\mid$ Frequency: per minute \\
        Metric: sMAPE score
\end{description}

\textbf{Perception - Feature Extraction}
\begin{description}[itemsep=0pt, topsep=1pt, font=\normalfont]
    \item[$\bullet$ Postprandial CGM iAUC Calculation] \hfill \\
        Inputs: 2hr CGM after meal starts $\mid$ Output: iAUC value \\
        Sequence Length: $<$100 $\mid$ Frequency: per minute \\
        Metric: sMAPE score
\end{description}

\textbf{Inference - Event Localization}
\begin{description}[itemsep=0pt, topsep=1pt, font=\normalfont]
    \item[$\bullet$ Meal Time Localization] \hfill \\
        Inputs: 2hr CGM window $\mid$ Output: meal start timestamp \\
        Sequence Length: $<$100 $\mid$ Frequency: per minute \\
        Metric: sMAPE score
\end{description}

\textbf{Inference - Physiological Classification}
\begin{description}[itemsep=0pt, topsep=1pt, font=\normalfont]
    \item[$\bullet$ Meal Image Classification from CGM] \hfill \\
        Inputs: 4hr CGM time window with meal event \& 4 options of meal image \\
        Output: identify the image corresponding to the subject’s meal \\
        Sequence Length: $<$100 $\mid$ Frequency: per minute \\
        Metric: Accuracy
\end{description}

\textbf{Inference - Subject Profiling}
\begin{description}[itemsep=0pt, topsep=1pt, font=\normalfont]
    \item[$\bullet$ A1c Prediction] \hfill \\
        Inputs: Full length CGM $\mid$ Output: Select A1c range (3 options) \\
        Sequence Length: 1K-10K $\mid$ Frequency: Per minute \\
        Metric: Accuracy
    
    \item[$\bullet$ Fasting GLU Prediction] \hfill \\
        Inputs: Full length CGM $\mid$ Output: Predict fasting GLU \\
        Sequence Length: 1K-10K $\mid$ Frequency: Per minute \\
        Metric: sMAPE score
    
    \item[$\bullet$ Postprandial CGM Response Comparison] \hfill \\
        Inputs: 4-hour CGM data following similar meals from two different subjects (carbohydrates and calories differ by $<$ 10\%)\\
        Output: Which subject exhibits normal glucose regulation \\
        Sequence Length: $<$100 $\mid$ Frequency: Per minute \\
        Metric: Accuracy
\end{description}

\textbf{Generation - Signal Imputation}
\begin{description}[itemsep=0pt, topsep=1pt, font=\normalfont]
    \item[$\bullet$ CGM Imputation] \hfill \\
        Inputs: 2hr CGM window with 30 minute missing $\mid$ Output: Missing 30 min CGM \\
        Sequence Length: $<$100 $\mid$ Frequency: Per minute \\
        Metric: sMAPE score
    
    \item[$\bullet$ CGM Imputation with Activity Calories Trajectory] \hfill \\
        Inputs: 2hr CGM window with 30 minute missing \& activity calories info within that window \\
        Output: Missing 30 min CGM \\
        Sequence Length: $<$100 $\mid$ Frequency: Per minute \\
        Metric: sMAPE score
    
    \item[$\bullet$ CGM Imputation with HR Trajectory] \hfill \\
        Inputs: 2hr CGM window with 30 minute missing \& HR within that window $\mid$ Output: Missing 30 min CGM \\
        Sequence Length: $<$100 $\mid$ Frequency: Per minute \\
        Metric: sMAPE score
\end{description}

\textbf{Generation - Future Forecasting}
\begin{description}[itemsep=0pt, topsep=1pt, font=\normalfont]
    \item[$\bullet$ CGM Forecasting] \hfill \\
        Inputs: 1hr CGM before meal $\mid$ Output: Next 30 min CGM \\
        Sequence Length: $<$100 $\mid$ Frequency: Per minute \\
        Metric: sMAPE score

    \item[$\bullet$ CGM Forecasting with Meal Information] \hfill \\
        Inputs: 1hr CGM before meal \& meal info $\mid$ Output: Next 30 min CGM \\
        Sequence Length: $<$100 $\mid$ Frequency: Per minute \\
        Metric: sMAPE score
        
    \item[$\bullet$ CGM Forecasting with History CGM and Meal Info] \hfill \\
        Inputs: Previous 3 days' CGM \& meal of same subject as reference, and 1hr CGM before meal\\
        Output: Next 30 min CGM \\
        Sequence Length: 100-1K $\mid$ Frequency: Per minute \\
        Metric: sMAPE score
    
    \item[$\bullet$ CGM Forecasting with Current Meal Information and History CGM] \hfill \\
        Inputs: Previous 3 days' CGM \& meal of same subject as reference, and 1hr CGM before meal \& meal info \\
        Output: Next 30 min CGM \\
        Sequence Length: 100-1K $\mid$ Frequency: Per minute \\
        Metric: sMAPE score
\end{description}

\subsubsection{VitalDB}
\textbf{Perception - Statistical Calculation}
\begin{description}[itemsep=0pt, topsep=1pt, font=\normalfont]
\item[$\bullet$ Mean arterial pressure (MBP) Time in Range 70-100] \hfill \\
    Inputs: Full length blood Pressure (Solar8000/ART\_MBP) $\mid$ Output: Percentage of time where blood pressure within normal range \\
    Sequence Length: 100K-1M $\mid$ Frequency: 0.5 Hz \\
    Metric: sMAPE score
\end{description}

\textbf{Inference - Event Localization}
\begin{description}[itemsep=0pt, topsep=1pt, font=\normalfont]
    \item[$\bullet$ Anesthesia Range Localization] \hfill \\
        Inputs: Full length annotation (BIS/EEG1\_WAV \& BIS/EEG2\_WAV) $\mid$ Output: Anesthesia time range [start time, end time] \\
        Sequence Length: 100K-1M $\mid$ Frequency: 128 Hz \\
        Metric: IoU
\end{description}

\textbf{Inference - Physiological Classification}
\begin{description}[itemsep=0pt, topsep=1pt, font=\normalfont]
    \item[$\bullet$ Hypotension Event Classification with MBP] \hfill \\
        Inputs: 600 sec blood pressure (Solar8000/ART\_MBP) $\mid$ Output: If next 5 minutes will have hypotension events \\
        Sequence Length: $<$100 $\mid$ Frequency: 0.5 Hz \\
        Metric: Accuracy

    \item[$\bullet$ Hypotension Event Classification with PPG] \hfill \\
        Inputs: 600 sec PPG (SNUADC/PLETH) $\mid$ Output: If next 5 minutes will have hypotension events \\
        Sequence Length: 10K-100K $\mid$ Frequency: 100 Hz \\
        Metric: Accuracy
\end{description}

\textbf{Inference - Subject Profiling}
\begin{description}[itemsep=0pt, topsep=1pt, font=\normalfont]
    \item[$\bullet$ MINS (myocardial injury after non-cardiac surgery) Prediction] \hfill \\
        Inputs: Full length Blood Pressure (Solar8000/ART\_MBP) $\mid$ Output: White blood cell count from lab result \\
        Sequence Length: 100-1K $\mid$ Frequency: 1 Hz \\
        Metric: Accuracy
\end{description}

\textbf{Generation - Future Forecasting}
\begin{description}[itemsep=0pt, topsep=1pt, font=\normalfont]
    \item[$\bullet$ MBP Forecasting] \hfill \\
        Inputs: 120 sec blood pressure (Solar8000/ART\_MBP) $\mid$ Output: Next 30 sec Blood Pressure \\
        Sequence Length: $<$100 $\mid$ Frequency: 0.5 Hz \\
        Metric: sMAPE score
    
    \item[$\bullet$ MBP Forecasting with Infusion Info] \hfill \\
        Inputs: 120 sec blood pressure (Solar8000/ART\_MBP) + All infusion data at this time window $\mid$ Output: 30s MBP data \\
        Sequence Length: $<$100 $\mid$ Frequency: 0.5 Hz \\
        Metric: sMAPE score
\end{description}

\textbf{Generation - Cross-modal Translation}
\begin{description}[itemsep=0pt, topsep=1pt, font=\normalfont]
    \item[$\bullet$ Drug Infusion Series to Depth of Anesthesia (BIS) Translation] \hfill \\
        Inputs: Drug infusion series (Orchestra/PPF20\_VOL \& Orchestra/RFTN20\_VOL \& Orchestra/PPF20\_CE \& Orchestra/RFTN20\_CE) \\
        Output: Corresponding BIS series \\
        Sequence Length: 100-1K $\mid$ Frequency: 10s intervals \\
        Metric: sMAPE score
    
    \item[$\bullet$ EEG to Bispectral Index (BIS) Translation] \hfill \\
        Inputs: 360 sec EEG (BIS/EEG1\_WAV \& BIS/EEG2\_WAV) $\mid$ Output: Corresponding BIS series \\
        Sequence Length: 100K-1M $\mid$ Frequency: 128 Hz \\
        Metric: sMAPE score
    
    \item[$\bullet$ PPG and ECG to Blood Pressure Translation] \hfill \\
        Inputs: 360 sec ECG (SNUADC/ECG\_II) + PPG (SNUADC/PLETH) \\
        Output: Corresponding blood pressure series \\
        Sequence Length: 100-1K $\mid$ Frequency: 100 Hz \\
        Metric: sMAPE score
\end{description}

\subsubsection{SHHS}

\textbf{Perception - Statistical Calculation}
\begin{description}[itemsep=0pt, topsep=1pt, font=\normalfont]
    \item[$\bullet$ REM Latency Calculation] \hfill \\
    Inputs: Whole night sleepstage annotations $\mid$ Output: REM latency value \\
    Sequence Length: 100-1K $\mid$ Frequency: Per 30sec epoch \\
    Metric: sMAPE score
    
    \item[$\bullet$ Sleep AHI Calculation] \hfill \\
    Inputs: Whole night sleepstage annotations and sleep event calculations $\mid$ Output: Sleep AHI value \\
    Sequence Length: 100-1K $\mid$ Frequency: Per 30sec epoch \\
    Metric: sMAPE score
\end{description}

\textbf{Perception - Feature Extraction}
\begin{description}[itemsep=0pt, topsep=1pt, font=\normalfont]
    \item[$\bullet$ Bandpower Calculation] \hfill \\
        Inputs: EEG C3/A2 in various length $\mid$ Output: Bandpower values \\
        Sequence Length: 100K-1M $\mid$ Frequency: 64 Hz \\
        Metric: sMAPE score

    \item[$\bullet$ Sleep Efficiency Calculation] \hfill \\
        Inputs: Whole night sleepstage annotations $\mid$ Output: Sleep efficiency value \\
        Sequence Length: 100-1K $\mid$ Frequency: 30s epochs \\
        Metric: sMAPE score
\end{description}

\textbf{Inference - Event Localization}
\begin{description}[itemsep=0pt, topsep=1pt, font=\normalfont]
    \item[$\bullet$ Hypopnea Detection] \hfill \\
        Inputs: Airflow \& Thoracic with 5x event duration $\mid$ Output: List of [start time, end time] \\
        Sequence Length: 1K-10K $\mid$ Frequency: 8 Hz \\
        Metric: IoU

    \item[$\bullet$ Arousal Detection (EEG)] \hfill \\
        Inputs: EEG C3/A2 with 5x event duration $\mid$ Output: List of [start time, end time] \\
        Sequence Length: 1K-10K $\mid$ Frequency: 64 Hz \\
        Metric: IoU
    
    \item[$\bullet$ Arousal Detection (EOG)] \hfill \\
        Inputs: EOG-E2A1 \& EOG-E1A2 with 5x event duration $\mid$ Output: List of [start time, end time] \\
        Sequence Length: 1K-10K $\mid$ Frequency: 64 Hz \\
        Metric: IoU 
\end{description}

\textbf{Inference - Physiological Classification}
\begin{description}[itemsep=0pt, topsep=1pt, font=\normalfont]
    \item[$\bullet$ REM/NREM Classification] \hfill \\
        Inputs: EOG-E2A1 \& EOG-E1A2 within REM/NREM sleep $\mid$ Output: REM or NREM \\
        Sequence Length: 100K-1M $\mid$ Frequency: 64 Hz \\
        Metric: Accuracy
    
    \item[$\bullet$ Sleep Stage Classification] \hfill \\
        Inputs: EEG C3/A2 of a specific sleep stage \& reference bandpower of different sleep stages of same subject \\
        Output: Sleep stage \\
        Sequence Length: 1K-10K $\mid$ Frequency: 64 Hz \\
        Metric: Accuracy
    
    \item[$\bullet$ Sleep Stage Transition Recognition] \hfill \\
        Inputs: EEG C3/A2 from two consecutive sleep stages \& bandpower of different sleep stages of same subject \\
        Output: Sleep stage transition sequence (e.g., N1$\to$N2) \\
        Sequence Length: 1K-10K $\mid$ Frequency: 64 Hz \\
        Metric: Accuracy 
\end{description}

\textbf{Inference - Subject Profiling}
\begin{description}[itemsep=0pt, topsep=1pt, font=\normalfont]
     \item[$\bullet$ Visit Level Ordering] \hfill \\
        Inputs: ECG \& Airflow \& EEG C3/A2 from different visit $\mid$ Output: Which one is ahead of another \\
        Sequence Length: $>$1M $\mid$ Frequency: 128 Hz \\
        Metric: Accuracy
    
    \item[$\bullet$ Half Night Level Ordering] \hfill \\
        Inputs: ECG \& Airflow \& EEG C3/A2 from first half / second half of the whole night signal \\
        Output: Which one is ahead of another \\
        Sequence Length: $>$1M $\mid$ Frequency: 128 Hz \\
        Metric: Accuracy
    
    \item[$\bullet$ Episode Level Ordering] \hfill \\
        Inputs: ECG \& Airflow \& EEG C3/A2 from different neighboring episodes \\
        Output: Which one is ahead of another \\
        Sequence Length: 1K-10K $\mid$ Frequency: 128 Hz \\
        Metric: Accuracy
\end{description}

\textbf{Generation - Signal Imputation}
\begin{description}[itemsep=0pt, topsep=1pt, font=\normalfont]
     \item[$\bullet$ Single-channel Imputation] \hfill \\
        Inputs: 150s ECG signal (missing interval: 60-90s) $\mid$ Output: Missing ECG signal \\
        Sequence Length: 10K-100K $\mid$ Frequency: 128 Hz \\
        Metric: sMAPE score
    
    \item[$\bullet$ Conditional Imputation] \hfill \\
        Inputs: 150 sec ECG channel (missing interval: 60-90s) \& 150 sec Airflow without missing \\
        Output: Missing ECG signal \\
        Sequence Length: 10K-100K $\mid$ Frequency: 128 Hz \\
        Metric: sMAPE score
\end{description}

\textbf{Generation - Future Forecasting}
\begin{description}[itemsep=0pt, topsep=1pt, font=\normalfont]
     \item[$\bullet$ Single-channel Forecasting] \hfill \\
        Inputs: 120 sec ECG channel $\mid$ Output: Next 30sec ECG signal \\
        Sequence Length: 10K-100K $\mid$ Frequency: 128 Hz \\
        Metric: sMAPE score
    
    \item[$\bullet$ Conditional Forecasting] \hfill \\
        Inputs: 120 sec ECG channel \& 120 sec Airflow signal $\mid$ Output: Next 30sec ECG signal \\
        Sequence Length: 10K-100K $\mid$ Frequency: 128 Hz \\
        Metric: sMAPE score
\end{description}

\textbf{Generation - Cross-modal Translation}
\begin{description}[itemsep=0pt, topsep=1pt, font=\normalfont]
     \item[$\bullet$ Cross-channel Translation (EEG C3/A2 to EEG C4/A1)] \hfill \\
        Inputs: 120 sec EEG C3/A2 channel $\mid$ Output: Corresponding 120 sec EEG C4/A1 channel \\
        Sequence Length: 1K-10K $\mid$ Frequency: 64 Hz \\
        Metric: sMAPE score
    
    \item[$\bullet$ Cross-channel Translation (ECG to HR)] \hfill \\
        Inputs: 120 sec ECG $\mid$ Output: Corresponding 120 sec HR \\
        Sequence Length: 1K-10K $\mid$ Frequency: 128 Hz \\
        Metric: sMAPE score
\end{description}

\textbf{Deduction - Temporal Ordering}
\begin{description}[itemsep=0pt, topsep=1pt, font=\normalfont]
     \item[$\bullet$ Smoker Classification] \hfill \\
        Inputs: Whole night Airflow \& ECG \& EEG C3/A2 $\mid$ Output: If this patient is a smoker (Yes/No) \\
        Sequence Length: $>$1M $\mid$ Frequency: 128 Hz \\
        Metric: Accuracy
    
    \item[$\bullet$ Atrial Fibrillation (AF) Classification] \hfill \\
        Inputs: Whole night ECG $\mid$ Output: If this patient has AF (Yes/No) \\
        Sequence Length: $>$1M $\mid$ Frequency: 128 Hz \\
        Metric: Accuracy
    
    \item[$\bullet$ Cardiovascular Disease (CVD) Death Prediction] \hfill \\
        Inputs: Whole night ECG \& Airflow \& EEG C3/A2 $\mid$ Output: If this patient will have fatal CVD (Yes/No) \\
        Sequence Length: $>$1M $\mid$ Frequency: 128 Hz \\
        Metric: Accuracy
    
    \item[$\bullet$ Stroke Prediction] \hfill \\
        Inputs: Whole night ECG \& Airflow \& EEG C3/A2 $\mid$ Output: If this patient will have stroke (Yes/No) \\
        Sequence Length: $>$1M $\mid$ Frequency: 128 Hz \\
        Metric: Accuracy
\end{description}

\textbf{Deduction - Trajectory Analysis}
\begin{description}[itemsep=0pt, topsep=1pt, font=\normalfont]
     \item[$\bullet$ BMI Comparison Between Visit] \hfill \\
        Inputs: Whole night ECG \& Airflow \& EEG C3/A2 from 2 visits of same subject \\
        Output: Which visit has higher BMI? (visit1 / visit2) \\
        Sequence Length: $>$1M $\mid$ Frequency: 128 Hz \\
        Metric: Accuracy
\end{description}

\subsubsection{Harespod}
\textbf{Inference - Physiological Classification}
\begin{description}[itemsep=0pt, topsep=1pt, font=\normalfont]
    \item[$\bullet$ Altitude Ranking with Respiration] \hfill \\
        Inputs: Given 3 segments of 5 min respiration signals, ranking altitude $\mid$ Output: Arrange from low to high \\
        Sequence Length: 10K-100K $\mid$ Frequency: 100 Hz \\
        Metric: Accuracy
    
    \item[$\bullet$ Altitude Ranking with SpO2] \hfill \\
        Inputs: Given 3 segments of 5 min SpO2 signals, ranking altitude $\mid$ Output: Arrange from low to high \\
        Sequence Length: 10K-100K $\mid$ Frequency: 100 Hz \\
        Metric: Accuracy
\end{description}

\textbf{Inference - Subject Profiling}
\begin{description}[itemsep=0pt, topsep=1pt, font=\normalfont]
    \item[$\bullet$ Respiration and SpO2 Pairing] \hfill \\
        Inputs: Given 2 sets of 10min SpO2 and respiration signal \\
        Output: Pairing corresponding SpO2 signal and respiration signal \\
        Sequence Length: 10K-100K $\mid$ Frequency: 100 Hz \\
        Metric: Accuracy
    
    \item[$\bullet$ Respiration and HR Pairing] \hfill \\
        Inputs: Given 2 sets 10min HR and respiration signal \\
        Output: Pairing corresponding respiration signal and heart rate signal \\
        Sequence Length: 10K-100K $\mid$ Frequency: 100 Hz \\
        Metric: Accuracy
\end{description}

\subsubsection{PhyMER}
\textbf{Inference - Physiological Classification}
\begin{description}[itemsep=0pt, topsep=1pt, font=\normalfont]
    \item[$\bullet$ Emotion Type Classification] \hfill \\
        Inputs: All E4 signals \& self-reported arousal and valence value $\mid$ Output: Emotion Type \\
        Sequence Length: 100K-1M $\mid$ Frequency: 64 Hz \\
        Metric: Accuracy

    \item[$\bullet$ Emotion Type Classification with Only EDA Input] \hfill \\
        Inputs: EDA \& self-reported arousal and valence value $\mid$ Output: Emotion Type \\
        Sequence Length: 1K-10K $\mid$ Frequency: 4 Hz \\
        Metric: Accuracy
\end{description}

\textbf{Inference - Subject Profiling}
\begin{description}[itemsep=0pt, topsep=1pt, font=\normalfont]
    \item[$\bullet$ Cross Subject Arousal Ranking] \hfill \\
        Inputs: All E4 signals of two subjects watching same video \& emotion label of each subject \\
        Output: Which subject has higher arousal? \\
        Sequence Length: 100K-1M $\mid$ Frequency: 64 Hz \\
        Metric: Accuracy
    
    \item[$\bullet$ Inter-subject Emotion Recognition] \hfill \\
        Inputs: All E4 signals of one subject watching different video \& emotion type of the two sessions \\
        Output: Identify correct emotion (e.g., which one's emotion is angry) \\
        Sequence Length: 100K-1M $\mid$ Frequency: 64 Hz \\
        Metric: Accuracy
    
    \item[$\bullet$ Personality Analysis] \hfill \\
        Inputs: All video labels for one subject $\mid$ Output: Big-5 Personality \\
        Sequence Length: 100K-1M $\mid$ Frequency: 64 Hz \\
        Metric: Accuracy
\end{description}

\textbf{Generation - Signal Imputation}
\begin{description}[itemsep=0pt, topsep=1pt, font=\normalfont]
    \item[$\bullet$ Single-channel Imputation] \hfill \\
        Inputs: HR with 20 sec missing $\mid$ Output: 20sec missing HR \\
        Sequence Length: $<$100 $\mid$ Frequency: 1 Hz \\
        Metric: sMAPE score
    
    \item[$\bullet$ Conditional Imputation] \hfill \\
        Inputs: HR with 20 sec missing \& corresponding BVP, EDA, TEMP in that time range \& emotion labels \\
        Output: 20 sec missing HR \\
        Sequence Length: 1K-10K $\mid$ Frequency: 64 Hz \\
        Metric: sMAPE score
\end{description}

\textbf{Generation - Future Forecasting}
\begin{description}[itemsep=0pt, topsep=1pt, font=\normalfont]
    \item[$\bullet$ Single-channel Forecasting] \hfill \\
        Inputs: HR $\mid$ Output: Next 20 sec HR \\
        Sequence Length: $<$100 $\mid$ Frequency: 1 Hz \\
        Metric: sMAPE score
    
    \item[$\bullet$ Conditional Forecasting] \hfill \\
        Inputs: All E4 signals \& emotion labels $\mid$ Output: Next 20 sec HR \\
        Sequence Length: 1K-10K $\mid$ Frequency: 1 Hz \\
        Metric: sMAPE score
\end{description}

\textbf{Generation - Cross-modal Translation}
\begin{description}[itemsep=0pt, topsep=1pt, font=\normalfont]
    \item[$\bullet$ Cross-channel Translation] \hfill \\
        Inputs: Full length BVP \& EDA \& TEMP \& Emotion $\mid$ Output: Corresponding HR \\
        Sequence Length: 1K-10K $\mid$ Frequency: 64 Hz \\
        Metric: sMAPE score
\end{description}

\subsubsection{PERG-IOBA}
\textbf{Perception - Feature Extraction}
\begin{description}[itemsep=0pt, topsep=1pt, font=\normalfont]
    \item[$\bullet$ N35, P50, N95 Feature Extraction] \hfill \\
        Inputs: 1 PERG signal recording $\mid$ Output: N35, P50 and N95 amplitude \\
        Sequence Length: 100-1K $\mid$ Frequency: 1700 Hz \\
        Metric: sMAPE score
\end{description}

\textbf{Inference - Subject Profiling}
\begin{description}[itemsep=0pt, topsep=1pt, font=\normalfont]
    \item[$\bullet$ Eye Health Status Classification] \hfill \\
        Inputs: 1 PERG signal recording $\mid$ Output: whether the subject had eye disease \\
        Sequence Length: 100-1K $\mid$ Frequency: 1700 Hz \\
        Metric: Accuracy
    \item[$\bullet$ Eye Disease Type Classification] \hfill \\
        Inputs: 1 PERG signal recording $\mid$ Output: Out of 4 choices, which is most likely the disease that the subject had \\
        Sequence Length: 100-1K $\mid$ Frequency: 1700 Hz \\
        Metric: Accuracy
    \item[$\bullet$ Eye Disease Type Classification with Patient's Meta Information] \hfill \\
        Inputs: 1 PERG signal recording and subject's age, visual acuity and sex information $\mid$ Output: Out of 4 choices, which is most likely the disease that the subject had \\
        Sequence Length: 100-1K $\mid$ Frequency: 1700 Hz \\
        Metric: Accuracy
    \item[$\bullet$ Disease Differentiation: between Macular Disease \& Optic Nerve Disease \& Normal] \hfill \\
        Inputs: 1 PERG signal recording $\mid$ Output: Differentiate 2 types of disease and normal status \\
        Sequence Length: 100-1K $\mid$ Frequency: 1700 Hz \\
        Metric: Accuracy
\end{description}

\subsubsection{GazeBase}
\textbf{Inference - Event Localization}
\begin{description}[itemsep=0pt, topsep=1pt, font=\normalfont]
    \item[$\bullet$ Fixation Point Localization] \hfill \\
        Inputs: full length eye tracking data from fixation task $\mid$ Output: Coordinate of the fixation point \\
        Sequence Length: 10K-100K $\mid$ Frequency: 1000 Hz \\
        Metric: sMAPE score
\end{description}

\textbf{Inference - Physiological Classification}
\begin{description}[itemsep=0pt, topsep=1pt, font=\normalfont]
    \item[$\bullet$ Eye-tracking Task Classification] \hfill \\
        Inputs: 20 sec eye tracking data segment $\mid$ Output: Task type \\
        Sequence Length: 10K-100K $\mid$ Frequency: 1000 Hz \\
        Metric: Accuracy
\end{description}

\textbf{Generation - Signal Imputation}
\begin{description}[itemsep=0pt, topsep=1pt, font=\normalfont]
    \item[$\bullet$ Text Reading Imputation] \hfill \\
        Inputs: 50 sec eye tracking sequence from reading task with 10 sec missing $\mid$ Output: Missing 10 sec data \\
        Sequence Length: 10K-100K $\mid$ Frequency: 1000 Hz \\
        Metric: sMAPE score
\end{description}

\textbf{Generation - Future Forecasting}
\begin{description}[itemsep=0pt, topsep=1pt, font=\normalfont]
    \item[$\bullet$ Horizontal Saccade Track Forecasting] \hfill \\
        Inputs: 90 sec eye tracking sequence $\mid$ Output: Next 10sec data \\
        Sequence Length: 10K-100K $\mid$ Frequency: 1000 Hz \\
        Metric: sMAPE score
\end{description}

\textbf{Deduction - Temporal Ordering}
\begin{description}[itemsep=0pt, topsep=1pt, font=\normalfont]
    \item[$\bullet$ Reading Sequence Recognition] \hfill \\
        Inputs: First \& middle \& last 10 sec of one eye tracking sequence from reading task $\mid$ Output: Sequence of these 10sec clips (e.g. A$\to$B$\to$C) \\
        Sequence Length: 10K-100K $\mid$ Frequency: 1000 Hz \\
        Metric: Accuracy
\end{description}

\subsubsection{GLOBEM}
\textbf{Perception - Feature Extraction}
\begin{description}[itemsep=0pt, topsep=1pt, font=\normalfont]
    \item[$\bullet$ Location Entropy Extraction] \hfill \\
        Inputs: 2 subject's 7-days location log $\mid$ Output: Which one has higher significant location entropy? \\
        Sequence Length: $<$100 $\mid$ Frequency: Daily aggregates \\
        Metric: Accuracy
\end{description}

\textbf{Inference - Event Localization}
\begin{description}[itemsep=0pt, topsep=1pt, font=\normalfont]
    \item[$\bullet$ Peak Stress Week Localization] \hfill \\
        Inputs: 10 weeks sleep \& location \& screen\& step data $\mid$ Output: Which week is the 'peak stress week'? \\
        Sequence Length: $<$100 $\mid$ Frequency: Daily aggregates \\
        Metric: Accuracy
\end{description}

\textbf{Inference - Physiological Classification}
\begin{description}[itemsep=0pt, topsep=1pt, font=\normalfont]
    \item[$\bullet$ Depression Trajectory Classification] \hfill \\
        Inputs: 10 weeks data $\mid$ Output: Whether the subject has depression at the end of term? \\
        Sequence Length: $<$100 $\mid$ Frequency: Daily aggregates \\
        Metric: Accuracy
    
    \item[$\bullet$ COVID Year Recognition] \hfill \\
        Inputs: General behavior patterns (travel distance, step count) of different years \\ Output: which one is from post COVID? \\
        Sequence Length: $<$100 $\mid$ Frequency: Daily aggregates \\
        Metric: Accuracy
\end{description}

\textbf{Generation - Future Forecasting}
\begin{description}[itemsep=0pt, topsep=1pt, font=\normalfont]
    \item[$\bullet$ Step Count Forecasting] \hfill \\
        Inputs: Given past 14 days step count $\mid$ Output: Next 1day's step count \\
        Sequence Length: $<$100 $\mid$ Frequency: Daily aggregates \\
        Metric: sMAPE score
\end{description}

\textbf{Inference - Subject Profiling}
\begin{description}[itemsep=0pt, topsep=1pt, font=\normalfont]
    \item[$\bullet$ Circadian Routine Comparison] \hfill \\
        Inputs: 2 subject's ten week sleep log (get up / go to sleep time), location log and step count\\
        Output: Which one has higher circadian routine score (circdnrtn)? \\
        Sequence Length: $<$100 $\mid$ Frequency: Daily aggregates \\
        Metric: Accuracy
\end{description}

\subsubsection{Bridge2AI-voice}
\textbf{Perception - Feature Extraction}
\begin{description}[itemsep=0pt, topsep=1pt, font=\normalfont]
    \item[$\bullet$ F0 Range Extraction] \hfill \\
        Inputs: 1 spectrogram of a recording $\mid$ Output: Value of mean F0 (fundamental frequency) \\
        Sequence Length: 10K - 100K $\mid$ Frequency: 100 Hz \\
        Metric: sMAPE score
    
    \item[$\bullet$ Harmonics to Noise Ratio (HNR) Extraction] \hfill \\
        Inputs: 1 spectrogram of a recording $\mid$ Output: Value of mean HNR \\
        Sequence Length: 10K - 100K $\mid$ Frequency: 100 Hz \\
        Metric: sMAPE score

    \item[$\bullet$ Shimmer Extraction] \hfill \\
        Inputs: 1 spectrogram of a recording $\mid$ Output: Value of mean shimmer \\
        Sequence Length: 10K - 100K $\mid$ Frequency: 100 Hz \\
        Metric: sMAPE score
    
    \item[$\bullet$ Jitter Extraction] \hfill \\
        Inputs: 1 spectrogram of a recording $\mid$ Output: Value of mean jitter \\
        Sequence Length: 10K - 100K $\mid$ Frequency: 100 Hz \\
        Metric: sMAPE score
\end{description}

\textbf{Perception - Feature Extraction}
\begin{description}[itemsep=0pt, topsep=1pt, font=\normalfont]
    \item[$\bullet$ Articulation Rate Calculation] \hfill \\
        Inputs: 1 spectrogram of a recording $\mid$ Output: Articulation rate \\
        Sequence Length: 10K - 100K $\mid$ Frequency: 100 Hz \\
        Metric: sMAPE score
\end{description}

\textbf{Inference - Physiological Classification}
\begin{description}[itemsep=0pt, topsep=1pt, font=\normalfont]
    \item[$\bullet$ Cross-task Voice Comparison] \hfill \\
        Inputs: Spectrograms of 2 recordings $\mid$ Output: Matching transcripts with the spectrogram \\
        Sequence Length: 10K - 100K $\mid$ Frequency: 100 Hz \\
        Metric: Accuracy
\end{description}

\textbf{Inference - Subject Profiling}
\begin{description}[itemsep=0pt, topsep=1pt, font=\normalfont]
    \item[$\bullet$ Parkinson's Diagnosis] \hfill \\
        Inputs: 1 spectrogram of a recording $\mid$ Output: Whether the subject has Parkinson's disease \\
        Sequence Length: 10K - 100K $\mid$ Frequency: 100 Hz \\
        Metric: Accuracy
\end{description}

\textbf{Deduction - Temporal Ordering}
\begin{description}[itemsep=0pt, topsep=1pt, font=\normalfont]
    \item[$\bullet$ Reversed Signal Detection] \hfill \\
        Inputs: 1 spectrogram of a recording $\mid$ Output: Whether the spectrogram was reversed \\
        Sequence Length: 10K - 100K $\mid$ Frequency: 100 Hz \\
        Metric: Accuracy
\end{description}

\subsubsection{VCTK}

\textbf{Deduction - Temporal Ordering}
\begin{description}[itemsep=0pt, topsep=1pt, font=\normalfont]
    \item[$\bullet$ Reversed Signal Detection] \hfill \\
        Inputs: 1s audio waveform $\mid$ Output: Whether the waveform was reversed \\
        Sequence Length: 100K - 1M $\mid$ Frequency: 48000 Hz \\
        Metric: Accuracy
\end{description}

\subsubsection{GrabMyo}

\textbf{Inference - Physiological Classification}
\begin{description}[itemsep=0pt, topsep=1pt, font=\normalfont]
    \item[$\bullet$ Gesture Classification with Reference] \hfill \\
        Inputs: 28 channels of EMG signal and reference EMG signals of all gestures $\mid$ Output: Gesture prediction \\
        Sequence Length: 1k - 10K $\mid$ Frequency: 2048 Hz \\
        Metric: Accuracy
    \item[$\bullet$ Subject Identification] \hfill \\
        Inputs: 28 channels of EMG signal and reference EMG signals of same gesture from a subject recorded in different session $\mid$ Output: Whether EMG signal was from the same subject \\
        Sequence Length: 1k - 10K $\mid$ Frequency: 2048 Hz \\
        Metric: Accuracy
\end{description}

\subsubsection{CoughVID}

\textbf{Perception - Feature Extraction}
\begin{description}[itemsep=0pt, topsep=1pt, font=\normalfont]
    \item[$\bullet$ MFCC Mean \& STD Calculation] \hfill \\
        Inputs: 1-12 seconds audio signal $\mid$ Output: Mean and standard deviation of MFCC values \\
        Sequence Length: 100K-1M $\mid$ Frequency: 48K Hz \\
        Metric: sMAPE score
\end{description}

\textbf{Inference - Physiological Classification}
\begin{description}[itemsep=0pt, topsep=1pt, font=\normalfont]
    \item[$\bullet$ Health Status Classification] \hfill \\
        Inputs: 1-12 seconds audio signal $\mid$ Output: Whether the subject is healthy \\
        Sequence Length: 100K-1M $\mid$ Frequency: 48K Hz \\
        Metric: Accuracy
    \item[$\bullet$ COVID Status Classification] \hfill \\
        Inputs: 1-12 seconds audio signal $\mid$ Output: COVID-19 positive/negative prediction \\
        Sequence Length: 100K-1M $\mid$ Frequency: 48K Hz \\
        Metric: Accuracy
    \item[$\bullet$ Diagnosis Classification] \hfill \\
        Inputs: 1-12 seconds audio signal $\mid$ Output: Choose one most likely diagnosis out of 5 choices \\
        Sequence Length: 100K-1M $\mid$ Frequency: 48K Hz \\
        Metric: Accuracy
\end{description}

\textbf{Inference - Subject Profiling}
\begin{description}[itemsep=0pt, topsep=1pt, font=\normalfont]
    \item[$\bullet$ Cough Detection with Good Quality Samples] \hfill \\
        Inputs: 1-12 seconds audio signal $\mid$ Output: Binary detection of cough presence \\
        Sequence Length: 100K-1M $\mid$ Frequency: 48K Hz \\
        Metric: Accuracy
    \item[$\bullet$ Cough Detection with Poor Quality Samples] \hfill \\
        Inputs: 1-12 seconds audio signal $\mid$ Output: Binary detection of cough presence \\
        Sequence Length: 100K-1M $\mid$ Frequency: 48K Hz \\
        Metric: Accuracy
\end{description}

\subsubsection{Coswara}
\textbf{Inference - Physiological Classification}
\begin{description}[itemsep=0pt, topsep=1pt, font=\normalfont]
    \item[$\bullet$ Audio Type Classification] \hfill \\
        Inputs: Audio $\mid$ Output: Audio type (breathing/speech/cough) \\
        Sequence Length: $>$1M $\mid$ Frequency: 48 kHz \\
        Metric: Accuracy
    
    \item[$\bullet$ Diagnosis Classification with Speech Audio] \hfill \\
        Inputs: Speech audio $\mid$ Output: Healthy or COVID positive \\
        Sequence Length: $>$1M $\mid$ Frequency: 48 kHz \\
        Metric: Accuracy
\end{description}

\textbf{Inference - Subject Profiling}
\begin{description}[itemsep=0pt, topsep=1pt, font=\normalfont]
    \item[$\bullet$ Diagnosis Classification with Cough Audio] \hfill \\
        Inputs: Cough audio $\mid$ Output: Healthy or COVID positive \\
        Sequence Length: 100K-1M $\mid$ Frequency: 48 kHz \\
        Metric: Accuracy
    
    \item[$\bullet$ Diagnosis Classification with Cough Audio and Symptom Information] \hfill \\
        Inputs: cough audio \& symptom info $\mid$ Output: Healthy or COVID positive \\
        Sequence Length: $>$1M $\mid$ Frequency: 48 kHz \\
        Metric: Accuracy
    
    \item[$\bullet$ Diagnosis Classification with Symptom Information Only] \hfill \\
        Inputs: Symptom info $\mid$ Output: Healthy or COVID positive \\
        Sequence Length: NA $\mid$ Frequency: N/A \\
        Metric: Accuracy
\end{description}

\subsection{Example Prompt}
\label{app_subsec:example_prompt}
Here we showcase the example prompts given to the LLM for each task category.

\subsubsection{Perception}

\begin{promptbox}[title=Satistical Calculation]
\textbf{Prompt:} The continuous glucose monitors (CGM) data for this subject is provided in \texttt{input/data.csv}. There are two columns in this csv file: one is timestamp containing the time of each reading, and the other column \texttt{Libre GL} contains glucose values (mg/dL). 
    
Calculate percentage of time CGM is below and above normal range ($70 - 180$ mg/dL). Please calculate and output your final answer as a JSON object without any other text in the following format:

\{\newline
\texttt{"below": [float, percentage of time CGM $<$ 70 mg/dL],
\newline
"above": [float, percentage of time CGM $>$ 180 mg/dL]}
\newline\}
\end{promptbox}

\begin{promptbox}[title=Feature Extraction]
\textbf{Prompt:} One EEG C3/A2 raw signal is saved in the file \texttt{input/data.npy}. Its sampling frequency is $64$ Hz. Calculate the relative bandpower of the signal in frequency range $0.5-40$ Hz.

Please use Welch's method with \texttt{nperseg=1024} to estimate the power spectral density, and use Simpson's method to integrate the power spectral density. Please output your final answer in the following JSON format without any other text:

\{\newline
\texttt{
"delta": [ratio of delta band (0.5-4 Hz)],\newline
"theta": [ratio of theta band (4-8 Hz)],\newline
"alpha": [ratio of alpha band (8-12 Hz)],\newline
"sigma": [ratio of sigma band (12-16 Hz)],\newline
"beta": ~[ratio of beta band (16-30 Hz)],\newline
"gamma": [ratio of gamma band (30-40 Hz)]
}
\newline\}
\end{promptbox}

\subsubsection{Inference}

\begin{promptbox}[title=Event Localization]
\textbf{Prompt:} The AF (Air Flow) and THX (Thoracic respiratory) signals are provided at \texttt{input/AF.npy} and \texttt{input/THX.npy}. Both signals are sampled at $8$Hz, and this time window contains hypopnea events.

Please analyze the signals and output your final answer of a list of hypopnea events (start and end timestamps) in the following JSON format without any other text:

\{\newline
\texttt{
"start": [float, start timestamp in seconds],\newline
"end": ~~[float, end timestamp in seconds]
}
\newline\}
\end{promptbox}

\begin{promptbox}[title=Physiological Classification]
\textbf{Prompt:} There are two EOG signals saved as numpy arrays: \texttt{input/EOG\_E2\_A1.npy} (EOG channel E2-A1), and \texttt{input/EOG\_E1\_A2.npy} (EOG channel E1-A2). Both signals are sampled at $64$Hz.

Please analyze the signals and determine whether this is REM sleep or NREM sleep in JSON format. Please output your final result in a JSON object without any other text:

\{\newline
\texttt{
"stage": [NR|R]
}
\newline\}
\end{promptbox}

\begin{promptbox}[title=Subject Profiling]
\textbf{Prompt:} A PERG (Pattern Electroretinogram) recording from both eyes is provided.
The right eye signal is in \texttt{input/re\_signal.csv} and left eye signal is in \texttt{input/le\_signal.csv}. Each file contains two columns: \texttt{time\_ms} (time in milliseconds) and \texttt{amplitude} (signal amplitude in $\mu V$).

Your task is to analyze the PERG recording and determine if the subject has normal eyes or has disease on eye.

Possible classifications:
\newline
\texttt{
- "normal": The subject has normal eyes\newline
- "disease": The subject has disease on eyes
}

Based on your analysis of the waveform characteristics, choose the most appropriate classification.

Output your answer in JSON format:

\{\newline
\texttt{
"eye status": [your classification]
}
\newline\}
\end{promptbox}

\subsubsection{Generation}

\begin{promptbox}[title=Signal Imputation]
\textbf{Prompt:}  The ECG signal with missing values is provided at \texttt{input/ECG\_missing.npy}. The signal is a $128$ Hz time series of total length $150$ seconds, where missing values are represented as $0$. Please fill in the missing values in the ECG signal. Note that the missing values are consecutive and correspond to a $30$-second segment within the $150$-second window.

Save your imputed signal in a Python dictionary format with the following structure:

\{\newline
\texttt{
"ECG": [your imputed signal as a NumPy array]
}
\newline\}

Then, save this dictionary to the output path \texttt{output/imputed\_signal.npz}. The numpy array should only contain the imputed values for the missing segments, while the observed values should not included. Your array should have the shape ($3840$,). Ensure that your imputed values are realistic and consistent with the observed data.
\end{promptbox}

\begin{promptbox}[title=Future Forecasting]
\textbf{Prompt:}  The ECG signal for the past $120$ seconds is provided at \texttt{input/ECG.npy}. The signal is a $128$ Hz time series. Please forecast the next $30$ seconds of the ECG signal based on the provided data.

Save your forecasted signal in a Python dictionary format with the following structure:

\{\newline
\texttt{
"ECG": [your forecasted signal as a NumPy array]
}
\newline\}

Then, save this dictionary to the output path \texttt{output/forecasted-signal.npz}. The numpy array should have the shape ($3840$,). Ensure that your forecasted values are realistic and consistent with the observed data.
\end{promptbox}

\begin{promptbox}[title=Cross-modal Translation]
\textbf{Prompt:}  The EEG signal from channel EEG C3/A2 for the past $120$ seconds is provided at \texttt{input/EEG\_C3\_A2.npy}. The signal is a $64$ Hz time series.

Please translate this signal to the corresponding EEG C4/A1 channel for the same time period. The EEG C4/A1 channel is also a $64$ Hz time series.

Save your translated signal in a Python dictionary format with the following structure:

\{\newline
\texttt{
"EEG\_C4\_A1": [your translated signal as a NumPy array]
}
\newline\}
        
Then, save this dictionary to the output path \texttt{output/translated\_signal.npz}. The numpy array should have the shape ($7680$,). Ensure that your translated values are realistic and consistent with the observed data.
\end{promptbox}

\subsubsection{Deduction}

\begin{promptbox}[title=Temporal Ordering]
\textbf{Prompt:} Two sets of $3$-hour $3$-axis accelerometer signals are provided in the format \texttt{input/[axis]\_1.npy} and \texttt{input/[axis]\_2.npy} where \texttt{axis} is \texttt{x}, \texttt{y}, or \texttt{z}. The signals are sampled at $100$ Hz. One set is from daytime (11:00 AM - 2:00 PM) and the other is from nighttime (12:00 AM - 3:00 AM). 

Determine which set of signals is from daytime and which is from nighttime. Output your final answer as a JSON object without any other text, in the following format:

\{\newline
\texttt{
"is\_daytime\_first": [1|0]
}
\newline\}

 where \texttt{0} means \texttt{input/[axis]\_1.npy} is from daytime and \texttt{input/[axis]\_2.npy} is from nighttime, \texttt{1} means \texttt{input/[axis]\_2.npy} is from daytime and \texttt{input/[axis]\_1.npy} is from nighttime.
\end{promptbox}

\begin{promptbox}[title=Trajectory Analysis]
\textbf{Prompt:} You are given ECG, AF, EEG C3/A2 signals from two visits of the same subject. ECG represents Electrocardiogram; AF represents Airflow; EEG C3/A2 represents
Electroencephalogram. The signals are provided as 1D numpy arrays in \texttt{input/[ch]\_visit1.npy} and \texttt{input/[ch]\_visit2.npy}, where \texttt{[ch]} is one of [\texttt{ECG}, \texttt{AF}, \texttt{EEG\_C3\_A2}]. Your task is to predict which visit has higher BMI. Output your final answer as a JSON object without any other text, in the following format:

\{\newline
\texttt{
"higher\_bmi": ["visit1"|"visit2"]
}
\newline\}
\end{promptbox}

\subsection{Living Ecosystem}
\label{app_subsec:living_ecosys}

Unlike traditional benchmarks that function as static snapshots of performance, our framework establishes a ``Living Ecosystem'' designed for continuous evolution and community-driven expansion. At the core of this ecosystem lies a rigorous decoupled architecture, exemplified by the separation of the \lstinline|ExperimentBase| and \lstinline|AgentBase| abstract classes. By enforcing a unified API through Python’s ABC (Abstract Base Class) structures, we have created a standardized interface that allows researchers to inject new reasoning tasks or integrate novel agentic frameworks without altering the underlying evaluation engine. This modularity transforms the benchmark from a fixed dataset into a dynamic scaffolding where the ``Task'' and the ``Solver'' are interchangeable components, ensuring the system grows organically alongside the rapid pace of LLM advancement.

\subsubsection{Task Template}
\label{app_subsec:task_base_code}

The ``Living'' aspect of the ecosystem is powered by the \lstinline|ExperimentBase| class, which serves as a universal blueprint for community contribution. By abstracting the experiment lifecycle into five distinct, enforceable stages: \lstinline|prepare_data|, \lstinline|data_iterator|, \lstinline|run_agent|, \lstinline|parse_output|, and \lstinline|calculate_metrics|, we lower the barrier to entry for introducing complex reasoning domains. A community contributor need only implement these specific methods to plug a new task using existing dataset or a new set of tasks from a new dataset into the ecosystem. This rigid yet flexible structure ensures that while the content of the benchmark diversifies through community effort, the scientific rigor of the evaluation remains consistent. The framework handles the orchestration, logging, and state management, allowing contributors to focus purely on defining the logic of the task itself.

\begin{lstlisting}[language=Python]
class ExperimentBase(ABC):
    """
    Abstract base class for orchestrating agent-based experiments.

    This class defines the standard lifecycle of an experiment:
    1. Data preparation (`prepare_data`)
    2. Input iteration (`data_iterator`)
    3. Agent execution (`run_agent`)
    4. Output parsing (`parse_output`)
    5. Metric calculation (`calculate_metrics`)

    Attributes:
        task (str): The name of the specific task being experimented on.
        num_test (int): The maximum number of test samples to run.
        logs_dir (Path): The directory where logs will be stored. 
        agent (Optional[AgentBase]): The agent instance to be evaluated.
    """

    def __init__(
        self,
        task: str,
        num_test: int = 50,
        logs_dir: Optional[Path] = None,
        agent: Optional[AgentBase] = None,
    ):
        """
        Initialize the ExperimentBase.

        Args:
            task (str): The identifier for the experiment task
            num_test (int, optional): The number of samples to test. Defaults to 50.
            logs_dir (Optional[Path], optional): Custom path for logging. If None
            agent (Optional[AgentBase], optional): The agent instance to run the 
                experiment against. Defaults to None.
        """
        self.num_test = num_test
        # Set default logs directory if not provided and task is specified
        self.logs_dir = logs_dir
        if self.logs_dir and not self.logs_dir.exists():
            self.logs_dir.mkdir(parents=True, exist_ok=True)
        self.task = task
        self.agent = agent

    @abstractmethod
    def prepare_data(self) -> None:
        """
        Prepare the dataset and environment before the experiment loop begins.

        This method should handle tasks such as:
        - Downloading datasets.
        - Loading raw files into memory.
        - Data preprocessing
        - Setting up necessary global state for the experiment.
        """
        pass

    @abstractmethod
    def data_iterator(self) -> Generator[Dict[str, Any], None, None]:
        """
        Yield individual data samples for the experiment loop.

        Returns:
            Generator[Dict[str, Any], None, None]: A generator yielding a dictionary 
            representing a single test case (containing 'input_data', 'GT', 'metadata').
        """
        pass

    @abstractmethod
    async def run_agent(self, data: Dict[str, Any]) -> Dict[str, Any]:
        """
        Execute the agent logic for a single data sample.

        Args:
            data (Dict[str, Any]): The input data for the current step (yielded 
                by `data_iterator`).

        Returns:
            Dict[str, Any]: The raw result from the agent execution, usually containing
            the agent's response, trace, or internal state.
        """
        pass

    @abstractmethod
    def parse_output(
        self, content: Optional[str] = None, query_id: Optional[str] = None
    ) -> Tuple[Dict[str, Any], Any]:
        """
        Parse the raw output from the agent into a structured format for evaluation.

        This method supports two modes:
        1. Parsing a direct string (`content`).
        2. Loading output from a file associated with `query_id` if the agent 
           writes to disk.

        Args:
            content (Optional[str], optional): The raw string output from the agent.
            query_id (Optional[str], optional): The unique ID of the query, used to 
                locate output files if `content` is not sufficient.

        Returns:
            Tuple[Dict[str, Any], Any]: A tuple containing:
                - A dictionary with the parsed solution/answer.
                - A reason for failure (if any), or None if successful.
        """
        pass

    @abstractmethod
    def calculate_metrics(self, result_list: List[dict]) -> ExperimentMetrics:
        """
        Compute aggregate performance metrics for the entire experiment.

        Args:
            result_list (List[dict]): A list of dictionaries, where each dictionary
                contains the parsed results and ground truth for a single test case.

        Returns:
            ExperimentMetrics: A structured object containing calculated metrics 
            (e.g., Accuracy, IoU, SMAPE).
        """
        pass
\end{lstlisting}

\subsubsection{Agent Template}
\label{app_subsec:agent_base_code}

The robustness of this ecosystem is secured by the \lstinline|AgentBase| class, which standardizes the interaction between the evaluation environment and the rapidly proliferating landscape of LLMs. By distilling complex model interactions into a single, unified asynchronous query method, the framework becomes model-agnostic. Whether evaluating a proprietary model like GPT-4 or a local, open-source agent, the system treats them as uniform entities adhering to the same protocol. This polymorphism ensures that our benchmark is not brittle: it is ready to assess the agents of tomorrow immediately. Consequently, the framework acts as a bridge, allowing any standardized agent to attempt any community-contributed task, effectively creating a many-to-many testing matrix that defines a truly thriving, collaborative ecosystem.

\begin{lstlisting}[language=Python]
class AgentBase(ABC):
    """
    Abstract base class defining the interface for AI agents.

    This class serves as a blueprint for creating specific agent implementations
    that interact with various models (LLMs). It enforces a consistent structure
    for initialization and querying, ensuring that all subclasses provide the
    necessary logic to handle prompts, context data, logging, and response generation.

    Attributes:
        model_name (str): The identifier of the specific model being used (e.g., 'gpt-4', 'claude-3').
        name (str): The name of the agent instance, defaulting to the class name.
    """

    def __init__(self, model_name: str, **kwargs):
        """
        Initialize the AgentBase instance.

        Args:
            model_name (str): The name or identifier of the underlying model to be used
                by the agent (e.g., "gpt-4-turbo").
            **kwargs: Additional keyword arguments for configuration or model-specific
                parameters. These can be captured and used by subclasses.
        """
        self.model_name = model_name
        self.name = self.__class__.__name__

    @abstractmethod
    async def query(
        self, prompt: str, data: Dict[str, Any], logs_dir: Path, query_id: str
    ) -> str:
        """
        Execute an asynchronous query against the agent.

        This abstract method must be implemented by subclasses to define how the
        agent processes a prompt and data. It is responsible for formatting the
        input, interacting with the model API, handling the conversation history,
        and persisting logs to the specified directory.

        Args:
            prompt (str): The main instructional text or query to send to the agent.
            data (Dict[str, Any]): Contextual data required for the query. This may include
                signals, channel information, user history, or other relevant metadata.
            logs_dir (Path): The filesystem path where logs, artifacts, and conversation
                traces should be saved for debugging and auditing.
            query_id (str): A unique identifier for the specific query execution, used
                for tracing and organizing log files.

        Returns:
            str: The final textual response content generated by the agent.

        Raises:
            NotImplementedError: If the subclass does not implement this method.
        """
        pass
\end{lstlisting}

\section{Experiment Setting Details}
\label{app_sec:exp_details}

\subsection{Naive Baseline}
\label{app_subsec:naive_baseline}
To contextualize model performance across diverse metrics, we introduce a set of naive baselines tailored to each task type. The detailed calculation methods are: (1) For tasks evaluated via Accuracy, the baseline employs a uniform random guessing strategy ($1/\text{num\_options}$). (2) For IoU-based localization tasks, we generate a random interval $[t, t+d]$, where the start time $t$ is uniformly sampled from the valid range and the duration $d$ equals the average event length in the test set. (3) For tasks evaluated by the sMAPE-score, the baseline strategy varies by category: we utilize the population mean for Cross-modal Translation tasks and the input signal's mean as a constant prediction for Future Forecasting and Signal Imputation tasks.

Notably, the baseline for perception tasks defaults to a perfect reference score of 1.0, reflecting the clear, deterministic analytical solutions inherent to such tasks. For instance, calculating sleep efficiency relies on a strict mathematical definition (total sleep time / time in bed), naturally yielding a perfect score when executed correctly.

\subsection{Model Intelligence Index}
\label{app_subsec:intel_index}
To quantify the general reasoning capabilities of the LLMs in our study, we adopt the Intelligence Index methodology proposed by \citealp{kim2025towards}. This composite metric is designed to isolate intrinsic model performance across general-purpose reasoning, knowledge retrieval, mathematics, and coding, explicitly independent of agentic workflows or multi-agent collaboration structures. The index is constructed via an equal-weighted aggregation of eight standardized evaluation suites: MMLU-Pro \cite{wang2024mmlupro}, GPQA Diamond \cite{rein2024gpqa}, HLE \cite{phan2025hle}, AIME 2025, SciCode \cite{tian2024scicode}, LiveCodeBench \citep{jain2024livecodebench}, IFBench \citep{pyatkin2025ifbench}, and AA-LCR \citep{artificialanalysis2025aalcr}. To ensure consistency and reliability, all models are evaluated under uniform zero-shot, instruction-prompted conditions using robust equality checks and pass@1 scoring. Table~\ref{tab:intel_index} presents the calculated Intelligence Index for all evaluated LLMs. The intelligence indices for \glmfive and \geminiProthree are omitted, as \glmfive lacks official AIME 2025 and LiveCodeBench scores, and \geminiProthree lacks an official LiveCodeBench pass@1 scoring.

\input{tables/table_Appendix_intelligence_index}

\subsection{SOTA Machine Learning Baseline}
\label{app_subsec:sota_ml_baseline}
To strictly quantify the performance disparity between generalist LLMs and specialized models, we curate a registry of state-of-the-Art  machine learning baselines for a representative subset of 32 tasks within \name. We acknowledge that these external baselines are drawn from diverse literature, datasets, and may operate under varying experimental protocols (e.g., specific data splits or preprocessing pipelines). Consequently, these comparisons should not be interpreted as a strictly controlled competitive evaluation, but rather as an indicative upper bound of domain-specific capability. Despite these methodological variations, we ensure semantic alignment in task definitions to provide valid insights into the limitations of current LLM reasoning. Table~\ref{tab:sota_benchmarking_summary} details the specific model architectures, reported performance metrics, and original sources for these 32 reference tasks.

\input{tables/table_Appendix_sota_baseline}

\section{More Results}
\label{app_sec:more_results}
\subsection{Performance Details}
\label{app_subsec:performance_details}

\begin{figure*}[h!]
\centering
\includegraphics[width=0.9\textwidth]{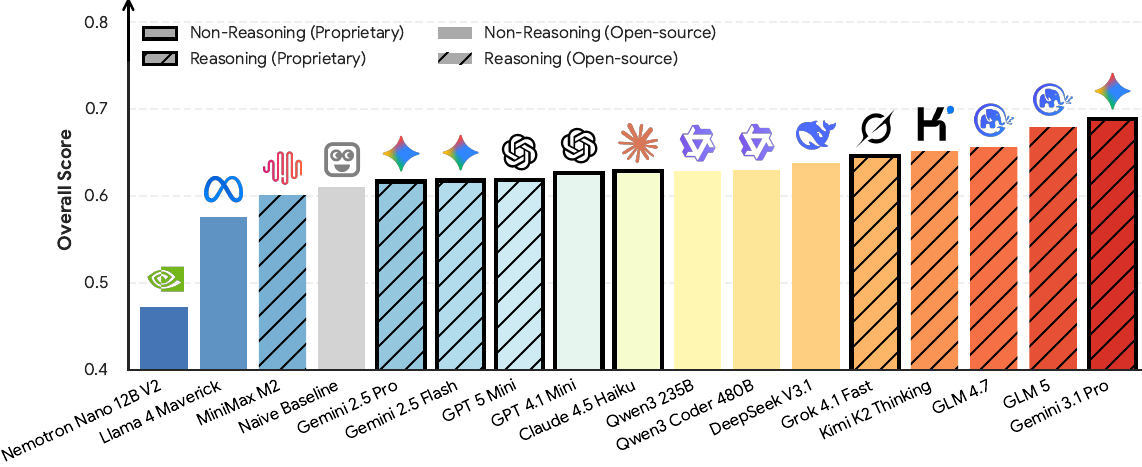}
\caption{\textbf{\name Leaderboard.} We report results for 16 state-of-the-art LLMs.}
\label{fig:llm_performance}
\vspace{-10pt}
\end{figure*}

 In this section we report the comparative performance of 16 LLMs on \name. Fig.~\ref{fig:llm_performance} presents leaderboard for \name. All models are ranked by their overall scores.

Table~\ref{tab:task_breakdown} provides a granular performance breakdown for all 16 LLMs and naive baselines across the full suite of 110 tasks.

\input{tables/table_Appendix_all_task_breakdown}

\subsection{Supplement Analysis Across Task Categories}
\label{app_subsec:supp_analysis_task_category}
To contextualize our analysis in Sec.~\ref{subsec:task_category}, we provide supplement analysis in this section.

\subsubsection{Perception Task}

\begin{figure}[h!]
    \vspace{-5pt}
    \centering
    \includegraphics[width=0.6\linewidth]{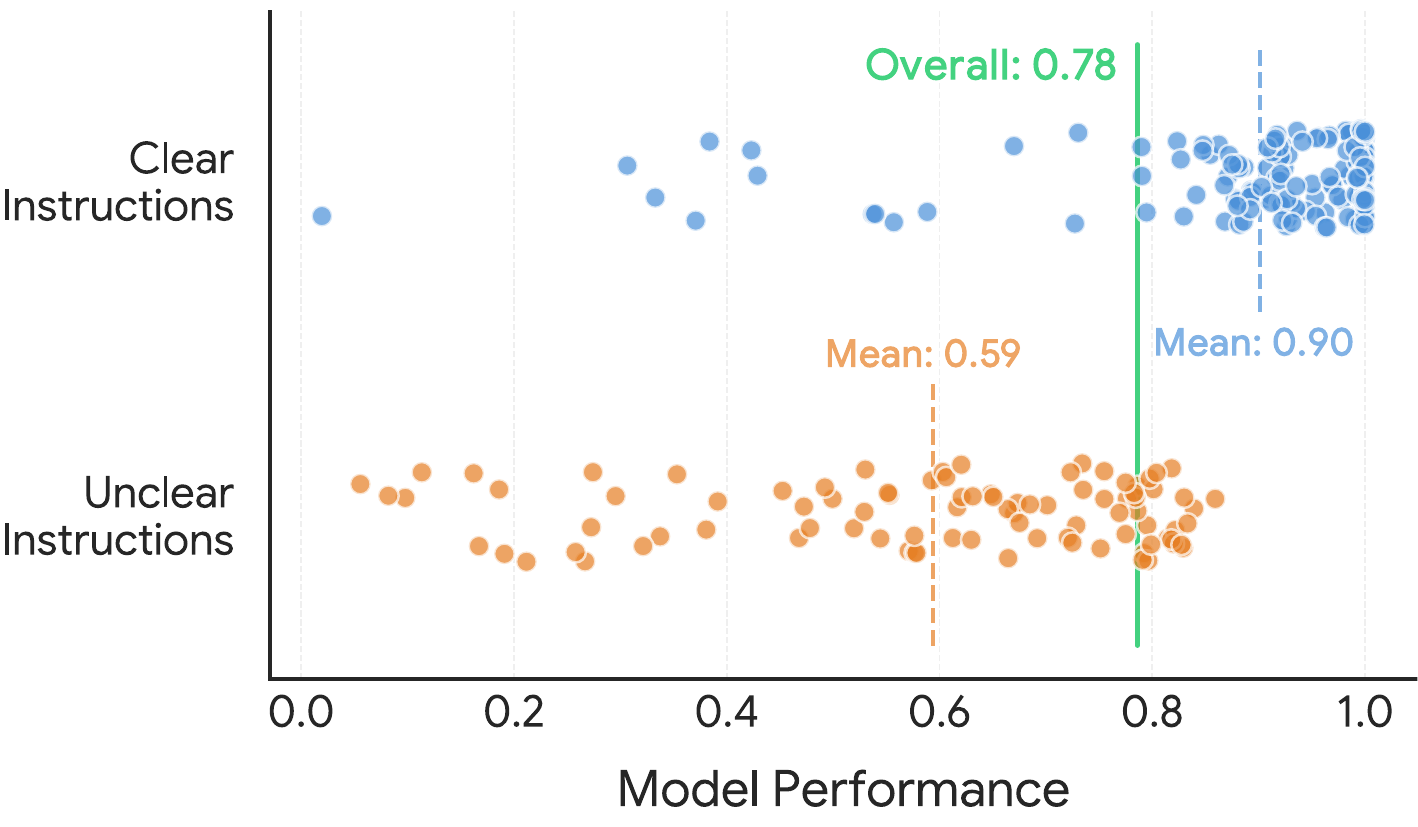}
    \caption{\textbf{Impact of instruction clarity in Perception category.} Each dot represents the performance of a single model on a specific perception task.}
    \label{fig:perception_instruction}
    \vspace{-5pt}
\end{figure}

The data visualized in Fig.~\ref{fig:perception_instruction} illustrates a critical insight: while LLMs possess a strong baseline level for perception tasks in general, their reliability is heavily dependent on how the problem is framed. It reveals a stark divergence between the two conditions. If LLMs were provided with step-by-step guidance, performance is propelled to a mean of 0.90, effectively unlocking the model's full potential. Conversely, the scattered orange distribution below shows that leaving instructions result-oriented creates significant volatility, dragging the mean score down to 0.59. A fundamental limitation is revealed: current LLMs still lack robust autonomous reasoning capabilities. They rely heavily on detailed instructions to perform effectively, struggling to bridge the gap from a simple target to a complex solution on their own.

\subsubsection{Generation Task}

\begin{figure*}[ht]
\centering
\begin{minipage}[b]{0.32\textwidth}
    \centering
    \includegraphics[width=\linewidth]{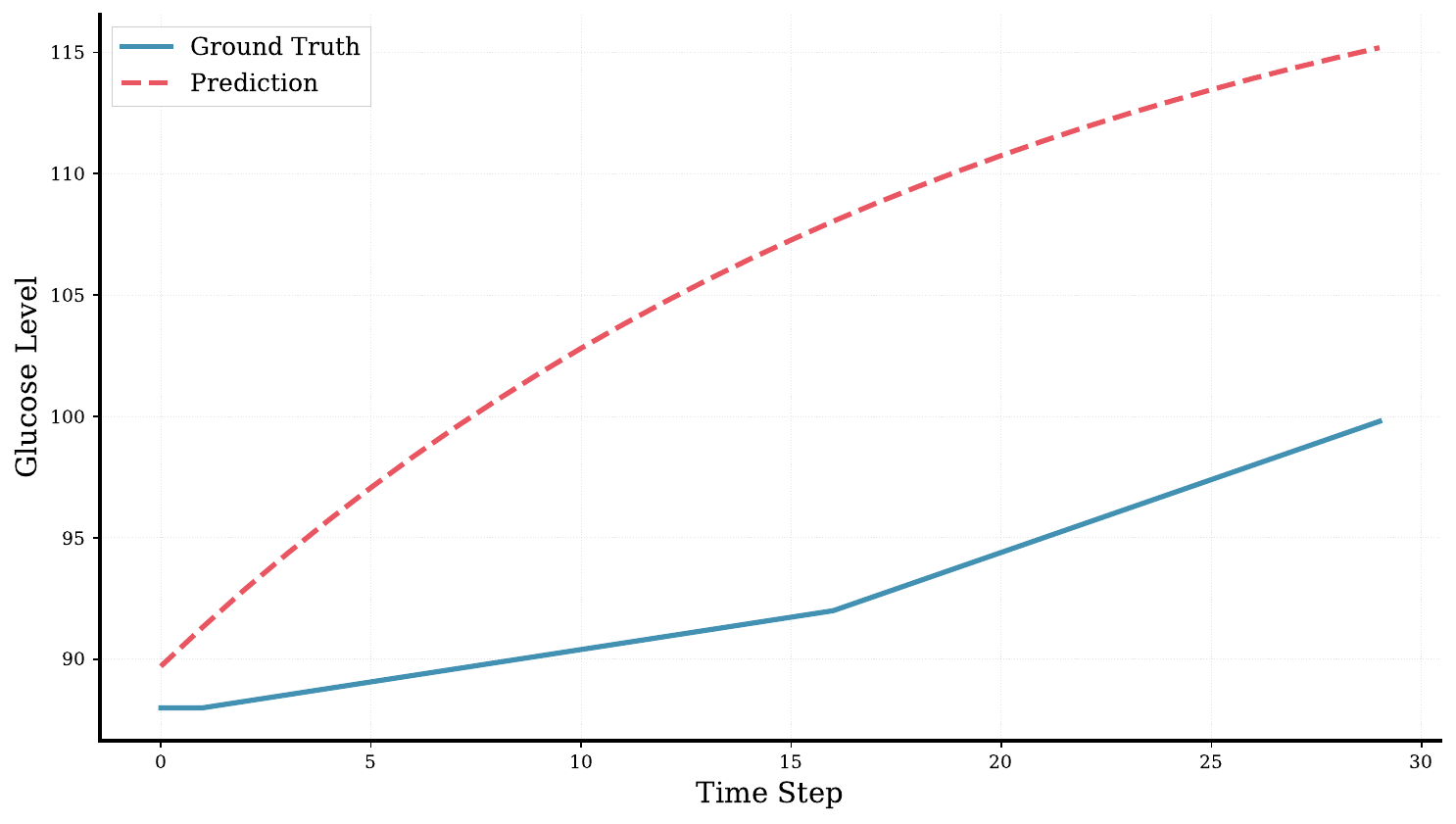}
    \caption*{Polynomial Fitting Pattern}
\end{minipage}
\hfill
\begin{minipage}[b]{0.32\textwidth}
    \centering
    \includegraphics[width=\linewidth]{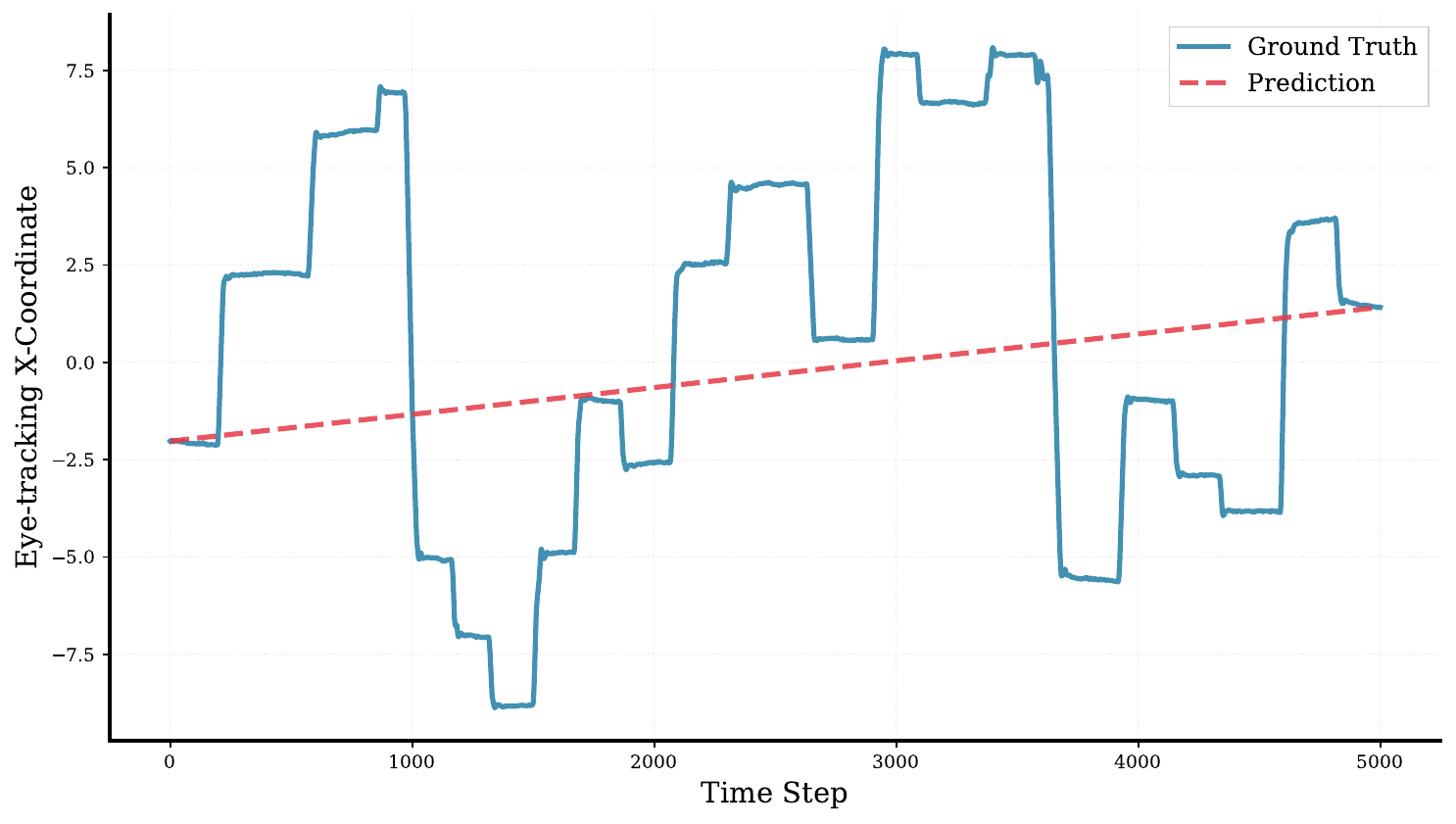}
    \caption*{Linear Interpolation Pattern}
\end{minipage}
\hfill
\begin{minipage}[b]{0.32\textwidth}
    \centering
    \includegraphics[width=\linewidth]{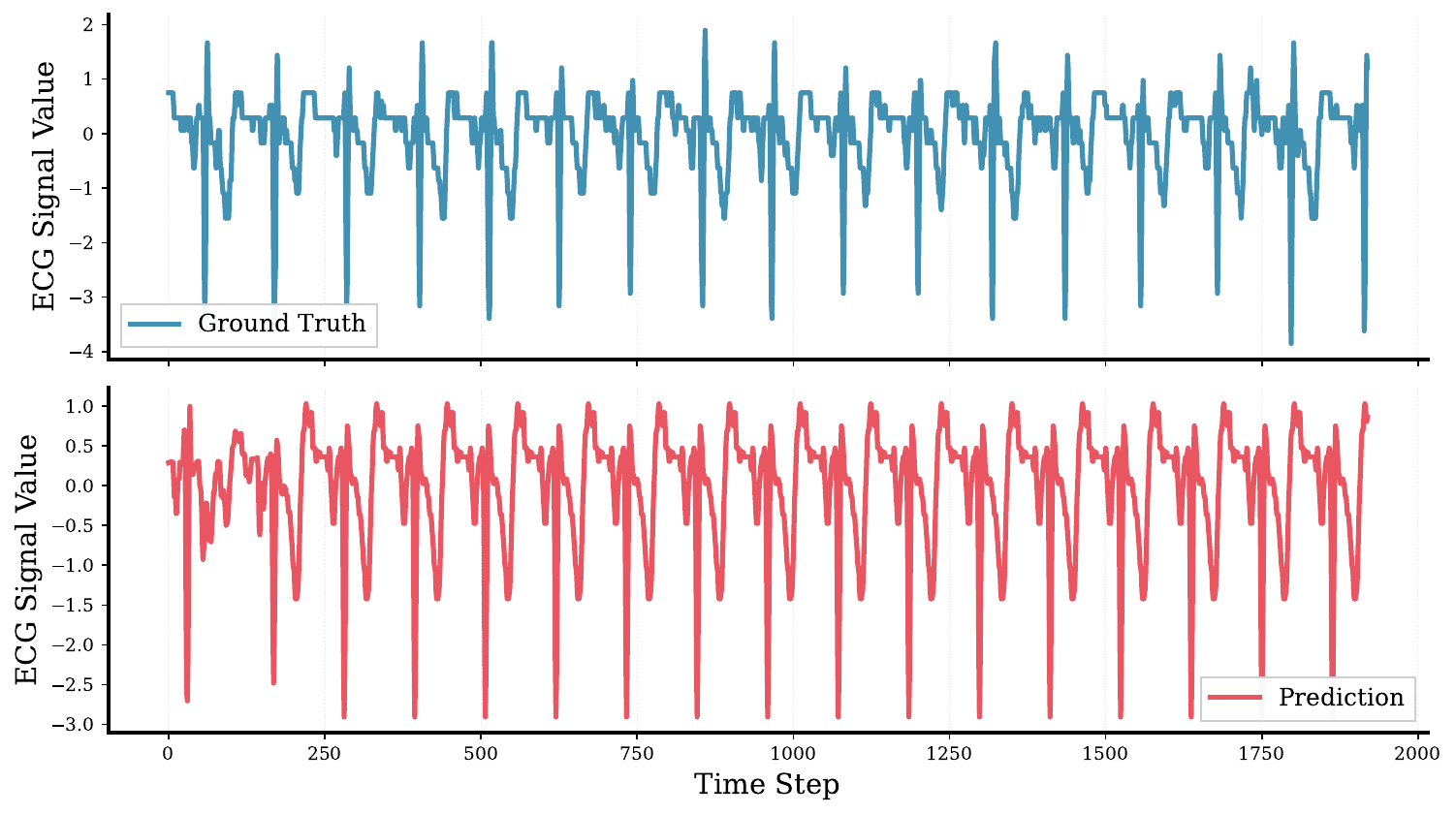}
    \caption*{Copy-Paste Pattern}
\end{minipage}
\caption{\textbf{Common Generation Patterns exhibited by LLMs for sequence imputation and forecasting.} The three most prevalent strategies are: polynomial curve fitting, linear interpolation, and copy-pasting observed segments.}
\label{fig:gen_patterns}
\end{figure*}

Fig.~\ref{fig:gen_patterns} shows 3 most common patterns of how LLMs approach time-series generation tasks. The visualization reveals a critical tendency: rather than modeling the intricate stochastic dynamics or causal dependencies of the data, LLMs frequently collapse predictions into elementary mathematical functions. As observed in the polynomial fitting pattern, the model smooths distinct glucose level fluctuations into a generic curve, drifting significantly from the actual trend. Similarly, the linear interpolation ignores the structured and quasi-periodic behavior of eye-tracking coordinates, cutting through the data with a naive straight line. The copy-paste behavior further highlights this lack of genuine reasoning capability, where the LLM simply replicates previous segments of an ECG signal. These failures suggest that instead of solving problems from deep analysis of time series itself, generation tasks are governed by low-complexity heuristics.

\subsection{Heatmap on Input Modality}
\label{app_subsec:modality_heatmap}
\begin{figure}[h!]
    \centering
    \includegraphics[width=0.7\linewidth]{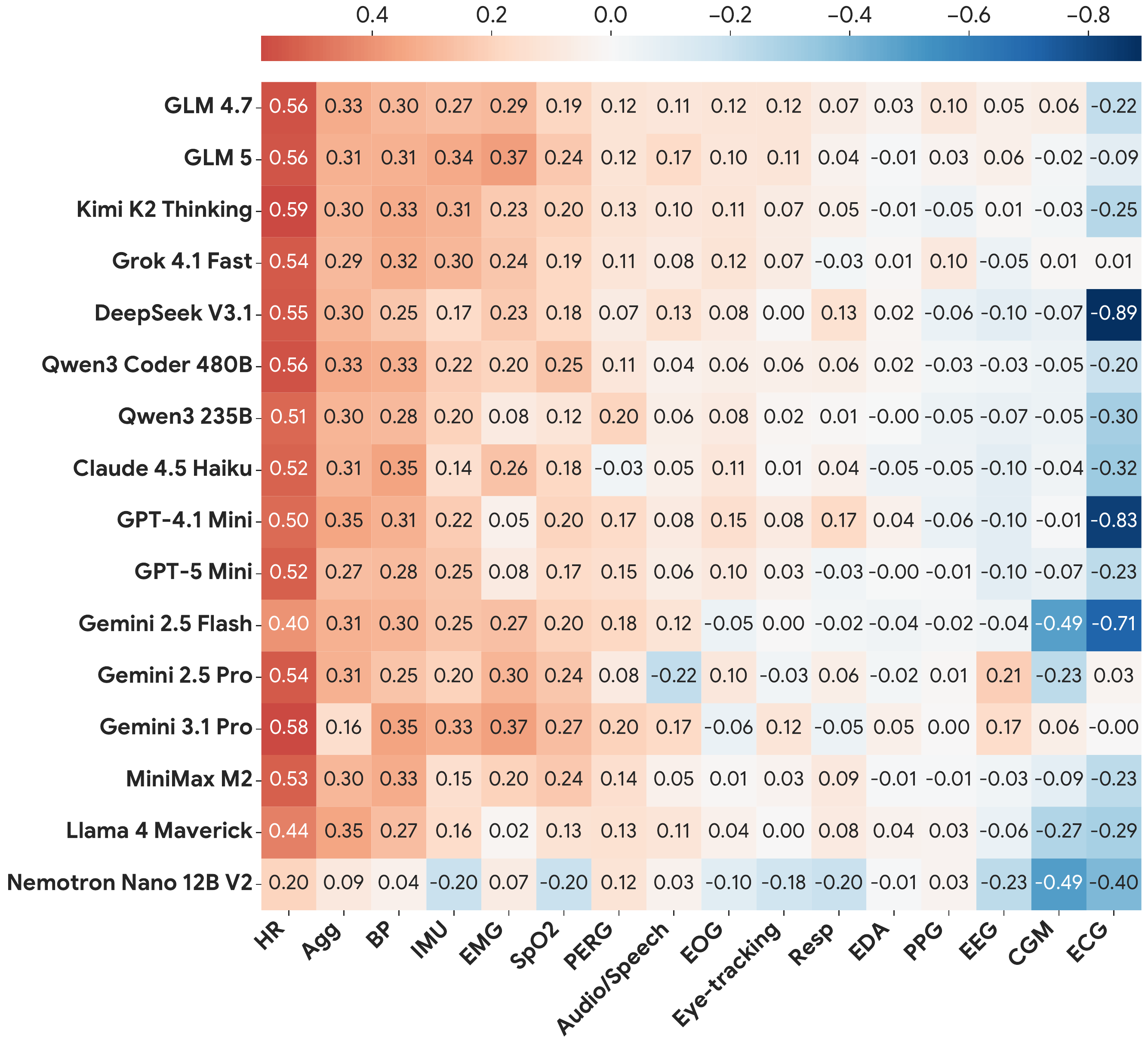}
    \caption{\textbf{Model performance heatmap across input modality.}}
    \label{fig:llm_modality_heatmap}
\end{figure}

We provide a granular visualization of model performance across distinct input modalities, supplementing the domain-level analysis in Sec.~\ref{subsec:temp_reso_data_property}. As illustrated in Fig.~\ref{fig:llm_modality_heatmap}, the heatmap reveals a striking stratification of difficulty inherent to specific signal types, mirroring the inter-model consistency observed in our domain analysis. This indicates that performance is primarily governed by the intrinsic semantic accessibility of the signal representation rather than model-specific architectural biases. Specifically, we observe a universal competence in processing high-level, structured signals (e.g., HR, Aggregated Data, and BP), which yield consistently positive Kappa scores across the model cohort. Conversely, waveform-dense modalities such as ECG and CGM impose a uniform performance bottleneck, resulting in negative scores even for advanced models. This suggests that current LLM tokenization strategies may be fundamentally misaligned with the continuous, high-frequency nature of these specific biosignals, creating a 'representation gap' that scaling alone cannot easily bridge.

\subsection{Input Format Experiments}
\label{app_subsec:input_format}

To support the input format evaluation discussed in Sec.~\ref{subsec:input_format}, we specify 10 representative tasks and their corresponding input time series, whose text-based representations fit within the LLM context window. Detailed performance results for each task are reported in Table~\ref{tab:input_format_exp}.

\input{tables/table_Appendix_input_format_exp}

\subsection{Supplementary Material for Additional Studies}
\label{app_subsec:supp_casestudy}

\subsubsection{Influence of Information Quality}

\begin{figure}
    \centering
    \includegraphics[width=0.95\linewidth]{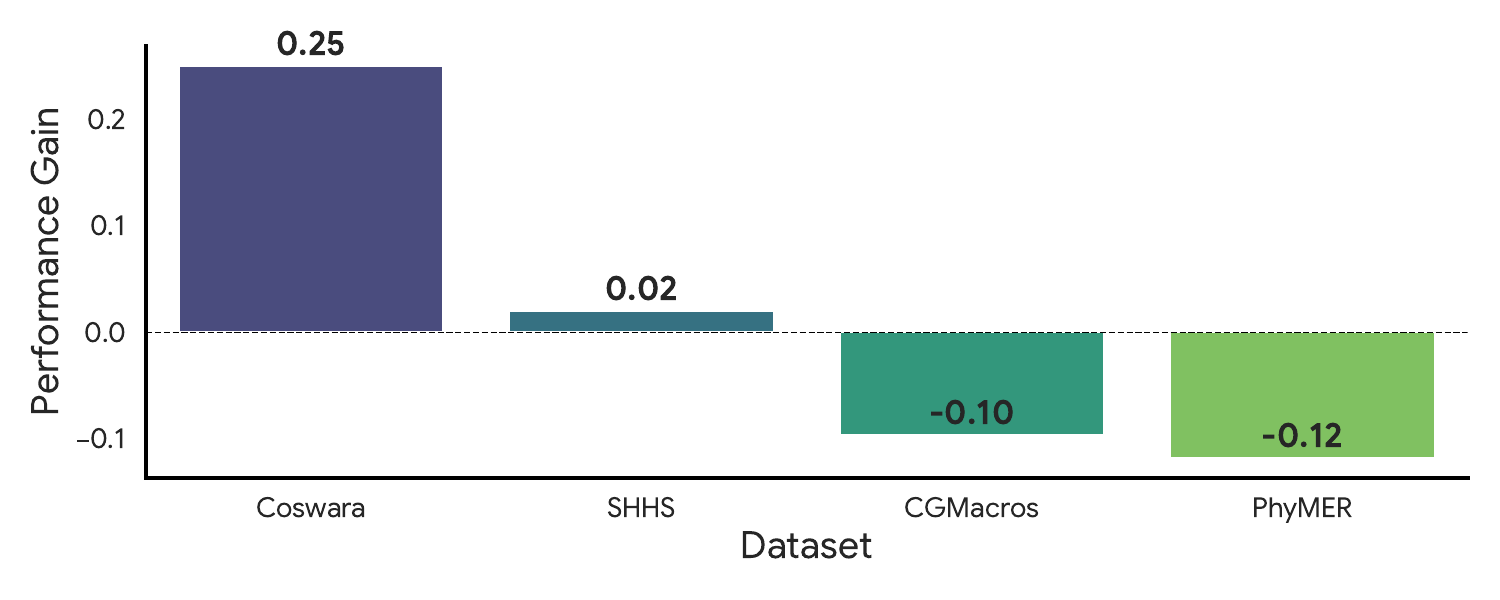}
    \vspace{-10pt}
    \caption{\textbf{Performance Gain with Auxiliary Information}}
    \label{fig:info_fusion}
\end{figure}

The experiments in this section investigated a critical question: can Large Language Models (LLMs) effectively synthesize auxiliary information to achieve their goals? This serves as a vital probe for their reasoning capabilities. As shown in Fig.~\ref{fig:info_fusion}, the results reveal that adding data does not automatically improve performance. In the Coswara dataset, where the additional input consisted of clear symptom descriptions, the model could directly apply this context to aid its diagnosis. However, in the other three datasets, the auxiliary inputs were raw signals, abstract data formats that the model could not intuitively interpret. Consequently, performance gains in these cases were negligible or even detrimental, exposing a significant limitation in the models' ability to synthesize complex, indirect data. This evidence consolidates our conclusion: more input information does not inherently result in better outcomes.

\subsubsection{Influence of Domain-Specific Agent}

As demonstrated in Sec.~\ref{subsec:additional_studies}, a domain-specialist agent does not inherently outperform generalist models. Comparing the trajectories from the CodeAct and Biomni frameworks below, it is evident that the behavior of two agents are similar, and the specialized agent (Biomni) failed to provide additional utility in resolving the task.

\textbf{CodeAct reasoning trajectory}
\input{reasoning_traces/codeact}

\textbf{Biomni reasoning trajectory}
\input{reasoning_traces/biomni}

\subsection{OpenTSLM Experiments}
\label{app_subsec:opentslm_exp}

To test OpenTSLM without exceeding the input time series length limit, we skipped samples with input time series length greater than the limit in Sleep Stage Classification task, and downsampled time series in Activity Classification task. To ensure fair comparison, input prompt was adapted to follow the style as shown in the official Github repository of OpenTSLM. 

\input{tables/table_Appendix_opentslm}

We observed that the output of OpenTSLM was invalid or repeated words on Activity Classification task, and only 1\% of the output contained valid answers.

\subsection{Direct Reasoning Experiments}
\label{app_subsec:direct_reasoning}

In \name, we employ the CodeAct framework to operationalize complex reasoning steps. Direct pure-text ingestion of health time series is fundamentally constrained by LLM context windows, which would render 40\% to 60\% of \name tasks unmeasurable due to massive sequence lengths.

To demonstrate that CodeAct accurately reflects the models' inherent reasoning abilities rather than mere Python proficiency, we compared direct in-context reasoning against CodeAct on a representative subset of 20 tasks where sequence lengths permit pure-text ingestion.

As detailed in Table~\ref{tab:direct_reasoning}, the results demonstrate a strong alignment between the two paradigms, with high Pearson correlations ($r \ge 0.89$) across all evaluated models. The performance gains under CodeAct, particularly for Perception tasks, merely reflect the deterministic precision of code-based calculations. Furthermore, Spearman correlation analysis confirms that direct reasoning capability significantly degrades with longer sequences ($\rho = -0.59$, $p = 6.4 \times 10^{-7}$) and higher sampling frequencies ($\rho = -0.45$, $p = 4.4 \times 10^{-7}$). These findings validate that CodeAct faithfully preserves underlying reasoning logic while overcoming the severe length bottlenecks of direct text ingestion.

\input{tables/table_Appendix_direct_reasoning}

\section{\lightname: A Cost-Efficient Evaluation Subset}
\label{app_sec:hearts_lightening}

\begin{figure*}[h!]
\centering
\includegraphics[width=0.9\textwidth]{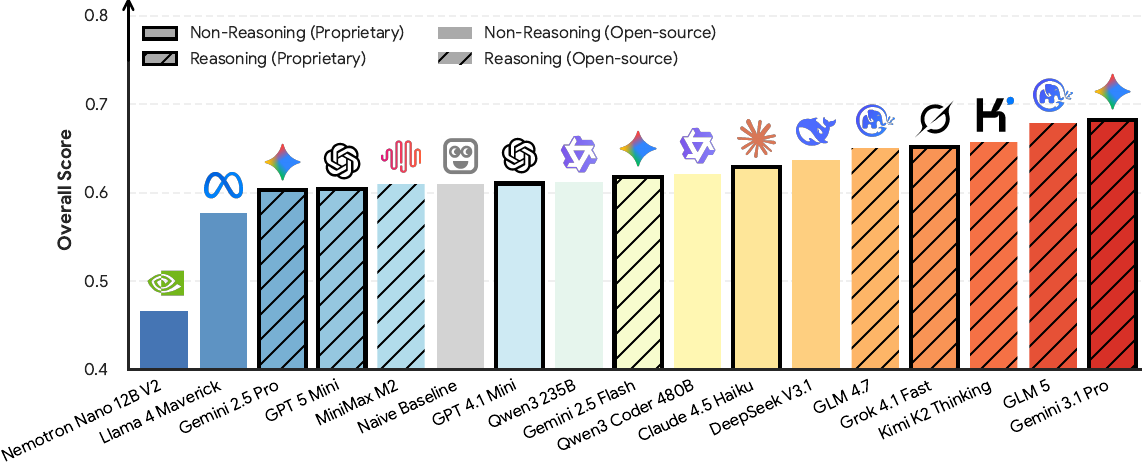}
\caption{\textbf{\lightname Leaderboard.} Model rankings and overall performance tiers remain highly consistent with the full \name.}
\label{fig:llm_performance_n50}
\vspace{-10pt}
\end{figure*}

To balance evaluation comprehensiveness with computational efficiency, we introduce \lightname, a cost-efficient subset comprising 5,417 test samples. This streamlined configuration preserves the complete taxonomy of all 110 tasks while reducing the number of test cases per task, thereby substantially lowering computational overhead without compromising the statistical stability of our evaluations.
The adjusted sample distribution is reported in Table~\ref{tab:lightening_data}, and the updated leaderboard is presented in Fig.~\ref{fig:llm_performance_n50}. Re-evaluating the models under this lightweight setting confirms that the overall performance tiers remain stable, with only minor ranking shifts observed among models with closely matched scores. Given its efficiency and strong alignment with the full benchmark, we adopt the \lightname configuration as a standard track for future evaluations.

\input{tables/table_Appendix_lightening_data}

%% file: tables/table_Appendix_data_statistics.tex
\begin{table}[ht]
    \small
    \centering
    \renewcommand{\arraystretch}{1.2}
    \caption{\textbf{Statistical summary of datasets in \name.}}
    \label{tab:data_stats}
    
    \begin{tabularx}{\textwidth}{l l >{\raggedright\arraybackslash}X c c}
        \toprule[1.5pt]
        \textbf{Dataset} & \textbf{Domain} & \textbf{Modality used in \name (Frequency)} & \textbf{\#  Test Samples} & \textbf{Data Accessibility} \\
        \midrule\midrule

        Capture24 & Motion & IMU (100Hz) & 1204 & Open Source \\
        PAMAP2 & Motion & IMU (100Hz) & 106 & Open Source \\
        Shanghai Diabetes & Metabolic & CGM (per minute) & 232 & Open Source \\
        CGMacros & Metabolic & CGM (per minute), HR (per minute), Annotation & 2333 & Open Source \\
        VitalDB & Surgery & MBP (10Hz/0.5Hz), EEG (100Hz), ECG (100Hz), PPG (100Hz), Annotation & 2000 & Open Source \\
        SHHS & Sleep & ECG (128Hz), EEG (64Hz), EOG (64Hz), Airflow (8Hz), Thoracic (8Hz), HR (1Hz), Annotation & 4799 & Restricted Access \\
        Harespod & Respiration & Respiration (100Hz), $\mathrm{SpO}_2$ (100Hz), HR (1Hz) & 632 & Open Source \\
        PhyMER & Emotion & BVP (64Hz), EDA (4Hz), TEMP (4Hz), HR (1Hz) & 1830 & Restricted Access \\
        PERG-IOBA & Ophthalmology & PERG (1700Hz) & 870 & Open Source \\
        GazeBase & Eye Movement & Eye Tracking (1000Hz) & 988 & Open Source \\
        GLOBEM & Behavior & Aggregated Data (per day) & 1140 & Restricted Access \\
        Bridge2AI-voice & Speech & Audio (100Hz) & 1600 & Restricted Access \\
        VCTK & Speech & Audio (16000Hz) & 200 & Open Source \\
        GrabMyo & Gesture & EMG (2048Hz) & 400 & Open Source \\
        CoughVID & COVID Cough & Audio (48000Hz) & 892 & Open Source \\
        Coswara & COVID Cough & Audio (48000Hz) & 1000 & Open Source \\
        
        \bottomrule[1.5pt]
    \end{tabularx}
\end{table}

%% file: tables/table_Appendix_intelligence_index.tex
\begin{table}[h!]
\centering
\setlength{\tabcolsep}{3pt} 
\caption{\textbf{Intelligence Index of all LLMs evaluated on \name.}}

\resizebox{\linewidth}{!}{
\begin{tabular}{lccccccccc}
\toprule[1.5pt]
\textbf{Model} & \textbf{Index} & \textbf{AA-LCR} & \textbf{HLE} & \textbf{MMLU-Pro} & \textbf{GPQA Diamond} & \textbf{AIME 25} & \textbf{LiveCode} & \textbf{SciCode} & \textbf{IFBench} \\
\midrule\midrule

\raisebox{-0.25em}{\includegraphics[height=1.2em]{icons/meta-color.png}} \llama & 42 & 46 & 5 & 81 & 67 & 19 & 40 & 33 & 43 \\
\raisebox{-0.25em}{\includegraphics[height=1.2em]{icons/qwen-color.png}} \qwenCoder & 45 & 42 & 4 & 79 & 62 & 39 & 59 & 36 & 41 \\
\raisebox{-0.25em}{\includegraphics[height=1.2em]{icons/openai.png}} \gptFourOneMini & 45 & 42 & 5 & 78 & 66 & 46 & 48 & 40 & 38 \\
\raisebox{-0.25em}{\includegraphics[height=1.2em]{icons/nvidia-color.png}} \nemotron & 48 & 40 & 5 & 76 & 57 & 75 & 69 & 26 & 32 \\
\raisebox{-0.25em}{\includegraphics[height=1.2em]{icons/claude-color.png}} \claudeHaiku & 58 & 70 & 4 & 80 & 67 & 84 & 62 & 43 & 54 \\
\raisebox{-0.25em}{\includegraphics[height=1.2em]{icons/gemini-color.png}} \geminiFlash & 58 & 57 & 13 & 84 & 79 & 78 & 63 & 41 & 52 \\
\raisebox{-0.25em}{\includegraphics[height=1.2em]{icons/minimax-color.png}} \minimax & 63 & 61 & 13 & 82 & 78 & 78 & 83 & 36 & 72 \\
\raisebox{-0.25em}{\includegraphics[height=1.2em]{icons/qwen-color.png}} \qwen & 63 & 67 & 15 & 83 & 79 & 91 & 79 & 42 & 51 \\
\raisebox{-0.25em}{\includegraphics[height=1.2em]{icons/deepseek-color.png}} \deepseek & 64 & 65 & 15 & 83 & 79 & 90 & 80 & 41 & 57 \\
\raisebox{-0.25em}{\includegraphics[height=1.2em]{icons/gemini-color.png}} \geminiPro & 65 & 66 & 21 & 86 & 84 & 88 & 80 & 43 & 49 \\
\raisebox{-0.25em}{\includegraphics[height=1.2em]{icons/grok.png}} \grok & 66 & 68 & 18 & 85 & 85 & 89 & 82 & 44 & 53 \\
\raisebox{-0.25em}{\includegraphics[height=1.2em]{icons/openai.png}} \gptFiveMini & 68 & 68 & 20 & 84 & 83 & 91 & 84 & 39 & 75 \\
\raisebox{-0.25em}{\includegraphics[height=1.2em]{icons/kimi-color.png}} \kimi & 68 & 66 & 22 & 85 & 84 & 95 & 85 & 42 & 68 \\
\raisebox{-0.25em}{\includegraphics[height=1.2em]{icons/chatglm-color.png}} \glm & 70 & 64 & 25 & 86 & 86 & 95 & 89 & 45 & 68 \\
\raisebox{-0.25em}{\includegraphics[height=1.2em]{icons/chatglm-color.png}} \glmfive & - & 63 & 27 & 86 & 82 & - & - & 46 & 72 \\
\raisebox{-0.25em}{\includegraphics[height=1.2em]{icons/gemini-color.png}} \geminiProthree & - & 73 & 45 & 91 & 94 & 95 & - & 59 & 77 \\

\bottomrule[1.5pt]
\end{tabular}
}
\label{tab:intel_index}
\end{table}

%% file: tables/table_Appendix_sota_baseline.tex
\renewcommand{\theadfont}{\bfseries\small}
\renewcommand{\theadalign}{bc}

\noindent
\begin{table*}[t]
\centering
\small
\caption{\textbf{Benchmarking LLM Agents against SOTA ML Models across 32 tasks on \name.} Scores represent the average and best performance achieved by the LLMs compared to SOTA baselines.}
\begin{tabular*}{0.81\textwidth}{l p{5cm} c c c r}
\toprule[1.5pt]
\textbf{Dataset} & \textbf{Task} & \thead{Avg.\\LLMs} & \thead{Best\\LLM} & \thead{SOTA \\ML} & \textbf{Source} \\
\midrule\midrule

\grayrow
\multicolumn{6}{l}{\textit{\textbf{Individual-level Analysis}}} \\
Bridge2AI-voice & Voice Parkinson’s Detection & 0.52 & 0.56 & 0.92 & \cite{jeong2024exploring} \\
CGMacros & CGM Diabetes Classification & 0.54 & 0.65 & 0.97 & \cite{ayat2024novel} \\
CGMacros & Glycemic Response Comparison & 0.83 & 0.89 & 0.87 & \cite{herrero2022identifying} \\
CoughVID & Cough Event Detection & 0.54 & 0.60 & 0.88 & \cite{orlandic2021coughvid} \\
PERG-IOBA & PERG Eye Disease Detection & 0.58 & 0.62 & 0.68 & \cite{bowd2009diagnostic} \\
PhyMER & Cross-session Emotion Match & 0.57 & 0.73 & 0.70 & \cite{singh2025feel} \\
SHHS & ECG Atrial Fibrillation & 0.56 & 0.68 & 0.99 & \cite{cai2020accurate} \\
SHHS & Cardiovasc. Mortality Pred. & 0.49 & 0.54 & 0.88 & \cite{thapa2026multimodal} \\
SHHS & PSG Smoker Classification & 0.52 & 0.56 & 0.94 & \cite{ccay2024eeg} \\
SHHS & PSG Stroke Prediction & 0.51 & 0.55 & 0.82 & \cite{thapa2026multimodal} \\
VitalDB & Perioperative Myocardial Injury & 0.52 & 0.60 & 0.79 & \cite{oh2023prediction} \\

\midrule
\grayrow
\multicolumn{6}{l}{\textit{\textbf{Physiology Classification}}} \\
Capture24 & Accel. Activity Recognition & 0.12 & 0.17 & 0.90 & \cite{burq2023human} \\
Coswara & Cough COVID-19 Detection & 0.46 & 0.56 & 0.97 & \cite{pahar2022covid} \\
Coswara & Cough COVID-19 (w/\,Symptoms) & 0.71 & 0.76 & 0.97 & \cite{pahar2022covid} \\
Coswara & Speech COVID-19 Detection & 0.47 & 0.58 & 0.87 & \cite{pahar2022covid} \\
CoughVID & Audio COVID-19 Class. & 0.50 & 0.56 & 0.91 & \cite{pahar2022covid} \\
CoughVID & Audio Health Status Class. & 0.50 & 0.57 & 0.91 & \cite{kumar2025cough} \\
GazeBase & Eye-tracking Task Class. & 0.31 & 0.46 & 0.82 & \cite{sadhu2025task} \\
GLOBEM & Behavioral Depression Pred. & 0.53 & 0.59 & 0.70 & \cite{bhattacharya2024imputation} \\
GrabMyo & EMG Gesture Classification & 0.29 & 0.48 & 0.90 & \cite{cansiz2025hierarchical} \\
GrabMyo & EMG Subject Identification & 0.55 & 0.61 & 0.79 & \cite{cansiz2025hierarchical} \\
Harespod & Alt.-Respiration Ranking & 0.20 & 0.31 & 0.96 & \cite{kim2025optimization} \\
PhyMER & Multimodal Emotion Class. & 0.38 & 0.43 & 0.70 & \cite{singh2025feel} \\
PhyMER & EDA Emotion Classification & 0.14 & 0.18 & 0.72 & \cite{singh2025feel} \\
SHHS & EOG REM/NREM Class. & 0.49 & 0.58 & 0.73 & \cite{yetton2016automatic} \\
SHHS & EEG Sleep Stage Class. & 0.41 & 0.54 & 0.81 & \cite{gunnarsdottir2018novel} \\
VitalDB & ABP Hypotension Prediction & 0.72 & 0.77 & 0.88 & \cite{shim2025machine} \\
VitalDB & PPG Hypotension Prediction & 0.50 & 0.55 & 0.88 & \cite{shim2025machine} \\

\midrule
\grayrow
\multicolumn{6}{l}{\textit{\textbf{Localization}}} \\
GazeBase & Gaze Fixation Localization & 0.36 & 0.40 & 0.87 & \cite{svaricek2025insight} \\
GLOBEM & Peak Stress Week Ident. & 0.31 & 0.40 & 0.48 & \cite{booth2022toward} \\
PAMAP2 & Activity Segment Localization & 0.17 & 0.25 & 0.99 & \cite{bollampally2024optimizing} \\

\midrule
\grayrow
\multicolumn{6}{l}{\textit{\textbf{Cross-visit Comparison}}} \\
SHHS & Longitudinal BMI Prediction & 0.49 & 0.55 & 0.74 & \cite{li2021neural} \\

\bottomrule[1.5pt]
\end{tabular*}
\label{tab:sota_benchmarking_summary}
\end{table*}

%% file: tables/table_Appendix_all_task_breakdown.tex
\newcommand{\mhead}[1]{\multicolumn{1}{>{\centering\arraybackslash}p{1.1cm}}{#1}}
\newcolumntype{L}[1]{>{\raggedright\arraybackslash}p{#1}}
\newcolumntype{C}[1]{>{\centering\arraybackslash}p{#1}}

\begin{landscape}
    \scriptsize
    \setlength{\tabcolsep}{1pt}
    \setlength{\LTleft}{0pt}
    \setlength{\LTright}{0pt}
    \begin{longtable}{ L{1.5cm} L{3cm} *{17}{C{1.0cm}} }
 
    \caption{\textbf{Performance across datasets and tasks of all LLMs}}
    \label{tab:task_breakdown} \\
    
    \toprule[1.5pt]
    \textbf{Dataset} & \textbf{Task} & {\tiny\textbf{Naive Baseline}} & {\tiny\textbf{\claudeHaiku}} & {\tiny\textbf{\deepseek}} & {\tiny\textbf{\geminiFlash}} & {\tiny\textbf{\geminiPro}} & {\tiny\textbf{\geminiProthree}} & {\tiny\textbf{\glm}} & {\tiny\textbf{\glmfive}} & {\tiny\textbf{\gptFourOneMini}} & {\tiny\textbf{\gptFiveMini}} & {\tiny\textbf{\grok}} & {\tiny\textbf{\kimi}} & {\tiny\textbf{\llama}} & {\tiny\textbf{\minimax}} & {\tiny\textbf{\nemotron}} & {\tiny\textbf{\qwen}} & {\tiny\textbf{\qwenCoder}} \\
    \midrule\midrule
    \endfirsthead
    
    \toprule[1.5pt]
    \textbf{Dataset} & \textbf{Task} & {\tiny\textbf{Naive Baseline}} & {\tiny\textbf{\claudeHaiku}} & {\tiny\textbf{\deepseek}} & {\tiny\textbf{\geminiFlash}} & {\tiny\textbf{\geminiPro}} & {\tiny\textbf{\geminiProthree}} & {\tiny\textbf{\glm}} & {\tiny\textbf{\glmfive}} & {\tiny\textbf{\gptFourOneMini}} & {\tiny\textbf{\gptFiveMini}} & {\tiny\textbf{\grok}} & {\tiny\textbf{\kimi}} & {\tiny\textbf{\llama}} & {\tiny\textbf{\minimax}} & {\tiny\textbf{\nemotron}} & {\tiny\textbf{\qwen}} & {\tiny\textbf{\qwenCoder}} \\
    \midrule\midrule
    \endhead
 
    \bottomrule[1.5pt]
    \multicolumn{19}{r}{\emph{Continued on next page}} \\
    \endfoot
 
    \bottomrule[1.5pt]
    \endlastfoot
    Bridge2AI-voice & Articulation Rate Calculation & 1.00 & 0.79 & 0.75 & 0.69 & 0.83 & 0.81 & 0.80 & 0.86 & 0.67 & 0.73 & 0.84 & 0.80 & 0.47 & 0.78 & 0.55 & 0.79 & 0.81 \\
    Bridge2AI-voice & Cross-task Voice Comparison & 0.25 & 0.32 & 0.36 & 0.27 & 0.26 & 0.43 & 0.30 & 0.33 & 0.33 & 0.39 & 0.29 & 0.35 & 0.38 & 0.32 & 0.28 & 0.28 & 0.32 \\
    Bridge2AI-voice & F0 Range Extraction & 1.00 & 0.75 & 0.80 & 0.82 & 0.82 & 0.80 & 0.83 & 0.78 & 0.66 & 0.67 & 0.82 & 0.78 & 0.65 & 0.83 & 0.80 & 0.79 & 0.82 \\
    Bridge2AI-voice & Harmonics to Noise Ratio (HNR) Extraction & 1.00 & 0.52 & 0.48 & 0.57 & 0.53 & 0.69 & 0.58 & 0.72 & 0.35 & 0.34 & 0.50 & 0.61 & 0.30 & 0.60 & 0.59 & 0.47 & 0.66 \\
    Bridge2AI-voice & Jitter Extraction & 1.00 & 0.17 & 0.27 & 0.19 & 0.11 & 0.06 & 0.27 & 0.26 & 0.16 & 0.06 & 0.21 & 0.39 & 0.10 & 0.08 & 0.19 & 0.32 & 0.27 \\
    Bridge2AI-voice & Parkinsons Diagnosis & 0.50 & 0.52 & 0.56 & 0.53 & 0.49 & 0.51 & 0.51 & 0.49 & 0.56 & 0.48 & 0.51 & 0.50 & 0.52 & 0.51 & 0.54 & 0.51 & 0.48 \\
    Bridge2AI-voice & Shimmer Extraction & 1.00 & 0.58 & 0.62 & 0.55 & 0.65 & 0.75 & 0.63 & 0.70 & 0.63 & 0.45 & 0.73 & 0.61 & 0.54 & 0.58 & 0.38 & 0.49 & 0.62 \\
    Bridge2AI-voice & Reversed Signal Detection & 0.50 & 0.57 & 0.83 & 0.60 & 0.75 & 0.63 & 0.64 & 0.77 & 0.54 & 0.57 & 0.64 & 0.66 & 0.48 & 0.48 & 0.49 & 0.67 & 0.54 \\
    Capture24 & Activity Classification & 0.07 & 0.10 & 0.10 & 0.08 & 0.13 & 0.13 & 0.14 & 0.14 & 0.14 & 0.10 & 0.14 & 0.17 & 0.11 & 0.13 & 0.05 & 0.14 & 0.12 \\
    Capture24 & Activity Transition Recognition & 0.02 & 0.00 & 0.00 & 0.00 & 0.00 & 0.00 & 0.00 & 0.00 & 0.00 & 0.00 & 0.00 & 0.00 & 0.00 & 0.00 & 0.00 & 0.00 & 0.00 \\
    Capture24 & 1-axis Signal Imputation & 0.73 & 0.78 & 0.83 & 0.78 & 0.85 & 0.84 & 0.81 & 0.83 & 0.83 & 0.81 & 0.84 & 0.84 & 0.80 & 0.79 & 0.59 & 0.80 & 0.83 \\
    Capture24 & Step Count Calculation & 1.00 & 0.93 & 0.89 & 0.73 & 0.85 & 0.95 & 0.93 & 0.94 & 0.92 & 0.96 & 0.92 & 0.92 & 0.79 & 0.86 & 0.56 & 0.82 & 0.91 \\
    Capture24 & 3-axis Signal Forecasting & 0.72 & 0.82 & 0.78 & 0.83 & 0.80 & 0.81 & 0.81 & 0.81 & 0.76 & 0.79 & 0.81 & 0.81 & 0.82 & 0.76 & 0.73 & 0.80 & 0.83 \\
    Capture24 & 3-axis Signal Imputation & 0.73 & 0.70 & 0.63 & 0.82 & 0.67 & 0.80 & 0.79 & 0.83 & 0.76 & 0.71 & 0.77 & 0.83 & 0.76 & 0.76 & 0.58 & 0.68 & 0.75 \\
    Capture24 & Day and Night Signal Ordering & 0.50 & 0.73 & 0.89 & 0.79 & 0.83 & 0.94 & 0.88 & 0.95 & 0.82 & 0.98 & 0.91 & 0.83 & 0.60 & 0.68 & 0.46 & 0.91 & 0.74 \\
    CGMacros & A1c Prediction & 0.33 & 0.64 & 0.62 & 0.63 & 0.57 & 0.67 & 0.61 & 0.58 & 0.57 & 0.53 & 0.65 & 0.54 & 0.53 & 0.44 & 0.29 & 0.49 & 0.47 \\
    CGMacros & CGM Time in Range Calculation & 1.00 & 1.00 & 1.00 & 0.95 & 0.99 & 1.00 & 1.00 & 1.00 & 0.97 & 0.91 & 0.98 & 1.00 & 0.99 & 0.87 & 0.54 & 0.88 & 0.99 \\
    CGMacros & Fasting GLU Prediction & 0.90 & 0.78 & 0.74 & 0.32 & 0.50 & 0.75 & 0.76 & 0.77 & 0.78 & 0.78 & 0.78 & 0.72 & 0.65 & 0.72 & 0.69 & 0.78 & 0.76 \\
    CGMacros & Postprandial CGM iAUC Calculation & 1.00 & 0.93 & 0.89 & 0.94 & 0.87 & 0.99 & 0.95 & 0.92 & 0.99 & 0.99 & 0.96 & 0.95 & 0.79 & 0.88 & 0.33 & 0.93 & 0.83 \\
    CGMacros & CGM Forecasting with Histroy CGM and Meal Info & 0.67 & 0.55 & 0.54 & 0.54 & 0.55 & 0.58 & 0.57 & 0.54 & 0.49 & 0.52 & 0.58 & 0.59 & 0.41 & 0.55 & 0.42 & 0.54 & 0.53 \\
    CGMacros & CGM Forecasting with Current Meal Information and Histroy CGM and Meal Information & 0.66 & 0.52 & 0.48 & 0.39 & 0.57 & 0.61 & 0.58 & 0.54 & 0.52 & 0.53 & 0.51 & 0.54 & 0.33 & 0.51 & 0.45 & 0.55 & 0.54 \\
    CGMacros & CGM Forecasting with Meal Information & 0.66 & 0.52 & 0.52 & 0.40 & 0.54 & 0.61 & 0.61 & 0.55 & 0.54 & 0.46 & 0.54 & 0.54 & 0.45 & 0.51 & 0.30 & 0.47 & 0.49 \\
    CGMacros & CGM Forecasting & 0.67 & 0.47 & 0.50 & 0.54 & 0.49 & 0.66 & 0.56 & 0.53 & 0.48 & 0.49 & 0.49 & 0.54 & 0.41 & 0.51 & 0.51 & 0.53 & 0.52 \\
    CGMacros & Meal Image Classification from CGM & 0.25 & 0.29 & - & 0.10 & 0.25 & 0.13 & - & 0.18 & 0.26 & 0.27 & 0.26 & - & 0.29 & - & 0.23 & - & - \\
    CGMacros & Postprandial CGM Response Comparison & 0.50 & 0.82 & 0.84 & 0.89 & 0.81 & 0.89 & 0.88 & 0.88 & 0.89 & 0.84 & 0.87 & 0.87 & 0.85 & 0.78 & 0.54 & 0.83 & 0.85 \\
    CGMacros & Meal Time Localization & 0.71 & 0.63 & 0.64 & 0.60 & 0.61 & 0.68 & 0.83 & 0.59 & 0.68 & 0.63 & 0.67 & 0.69 & 0.62 & 0.64 & 0.62 & 0.63 & 0.67 \\
    CGMacros & CGM imputation with Activity Calories Trajectory & 0.72 & 0.72 & 0.55 & 0.53 & 0.45 & 0.74 & 0.68 & 0.75 & 0.63 & 0.55 & 0.67 & 0.71 & 0.42 & 0.69 & 0.40 & 0.58 & 0.71 \\
    CGMacros & CGM imputation & 0.71 & 0.98 & 0.98 & 0.99 & 0.99 & 0.99 & 0.99 & 0.98 & 0.98 & 0.98 & 0.98 & 0.98 & 0.93 & 0.98 & 0.59 & 0.98 & 0.98 \\
    CGMacros & CGM Imputation with HR Trajectory & 0.70 & 0.98 & 0.95 & 0.94 & 0.94 & 0.97 & 0.97 & 0.98 & 0.95 & 0.97 & 0.97 & 0.96 & 0.96 & 0.96 & 0.71 & 0.95 & 0.98 \\
    Coswara & Audio Type Classification & 0.33 & 0.32 & 0.29 & 0.36 & 0.00 & 0.79 & 0.45 & 0.39 & 0.36 & 0.33 & 0.41 & 0.35 & 0.28 & 0.39 & 0.33 & 0.33 & 0.39 \\
    Coswara & Diagnosis Classification with Cough Audio & 0.50 & 0.47 & 0.53 & 0.53 & 0.00 & 0.33 & 0.56 & 0.49 & 0.47 & 0.47 & 0.48 & 0.47 & 0.52 & 0.51 & 0.46 & 0.51 & 0.52 \\
    Coswara & Diagnosis Classification with Symptom Information Only & 0.50 & 0.76 & 0.78 & 0.76 & 0.81 & 0.78 & 0.79 & 0.74 & 0.76 & 0.78 & 0.79 & 0.81 & 0.78 & 0.70 & 0.80 & 0.74 & 0.80 \\
    Coswara & Diagnosis Classification with Cough Audio and Symptom Information & 0.50 & 0.66 & 0.74 & 0.72 & 0.70 & 0.83 & 0.73 & 0.72 & 0.74 & 0.70 & 0.74 & 0.69 & 0.76 & 0.68 & 0.74 & 0.70 & 0.71 \\
    Coswara & Diagnosis Classification with Speech Audio & 0.50 & 0.58 & 0.47 & 0.49 & 0.00 & 0.51 & 0.49 & 0.50 & 0.52 & 0.48 & 0.54 & 0.51 & 0.48 & 0.52 & 0.54 & 0.47 & 0.53 \\
    Cough VID & Cough Detection wtih Good Quality Samples & 0.50 & 0.50 & 0.54 & 0.57 & 0.55 & 0.59 & 0.59 & 0.61 & 0.54 & 0.54 & 0.58 & 0.46 & 0.60 & 0.56 & 0.46 & 0.49 & 0.55 \\
    Cough VID & Cough Detection with Poor Quality Samples & 0.50 & 0.55 & 0.65 & 0.62 & 0.00 & 0.60 & 0.57 & 0.59 & 0.50 & 0.55 & 0.41 & 0.73 & 0.73 & 0.50 & 0.57 & 0.42 & 0.38 \\
    Cough VID & COVID Status Classification & 0.50 & 0.50 & 0.45 & 0.54 & 0.44 & 0.51 & 0.54 & 0.51 & 0.52 & 0.48 & 0.49 & 0.54 & 0.51 & 0.46 & 0.47 & 0.53 & 0.56 \\
    Cough VID & Diagnosis Classification & 0.20 & 0.20 & 0.18 & 0.17 & 0.10 & 0.07 & 0.20 & 0.18 & 0.23 & 0.11 & 0.20 & 0.19 & 0.22 & 0.19 & 0.13 & 0.23 & 0.17 \\
    Cough VID & Health Status Classification & 0.50 & 0.48 & 0.55 & 0.50 & 0.49 & 0.57 & 0.47 & 0.45 & 0.57 & 0.51 & 0.51 & 0.48 & 0.57 & 0.50 & 0.46 & 0.46 & 0.49 \\
    Cough VID & MFCC Mean \& STD Calculation & 1.00 & 0.99 & 0.99 & 1.00 & 0.79 & 1.00 & 0.97 & 0.99 & 0.99 & 1.00 & 0.73 & 0.88 & 0.98 & 0.94 & 0.42 & 0.98 & 1.00 \\
    GazeBase & Fixation Point Localization & 0.35 & 0.36 & 0.37 & 0.38 & 0.39 & 0.37 & 0.36 & 0.38 & 0.37 & 0.37 & 0.20 & 0.37 & 0.38 & 0.38 & 0.40 & 0.37 & 0.38 \\
    GazeBase & Horizontal Saccade Track Forecasting & 0.79 & 0.78 & 0.79 & 0.80 & 0.83 & 0.90 & 0.82 & 0.84 & 0.81 & 0.84 & 0.84 & 0.81 & 0.76 & 0.80 & 0.50 & 0.80 & 0.80 \\
    GazeBase & Text Reading Imputation & 0.86 & 0.83 & 0.84 & 0.85 & 0.78 & 0.83 & 0.84 & 0.84 & 0.82 & 0.84 & 0.85 & 0.83 & 0.85 & 0.84 & 0.91 & 0.84 & 0.84 \\
    GazeBase & Reading Sequence Recognition & 0.17 & 0.29 & 0.21 & 0.10 & 0.03 & 0.19 & 0.40 & 0.31 & 0.34 & 0.09 & 0.23 & 0.38 & 0.23 & 0.20 & 0.17 & 0.17 & 0.29 \\
    GazeBase & Eye-tracking Task Classification & 0.17 & 0.24 & 0.24 & 0.24 & 0.46 & 0.33 & 0.40 & 0.38 & 0.42 & 0.29 & 0.45 & 0.35 & 0.22 & 0.28 & 0.18 & 0.30 & 0.30 \\
    GLOBEM & Circadian Routine Comparison & 0.50 & 0.52 & 0.57 & 0.61 & 0.60 & 0.48 & 0.55 & 0.53 & 0.63 & 0.53 & 0.57 & 0.55 & 0.65 & 0.54 & 0.59 & 0.55 & 0.64 \\
    GLOBEM & COVID Year Recognition & 0.50 & 0.96 & 0.94 & 0.93 & 0.90 & 0.60 & 0.92 & 0.92 & 0.94 & 0.90 & 0.93 & 0.93 & 0.95 & 0.88 & 0.49 & 0.92 & 0.92 \\
    GLOBEM & Depression Trajectory Classification & 0.50 & 0.54 & 0.48 & 0.51 & 0.54 & 0.53 & 0.59 & 0.55 & 0.56 & 0.55 & 0.49 & 0.55 & 0.50 & 0.51 & 0.52 & 0.53 & 0.51 \\
    GLOBEM & Location Entropy Extraction & 1.00 & 0.80 & 0.72 & 0.62 & 0.68 & 0.74 & 0.74 & 0.78 & 0.80 & 0.76 & 0.83 & 0.72 & 0.78 & 0.79 & 0.53 & 0.69 & 0.77 \\
    GLOBEM & Peak Stress Week Localization & 0.10 & 0.28 & 0.29 & 0.26 & 0.31 & 0.33 & 0.30 & 0.32 & 0.30 & 0.31 & 0.30 & 0.30 & 0.30 & 0.38 & 0.40 & 0.29 & 0.31 \\
    GLOBEM & Step Count Forecasting & 0.68 & 0.78 & 0.77 & 0.76 & 0.76 & 0.78 & 0.78 & 0.79 & 0.76 & 0.74 & 0.76 & 0.75 & 0.78 & 0.78 & 0.65 & 0.77 & 0.78 \\
    GrabMyo & Gesture Classification with Reference & 0.06 & 0.46 & 0.34 & 0.38 & 0.42 & 0.56 & 0.48 & 0.52 & 0.11 & 0.24 & 0.32 & 0.39 & 0.14 & 0.35 & 0.06 & 0.10 & 0.32 \\
    GrabMyo & Subject Identification & 0.50 & 0.55 & 0.58 & 0.60 & 0.61 & 0.60 & 0.56 & 0.62 & 0.52 & 0.48 & 0.60 & 0.55 & 0.47 & 0.55 & 0.57 & 0.56 & 0.56 \\
    Harespod & Altitude Ranking with Respiration & 0.17 & 0.20 & 0.28 & 0.15 & 0.22 & 0.13 & 0.23 & 0.20 & 0.31 & 0.14 & 0.14 & 0.21 & 0.23 & 0.25 & 0.00 & 0.18 & 0.22 \\
    Harespod & Altitude Ranking with SpO2 & 0.17 & 0.32 & 0.32 & 0.33 & 0.37 & 0.39 & 0.33 & 0.37 & 0.33 & 0.31 & 0.32 & 0.33 & 0.27 & 0.37 & 0.00 & 0.26 & 0.38 \\
    Harespod & Respiration and HR Pairing & 0.50 & 0.60 & 0.53 & 0.46 & 0.58 & 0.57 & 0.56 & 0.67 & 0.69 & 0.61 & 0.65 & 0.61 & 0.56 & 0.69 & 0.50 & 0.55 & 0.60 \\
    Harespod & Respiration and SpO2 Pairing & 0.50 & 0.72 & 0.50 & 0.49 & 0.48 & 0.61 & 0.60 & 0.75 & 0.41 & 0.74 & 0.66 & 0.56 & 0.52 & 0.68 & 0.51 & 0.46 & 0.59 \\
    PAMAP2 & Activity Localization & 0.15 & 0.23 & 0.22 & 0.18 & 0.19 & 0.79 & 0.25 & 0.27 & 0.15 & 0.08 & 0.21 & 0.20 & 0.11 & 0.18 & 0.03 & 0.17 & 0.21 \\
    PERG-IOBA & Eye Disease Type Classification with Patient's Meta Information & 0.25 & 0.28 & 0.32 & 0.44 & 0.27 & 0.50 & 0.34 & 0.36 & 0.40 & 0.35 & 0.31 & 0.39 & 0.38 & 0.30 & 0.35 & 0.42 & 0.34 \\
    PERG-IOBA & Eye Disease Type Classification & 0.25 & 0.24 & 0.26 & 0.35 & 0.29 & 0.34 & 0.37 & 0.39 & 0.40 & 0.34 & 0.30 & 0.40 & 0.31 & 0.37 & 0.30 & 0.46 & 0.37 \\
    PERG-IOBA & Disease Differentiation: between Macular Disease \& Optic Nerve Disease \& Normal & 0.50 & 0.38 & 0.54 & 0.55 & 0.52 & 0.61 & 0.53 & 0.54 & 0.58 & 0.56 & 0.53 & 0.52 & 0.56 & 0.56 & 0.57 & 0.56 & 0.49 \\
    PERG-IOBA & Eye Health Status Classification & 0.50 & 0.54 & 0.55 & 0.62 & 0.60 & 0.57 & 0.57 & 0.54 & 0.56 & 0.61 & 0.61 & 0.54 & 0.57 & 0.62 & 0.57 & 0.59 & 0.58 \\
    PERG-IOBA & N35, P50, N95 Feature Extraction & 1.00 & 1.00 & 1.00 & 0.88 & 0.85 & 0.99 & 1.00 & 1.00 & 1.00 & 0.98 & 0.92 & 1.00 & 1.00 & 0.96 & 0.54 & 1.00 & 1.00 \\
    PhyMER & Cross Subject Arousal Ranking & 0.50 & 0.50 & 0.48 & 0.51 & 0.52 & 0.53 & 0.54 & 0.53 & 0.51 & 0.56 & 0.51 & 0.58 & 0.50 & 0.44 & 0.60 & 0.51 & 0.54 \\
    PhyMER & Conditional Forecasting & 0.21 & 0.54 & 0.39 & 0.38 & 0.50 & 0.57 & 0.58 & 0.55 & 0.40 & 0.40 & 0.47 & 0.50 & 0.42 & 0.43 & 0.07 & 0.45 & 0.49 \\
    PhyMER & Conditional Imputation & 0.23 & 0.62 & 0.54 & 0.60 & 0.62 & 0.69 & 0.68 & 0.66 & 0.56 & 0.57 & 0.63 & 0.58 & 0.33 & 0.56 & 0.26 & 0.52 & 0.59 \\
    PhyMER & Cross-channel Translation & 0.43 & 0.72 & 0.70 & 0.69 & 0.78 & 0.77 & 0.80 & 0.80 & 0.68 & 0.79 & 0.79 & 0.62 & 0.65 & 0.67 & 0.35 & 0.68 & 0.77 \\
    PhyMER & Emotion Type Classification & 0.14 & 0.38 & 0.39 & 0.35 & 0.34 & 0.38 & 0.41 & 0.35 & 0.42 & 0.43 & 0.41 & 0.37 & 0.30 & 0.35 & 0.29 & 0.43 & 0.41 \\
    PhyMER & Emotion Type Classification with Only EDA Input & 0.14 & 0.10 & 0.16 & 0.11 & 0.12 & 0.19 & 0.17 & 0.14 & 0.18 & 0.14 & 0.15 & 0.13 & 0.17 & 0.13 & 0.13 & 0.14 & 0.16 \\
    PhyMER & Inter-subject Emotion Recognition & 0.50 & 0.52 & 0.54 & 0.55 & 0.55 & 0.54 & 0.52 & 0.60 & 0.61 & 0.63 & 0.55 & 0.63 & 0.55 & 0.46 & 0.73 & 0.60 & 0.56 \\
    PhyMER & Personality Analysis & 0.25 & 0.27 & 0.30 & 0.20 & 0.19 & 0.23 & 0.28 & 0.24 & 0.24 & 0.24 & 0.21 & 0.19 & 0.23 & 0.16 & 0.21 & 0.31 & 0.27 \\
    PhyMER & Single-channel Forecasting & 0.21 & 0.56 & 0.56 & 0.34 & 0.56 & 0.58 & 0.59 & 0.59 & 0.50 & 0.54 & 0.56 & 0.62 & 0.43 & 0.53 & 0.35 & 0.50 & 0.60 \\
    PhyMER & Single-channel Imputation & 0.23 & 0.69 & 0.73 & 0.72 & 0.72 & 0.76 & 0.73 & 0.71 & 0.72 & 0.72 & 0.73 & 0.73 & 0.70 & 0.74 & 0.40 & 0.73 & 0.72 \\
    Shanghai Diabetes & Diabetes Type Classification & 0.50 & 0.60 & 0.61 & 0.54 & 0.78 & 0.60 & 0.52 & 0.72 & 0.69 & 0.62 & 0.62 & 0.73 & 0.75 & 0.78 & 0.54 & 0.59 & 0.61 \\
    Shanghai Diabetes & Cross-Subject Diabetes Type Comparison & 0.50 & 0.76 & 0.87 & 0.87 & 0.88 & 0.89 & 0.82 & 0.85 & 0.86 & 0.81 & 0.82 & 0.80 & 0.81 & 0.76 & 0.48 & 0.85 & 0.81 \\
    SHHS & Atrial Fibrillation (AF) Classification & 0.50 & 0.54 & 0.55 & 0.57 & 0.62 & 0.60 & 0.62 & 0.55 & 0.48 & 0.57 & 0.68 & 0.55 & 0.53 & 0.53 & 0.53 & 0.53 & 0.55 \\
    SHHS & Sleep AHI Calculation & 1.00 & 0.92 & 0.87 & 0.85 & 0.37 & 0.96 & 0.91 & 0.93 & 0.89 & 0.83 & 0.97 & 0.89 & 0.38 & 0.88 & 0.31 & 0.84 & 0.89 \\
    SHHS & Arousal Detection (EEG) & 0.00 & 0.26 & 0.19 & 0.09 & 0.15 & 0.39 & 0.27 & 0.34 & 0.16 & 0.11 & 0.26 & 0.30 & 0.10 & 0.19 & 0.10 & 0.14 & 0.20 \\
    SHHS & Arousal Detection (EOG) & 0.00 & 0.20 & 0.12 & 0.10 & 0.09 & 0.29 & 0.18 & 0.27 & 0.15 & 0.11 & 0.20 & 0.21 & 0.10 & 0.20 & 0.09 & 0.12 & 0.16 \\
    SHHS & Bandpower Calculation & 1.00 & 1.00 & 1.00 & 0.92 & 0.99 & 1.00 & 0.99 & 1.00 & 1.00 & 0.98 & 1.00 & 0.98 & 0.99 & 0.97 & 0.43 & 0.97 & 1.00 \\
    SHHS & BMI Comparison Between Visit & 0.50 & 0.50 & 0.53 & 0.46 & 0.45 & 0.39 & 0.47 & 0.52 & 0.55 & 0.48 & 0.45 & 0.43 & 0.49 & 0.55 & 0.54 & 0.48 & 0.47 \\
    SHHS & Conditional Forecasting & 0.94 & 0.91 & 0.91 & 0.92 & 0.93 & 0.92 & 0.91 & 0.91 & 0.93 & 0.90 & 0.91 & 0.91 & 0.92 & 0.91 & 0.91 & 0.92 & 0.92 \\
    SHHS & Conditional Imputation & 0.93 & 0.88 & 0.90 & 0.77 & 0.92 & 0.90 & 0.89 & 0.91 & 0.90 & 0.90 & 0.90 & 0.89 & 0.87 & 0.90 & 0.76 & 0.91 & 0.89 \\
    SHHS & Cross-channel Translation (ECG to HR) & 0.38 & 0.53 & 0.58 & 0.61 & 0.71 & 0.74 & 0.56 & 0.56 & 0.37 & 0.58 & 0.69 & 0.66 & 0.54 & 0.59 & 0.22 & 0.59 & 0.66 \\
    SHHS & Cross-channel Translation (EEG-C3A2 to EEG-C4A1) & 0.92 & 0.90 & 0.90 & 0.92 & 0.92 & 0.92 & 0.91 & 0.90 & 0.91 & 0.91 & 0.91 & 0.91 & 0.90 & 0.91 & 0.90 & 0.90 & 0.90 \\
    SHHS & Cardiovascular Disease (CVD) Death Prediction & 0.50 & 0.45 & 0.44 & 0.52 & 0.54 & 0.53 & 0.46 & 0.49 & 0.53 & 0.51 & 0.47 & 0.46 & 0.52 & 0.40 & 0.51 & 0.50 & 0.52 \\
    SHHS & Hypopnea Detection & 0.02 & 0.31 & 0.23 & 0.23 & 0.12 & 0.30 & 0.31 & 0.36 & 0.14 & 0.09 & 0.22 & 0.30 & 0.06 & 0.28 & 0.11 & 0.12 & 0.21 \\
    SHHS & REM Latency Calculation & 1.00 & 0.92 & 0.92 & 0.90 & 0.91 & 0.93 & 0.91 & 0.91 & 0.92 & 0.92 & 0.92 & 0.88 & 0.93 & 0.89 & 0.59 & 0.92 & 0.92 \\
    SHHS & REM/NREM Classification & 0.50 & 0.51 & 0.52 & 0.40 & 0.56 & 0.30 & 0.53 & 0.47 & 0.58 & 0.54 & 0.52 & 0.51 & 0.49 & 0.41 & 0.35 & 0.52 & 0.48 \\
    SHHS & Single-channel Forecasting & 0.93 & 0.90 & 0.91 & 0.91 & 0.91 & 0.91 & 0.91 & 0.91 & 0.92 & 0.89 & 0.91 & 0.90 & 0.91 & 0.91 & 0.86 & 0.91 & 0.91 \\
    SHHS & Single-channel Imputation & 0.93 & 0.86 & 0.70 & 0.74 & 0.92 & 0.91 & 0.87 & 0.91 & 0.73 & 0.89 & 0.90 & 0.86 & 0.87 & 0.88 & 0.92 & 0.86 & 0.87 \\
    SHHS & Sleep Efficiency Calculation & 1.00 & 1.00 & 0.99 & 0.95 & 0.99 & 1.00 & 0.94 & 0.88 & 1.00 & 1.00 & 0.99 & 0.97 & 0.87 & 0.02 & 0.67 & 0.94 & 0.96 \\
    SHHS & Smoker Classification & 0.50 & 0.53 & 0.48 & 0.53 & 0.50 & 0.53 & 0.52 & 0.50 & 0.52 & 0.56 & 0.56 & 0.51 & 0.47 & 0.55 & 0.48 & 0.49 & 0.54 \\
    SHHS & Sleep Stage Classification & 0.25 & 0.45 & 0.45 & 0.45 & 0.42 & 0.17 & 0.42 & 0.24 & 0.43 & 0.41 & 0.45 & 0.41 & 0.28 & 0.54 & 0.29 & 0.39 & 0.42 \\
    SHHS & Sleep Stage Transition Recognition & 0.08 & 0.11 & 0.23 & 0.43 & 0.84 & 0.39 & 0.31 & 0.22 & 0.09 & 0.10 & 0.20 & 0.28 & 0.08 & 0.16 & 0.16 & 0.13 & 0.09 \\
    SHHS & Stroke Prediction & 0.50 & 0.47 & 0.55 & 0.47 & 0.47 & 0.62 & 0.52 & 0.51 & 0.49 & 0.48 & 0.52 & 0.53 & 0.53 & 0.49 & 0.53 & 0.51 & 0.54 \\
    SHHS & Episode Level Ordering & 0.50 & 0.52 & 0.55 & 0.53 & 0.52 & 0.94 & 0.54 & 0.51 & 0.50 & 0.48 & 0.55 & 0.48 & 0.53 & 0.46 & 0.51 & 0.54 & 0.47 \\
    SHHS & Half Night Level Ordering & 0.50 & 0.70 & 0.76 & 0.83 & 0.85 & 0.83 & 0.67 & 0.74 & 0.73 & 0.74 & 0.76 & 0.75 & 0.60 & 0.67 & 0.52 & 0.81 & 0.57 \\
    SHHS & Visit Level Ordering & 0.50 & 0.53 & 0.56 & 0.68 & 0.69 & 0.67 & 0.56 & 0.59 & 0.52 & 0.53 & 0.55 & 0.55 & 0.56 & 0.54 & 0.48 & 0.59 & 0.56 \\
    VCTK & Reversed Signal Detection & 0.50 & 0.49 & 0.50 & 0.67 & 0.47 & 0.67 & 0.48 & 0.89 & 0.47 & 0.59 & 0.55 & 0.58 & 0.49 & 0.50 & 0.51 & 0.58 & 0.43 \\
    VitalDB & Anesthesia Range & 0.33 & 0.30 & 0.23 & 0.33 & 0.70 & 0.86 & 0.56 & 0.68 & 0.36 & 0.16 & 0.66 & 0.54 & 0.75 & 0.21 & 0.04 & 0.44 & 0.40 \\
    VitalDB & MINS (myocardial injury after non-cardiac surgery) Prediction with MBP & 0.50 & 0.60 & 0.47 & 0.55 & 0.53 & 0.50 & 0.52 & 0.53 & 0.51 & 0.51 & 0.54 & 0.54 & 0.52 & 0.54 & 0.49 & 0.49 & 0.55 \\
    VitalDB & Drug Infusion Series to Depth of Anesthesia (BIS) Translation & 0.72 & 0.41 & 0.46 & 0.48 & 0.40 & 0.61 & 0.54 & 0.52 & 0.52 & 0.49 & 0.53 & 0.58 & 0.16 & 0.58 & 0.36 & 0.36 & 0.55 \\
    VitalDB & PPG and ECG to Blood Pressure Translation & 0.40 & 0.54 & 0.48 & 0.50 & 0.51 & 0.50 & 0.51 & 0.50 & 0.52 & 0.51 & 0.51 & 0.53 & 0.52 & 0.52 & 0.34 & 0.49 & 0.52 \\
    VitalDB & EEG to Bispectral Index (BIS) Translation & 0.70 & 0.50 & 0.48 & 0.46 & 0.59 & 0.60 & 0.53 & 0.61 & 0.44 & 0.55 & 0.35 & 0.48 & 0.47 & 0.58 & 0.46 & 0.52 & 0.58 \\
    VitalDB & MBP Forecasting & 0.29 & 0.58 & 0.54 & 0.53 & 0.52 & 0.61 & 0.54 & 0.58 & 0.57 & 0.55 & 0.58 & 0.58 & 0.54 & 0.55 & 0.33 & 0.58 & 0.57 \\
    VitalDB & Hypotension Event Classification with MBP & 0.50 & 0.73 & 0.72 & 0.73 & 0.68 & 0.80 & 0.75 & 0.74 & 0.77 & 0.72 & 0.74 & 0.75 & 0.70 & 0.77 & 0.55 & 0.74 & 0.75 \\
    VitalDB & MBP Forecasting with Infusion Info & 0.27 & 0.55 & 0.54 & 0.46 & 0.52 & 0.60 & 0.55 & 0.56 & 0.54 & 0.53 & 0.58 & 0.57 & 0.55 & 0.54 & 0.33 & 0.54 & 0.57 \\
    VitalDB & Hypotension Event Classification with PPG & 0.50 & 0.48 & 0.47 & 0.49 & 0.51 & 0.50 & 0.55 & 0.52 & 0.47 & 0.49 & 0.55 & 0.48 & 0.52 & 0.50 & 0.51 & 0.48 & 0.48 \\
    VitalDB & Mean arterial pressure (MBP) Time in Range 70-100 & 1.00 & 1.00 & 1.00 & 1.00 & 1.00 & 1.00 & 1.00 & 1.00 & 1.00 & 1.00 & 0.93 & 1.00 & 1.00 & 0.99 & 0.91 & 0.99 & 1.00 \\
    
    \end{longtable}
\end{landscape}

%% file: tables/table_Appendix_input_format_exp.tex
\begin{table}[ht]
\small
\centering
\caption{\textbf{Results breakdown for input format experiments.}}
\label{tab:input_format_exp}
\resizebox{\textwidth}{!}{%
\begin{tabular}{llcccccc}
\toprule[1.5pt]
\multirow{2}{*}{\textbf{Dataset}} & \multirow{2}{*}{\textbf{Task}} & \multicolumn{3}{c}{\raisebox{-0.25em}{\includegraphics[height=1.2em]{icons/grok.png}} \textbf{\grok}} & \multicolumn{3}{c}{\raisebox{-0.25em}{\includegraphics[height=1.2em]{icons/meta-color.png}} \textbf{\llama}} \\
\cmidrule(lr){3-5} \cmidrule(lr){6-8}
 & & \textbf{File} & \textbf{Text} & \textbf{Image} & \textbf{File} & \textbf{Text} & \textbf{Image} \\
\midrule\midrule
CGMacros & CGM Forecasting & 0.49 & 0.55 & 0.47 & 0.41 & 0.54 & 0.44 \\
 & CGM imputation & 0.98 & 0.98 & 0.73 & 0.93 & 0.99 & 0.96 \\
\midrule
GLOBEM & Circadian Routine Comparison & 0.57 & 0.57 & 0.49 & 0.65 & 0.53 & 0.57 \\
 & Depression Trajectory Classification & 0.50 & 0.50 & 0.50 & 0.50 & 0.54 & 0.53 \\
 & Location Entropy Extraction & 0.83 & 0.83 & 0.52 & 0.78 & 0.82 & 0.69 \\
 & Peak Stress Week Localization & 0.30 & 0.30 & 0.36 & 0.30 & 0.30 & 0.29 \\
 & Step Count Forecasting & 0.76 & 0.78 & 0.73 & 0.78 & 0.78 & 0.77 \\
\midrule
PERG\_IOBA & Eye Health Status Classification & 0.62 & 0.56 & 0.60 & 0.58 & 0.52 & 0.60 \\
\midrule
VitalDb & MBP Forecasting & 0.58 & 0.60 & 0.33 & 0.54 & 0.55 & 0.51 \\
 & Hypotension Event Classification with PPG & 0.74 & 0.72 & 0.68 & 0.71 & 0.70 & 0.71 \\
\bottomrule[1.5pt]
\end{tabular}%
}
\end{table}

%% file: reasoning_traces/codeact.tex
\begin{userbox}
You are provided with 3-axis accelerometer data from a wearable device. 
Your task is to classify the physical activity being performed during the recorded segment. 
The 14 possible activities are: sleeping, manual-work, sitting, heavy loads, leisure, sports/gym, standing, transportation, occupation, activity, vehicle, mixed-activity, walking, and household-chores.
The 3-axis data is provided as three separate arrays corresponding to the x, y, and z axes of the accelerometer in 'input/x.npy', 'input/y.npy', and 'input/z.npy'. The signals are sampled at 100 Hz.
Please analyze the signal and determine which activity it represents in json format. Your answer must be one of the 14 possible activities. Please output your final result in a json object without any other text:
\{
    "activity": "activity\_name"
\}
\par
\end{userbox}
\vspace{0.2cm}
\begin{agentbox}
\begin{thoughtbox}
I need to analyze 3-axis accelerometer data to classify physical activity. Let me first load the data from the three files and explore the signal characteristics to determine what activity is being performed.

The 14 possible activities are:
- sleeping
- manual-work
- sitting
- heavy loads
- leisure
- sports/gym
- standing
- transportation
- occupation
- activity
- vehicle
- mixed-activity
- walking
- household-chores

Let me start by loading the data and examining its properties.
\end{thoughtbox}
\begin{lstlisting}[language=Python]
import numpy as np
import json

# Load the accelerometer data
x = np.load('input/x.npy')
y = np.load('input/y.npy')
z = np.load('input/z.npy')

print(f"Shape of x: {x.shape}")
print(f"Shape of y: {y.shape}")
print(f"Shape of z: {z.shape}")
\end{lstlisting}
\begin{thoughtbox}
Observation: The file '/mnt/data/input/x.npy' doesn't exist.</think> The file path seems incorrect. Let me try the correct path 'input/x.npy' as specified in the task.
\end{thoughtbox}
\begin{lstlisting}[language=Python]
import numpy as np
import json

# Load the accelerometer data
x = np.load('input/x.npy')
y = np.load('input/y.npy')
z = np.load('input/z.npy')

print(f"Shape of x: {x.shape}")
print(f"Shape of y: {y.shape}")
print(f"Shape of z: {z.shape}")
print(f"Number of samples: {len(x)}")
print(f"Duration in seconds: {len(x)/100:.2f}")
\end{lstlisting}
\begin{thoughtbox}
Observation:
Shape of x: (6000,)
Shape of y: (6000,)
Shape of y: (6000,)
Shape of z: (6000,)
Shape of z: (6000,)
Shape of z: (6000,)
Shape of z: (6000,)
Shape of z: (6000,)
Shape of z: (6000,)
Shape of z: (6000,)
Shape of z: (6000,)
Shape of z: (6000,)
 activity classification. I'll calculate the signal magnitude and analyze its statistical features.</think> I have loaded the accelerometer data. There are 6000 samples at 100 Hz, so the recording duration is 60 seconds. Let me now analyze the signal to extract features that can help classify the activity. I'll calculate the signal magnitude and analyze its statistical characteristics.
\end{thoughtbox}
\begin{lstlisting}[language=Python]
# Calculate signal magnitude (acceleration magnitude)
magnitude = np.sqrt(x**2 + y**2 + z**2)

# Calculate basic statistics
print(f"Mean magnitude: {np.mean(magnitude):.4f}")
print(f"Std magnitude: {np.std(magnitude):.4f}")
print(f"Min magnitude: {np.min(magnitude):.4f}")
print(f"Max magnitude: {np.max(magnitude):.4f}")
print(f"Mean x: {np.mean(x):.4f}, Std x: {np.std(x):.4f}")
print(f"Mean y: {np.mean(y):.4f}, Std y: {np.std(y):.4f}")
print(f"Mean z: {np.mean(z):.4f}, Std z: {np.std(z):.4f}")
\end{lstlisting}
\begin{thoughtbox}
Observation:
The user wants me to classify the accelerometer data into one of 14 activity categories. I've already loaded the data and calculated basic statistics. Let me analyze this more carefully.

The data has:
- 6000 samples at 100 Hz = 60 seconds of recording
- Mean magnitude around 1.0 (suggesting gravity is present)
- Std of magnitude around 0.02
- x has mean close to 0, std around 0.014
- y has mean close to 0, std around 0.014
- z has mean around 1.0, std around 0.007

The z-axis having a mean around 1.0 suggests the device is mostly stationary with gravity aligned along the z-axis (horizontal orientation). The low standard deviations across all axes indicate very little movement, which strongly suggests a stationary activity like sleeping, sitting, or standing.

Let me analyze the signal more carefully by looking at:
1. The variance of the signal magnitude
2. Zero-crossing rate
3. Signal variability patterns

With such low std values, this is very likely a low-movement activity like sleeping, sitting, or standing.

I should provide my final classification based on this analysis.</think> I've loaded the accelerometer data which has 6000 samples at 100 Hz (60 seconds duration). The statistics show:
- Mean magnitude \textasciitilde{}1.0 (gravity is present)
- Very low standard deviations across all axes (0.007-0.014)
- z-axis mean is \textasciitilde{}1.0, suggesting the device is lying flat with gravity aligned
- x and y means close to 0

These characteristics strongly indicate a stationary, low-movement activity. The very low variability suggests this is likely "sleeping" since:
1. Sleeping typically has the lowest movement among all activities
2. The device would be relatively stationary
3. The gravitational alignment suggests lying down

Let me confirm by looking at more detailed signal characteristics.
\end{thoughtbox}
\begin{lstlisting}[language=Python]
# Calculate additional features
# Signal magnitude variance
mag_var = np.var(magnitude)
# Zero-crossing rate
zero_crossings = np.sum(np.diff(np.sign(magnitude - np.mean(magnitude))) != 0)
# Energy of the signal
energy = np.sum(magnitude**2)

print(f"Magnitude variance: {mag_var:.8f}")
print(f"Zero-crossing rate: {zero_crossings}")
print(f"Signal energy: {energy:.4f}")
print(f"Signal range (max-min): {np.max(magnitude) - np.min(magnitude):.6f}")

# Check for periodic patterns that might suggest breathing during sleep
# Look at frequency domain characteristics
import scipy.signal as signal
freqs, psd = signal.welch(magnitude, fs=100)
dominant_freq = freqs[np.argmax(psd)]
print(f"Dominant frequency: {dominant_freq:.4f} Hz")
\end{lstlisting}
\end{agentbox}
\vspace{0.2cm}
\begin{outputbox}
Shape of x: (14800,)
Shape of y: (14800,)
Shape of z: (14800,)
Shape of x: (14800,)
Shape of y: (14800,)
Shape of z: (14800,)
Number of samples: 14800
Duration in seconds: 148.00
Mean magnitude: 0.9934
Std magnitude: 0.2337
Min magnitude: 0.0514
Max magnitude: 5.1575
Mean x: -0.2258, Std x: 0.3409
Mean y: -0.7474, Std y: 0.2365
Mean z: -0.4546, Std z: 0.2301
Magnitude variance: 0.05462577
Zero-crossing rate: 1290
Signal energy: 15412.8530
Signal range (max-min): 5.106137
Dominant frequency: 1.9531 Hz
\par
\end{outputbox}
\vspace{0.2cm}
\begin{agentbox}
\begin{thoughtbox}
Let me analyze these signal characteristics:

1. **Duration**: 148 seconds at 100 Hz
2. **Mean magnitude**: 0.9934 g (confirms gravity is present)
3. **Std magnitude**: 0.2337 g (moderate movement variation)
4. **Max magnitude**: 5.1575 g (high peaks indicating vigorous movement)
5. **Signal range**: 5.106 g (wide range - significant intensity)
6. **Zero-crossing rate**: 1290 (frequent direction changes)
7. **Dominant frequency**: 1.9531 Hz (\textasciitilde{}117 cycles/minute)

The dominant frequency of \textasciitilde{}2 Hz (117 per minute) is very characteristic of:
- Walking cadence (typically 100-120 steps per minute)
- Aerobic exercise rhythms

The high magnitude peaks (up to 5.16g) combined with:
- Rhythmic pattern at 2 Hz
- Moderate overall variance
- Sustained energy over 148 seconds

This pattern suggests rhythmic, repetitive aerobic movement. The gravitational vector (negative in all axes) suggests a tilted orientation possibly during exercise.

Given:
- Rhythmic \textasciitilde{}2 Hz pattern (consistent with walking or exercise cadence)
- Sustained activity over 2.5 minutes (consistent with cardio workout)
- High but controlled acceleration (not chaotic, not stationary)

This most likely represents **walking** or an aerobic exercise. Given the repetitive nature and cadence, let me check the signal patterns more closely. However, with 117 cycles per minute and moderate variance, this aligns well with walking pace.

The rhythmic, sustained movement at walking cadence frequency strongly suggests "walking".
\end{thoughtbox}
\begin{lstlisting}[language=Python]
# Let me examine the signal segments to confirm the pattern
import matplotlib
matplotlib.use('Agg')
import matplotlib.pyplot as plt

# Sample a few segments to check patterns
fig, axes = plt.subplots(3, 1, figsize=(15, 10))
time = np.arange(0, min(len(x), 1000)) / 100.0  # First 10 seconds

axes[0].plot(time, x[:1000], label='X axis')
axes[0].set_ylabel('X Acceleration (g)')
axes[0].set_title('First 10 seconds of accelerometer data')
axes[0].grid(True)

axes[1].plot(time, y[:1000], label='Y axis')
axes[1].set_ylabel('Y Acceleration (g)')
axes[1].grid(True)

axes[2].plot(time, z[:1000], label='Z axis')
axes[2].set_ylabel('Z Acceleration (g)')
axes[2].set_xlabel('Time (s)')
axes[2].grid(True)

plt.tight_layout()
print("Signal visualization generated")
\end{lstlisting}
\end{agentbox}
\vspace{0.2cm}
\begin{outputbox}
Signal visualization generated
\par
\end{outputbox}
\vspace{0.2cm}
\begin{agentbox}
\begin{thoughtbox}
Based on my analysis of the accelerometer data:

Key findings:
1. **Duration**: 148 seconds of continuous activity
2. **Dominant frequency**: 1.95 Hz (\textasciitilde{}117 cycles/minute) - this is classic walking cadence range (100-120 steps/minute)
3. **Rhythmic pattern**: The zero-crossing rate and frequency analysis indicate repetitive, periodic movement
4. **Moderate variance**: 0.2337 g std magnitude suggests sustained, continuous movement
5. **Signal characteristics**: High peaks (up to 5.16g) consistent with foot impacts during locomotion

The \textasciitilde{}2 Hz rhythmic pattern at 117 cycles/minute is the strongest indicator - this matches human walking cadence perfectly. Other activities would have different signatures:
- Running: \textasciitilde{}2.7-3 Hz (160-180 steps/min)
- Sedentary (sitting/standing): much lower variance and frequency
- Sports/gym: variable patterns not this consistent

The sustained, rhythmic nature combined with the characteristic walking cadence strongly identifies this as walking activity.
\end{thoughtbox}
\begin{solutionbox}
\{
    "activity": "walking"
\}
\end{solutionbox}
\end{agentbox}
\vspace{0.2cm}

%% file: reasoning_traces/biomni.tex
\begin{userbox}
You are provided with 3-axis accelerometer data from a wearable device. 
Your task is to classify the physical activity being performed during the recorded segment. 
The 14 possible activities are: sleeping, manual-work, sitting, heavy loads, leisure, sports/gym, standing, transportation, occupation, activity, vehicle, mixed-activity, walking, and household-chores.
The 3-axis data is provided as three separate arrays corresponding to the x, y, and z axes of the accelerometer in 'input/x.npy', 'input/y.npy', and 'input/z.npy'. The signals are sampled at 100 Hz.
Please analyze the signal and determine which activity it represents in json format. Your answer must be one of the 14 possible activities. Please output your final result in a json object without any other text:
\{
    "activity": "activity\_name"
\}
\par
\end{userbox}
\vspace{0.2cm}
\begin{agentbox}
\begin{thoughtbox}
I'll analyze the 3-axis accelerometer data to classify the physical activity. Let me create a plan and start by loading and exploring the data.

Plan:
1. [ ] Load the accelerometer data from x.npy, y.npy, and z.npy files
2. [ ] Explore the data (shape, duration, basic statistics)
3. [ ] Calculate signal features (magnitude, variance, frequency characteristics)
4. [ ] Analyze activity patterns (movement intensity, periodicity)
5. [ ] Compare with typical patterns of each activity
6. [ ] Classify the activity and output in JSON format

Let me start:
\end{thoughtbox}
\begin{lstlisting}[language=Python]
import numpy as np
import json

# Load the accelerometer data
x = np.load('input/x.npy')
y = np.load('input/y.npy')
z = np.load('input/z.npy')

# Explore the data
print("Data shapes:")
print(f"X axis: {x.shape}")
print(f"Y axis: {y.shape}")
print(f"Z axis: {z.shape}")

print("\nData duration:")
 sampling_rate = 100  # Hz
duration = len(x) / sampling_rate
print(f"Duration: {duration:.2f} seconds")
print(f"Number of samples: {len(x)}")

print("\nBasic statistics:")
print(f"X - Mean: {np.mean(x):.4f}, Std: {np.std(x):.4f}, Min: {np.min(x):.4f}, Max: {np.max(x):.4f}")
print(f"Y - Mean: {np.mean(y):.4f}, Std: {np.std(y):.4f}, Min: {np.min(y):.4f}, Max: {np.max(y):.4f}")
print(f"Z - Mean: {np.mean(z):.4f}, Std: {np.std(z):.4f}, Min: {np.min(z):.4f}, Max: {np.max(z):.4f}")
\end{lstlisting}
\end{agentbox}
\vspace{0.2cm}
\begin{agentbox}
\begin{outputbox}
Error: unexpected indent (<string>, line 16)
\end{outputbox}
\end{agentbox}
\vspace{0.2cm}
\begin{agentbox}
\begin{lstlisting}[language=Python]
import numpy as np
import json

# Load the accelerometer data
x = np.load('input/x.npy')
y = np.load('input/y.npy')
z = np.load('input/z.npy')

# Explore the data
print("Data shapes:")
print(f"X axis: {x.shape}")
print(f"Y axis: {y.shape}")
print(f"Z axis: {z.shape}")

print("\nData duration:")
sampling_rate = 100  # Hz
duration = len(x) / sampling_rate
print(f"Duration: {duration:.2f} seconds")
print(f"Number of samples: {len(x)}")

print("\nBasic statistics:")
print(f"X - Mean: {np.mean(x):.4f}, Std: {np.std(x):.4f}, Min: {np.min(x):.4f}, Max: {np.max(x):.4f}")
print(f"Y - Mean: {np.mean(y):.4f}, Std: {np.std(y):.4f}, Min: {np.min(y):.4f}, Max: {np.max(y):.4f}")
print(f"Z - Mean: {np.mean(z):.4f}, Std: {np.std(z):.4f}, Min: {np.min(z):.4f}, Max: {np.max(z):.4f}")
\end{lstlisting}
\end{agentbox}
\vspace{0.2cm}
\begin{agentbox}
\begin{outputbox}
Data shapes:
X axis: (14800,)
Y axis: (14800,)
Z axis: (14800,)

Data duration:
Duration: 148.00 seconds
Number of samples: 14800

Basic statistics:
X - Mean: -0.2258, Std: 0.3409, Min: -5.1045, Max: 1.1073
Y - Mean: -0.7474, Std: 0.2365, Min: -2.3022, Max: 0.9180
Z - Mean: -0.4546, Std: 0.2301, Min: -1.6770, Max: 0.9072
\end{outputbox}
\end{agentbox}
\vspace{0.2cm}
\begin{agentbox}
\begin{lstlisting}[language=Python]
# Calculate signal magnitude (vector magnitude)
magnitude = np.sqrt(x**2 + y**2 + z**2)

print("Magnitude statistics:")
print(f"Mean: {np.mean(magnitude):.4f}, Std: {np.std(magnitude):.4f}, Min: {np.min(magnitude):.4f}, Max: {np.max(magnitude):.4f}")

# Calculate variance of magnitude
var_magnitude = np.var(magnitude)
print(f"Variance of magnitude: {var_magnitude:.4f}")

# Calculate signal amplitude range
range_x = np.max(x) - np.min(x)
range_y = np.max(y) - np.min(y)
range_z = np.max(z) - np.min(z)
print(f"\nAmplitude ranges:")
print(f"X range: {range_x:.4f}")
print(f"Y range: {range_y:.4f}")
print(f"Z range: {range_z:.4f}")

# Calculate average acceleration magnitude
avg_mag = np.mean(magnitude)
print(f"\nAverage magnitude: {avg_mag:.4f}")

# Calculate energy of the signal (sum of squares)
energy = np.sum(x**2 + y**2 + z**2)
print(f"Signal energy: {energy:.4f}")

# Calculate zero crossing rate (indicator of frequency content)
def zero_crossing_rate(signal):
    return np.sum(np.diff(np.sign(signal)) != 0) / len(signal)

zcr_x = zero_crossing_rate(x)
zcr_y = zero_crossing_rate(y)
zcr_z = zero_crossing_rate(z)
print(f"\nZero crossing rates:")
print(f"X: {zcr_x:.4f}")
print(f"Y: {zcr_y:.4f}")
print(f"Z: {zcr_z:.4f}")
\end{lstlisting}
\end{agentbox}
\vspace{0.2cm}
\begin{agentbox}
\begin{outputbox}
Magnitude statistics:
Mean: 0.9934, Std: 0.2337, Min: 0.0514, Max: 5.1575
Variance of magnitude: 0.0546

Amplitude ranges:
X range: 6.2119
Y range: 3.2202
Z range: 2.5841

Average magnitude: 0.9934
Signal energy: 15412.8530

Zero crossing rates:
X: 0.0664
Y: 0.0016
Z: 0.0124
\end{outputbox}
\end{agentbox}
\vspace{0.2cm}
\begin{agentbox}
\begin{lstlisting}[language=Python]
# Analyze frequency content using FFT
from scipy.fft import fft, fftfreq

def analyze_frequency(signal, sampling_rate):
    n = len(signal)
    yf = fft(signal)
    xf = fftfreq(n, 1/sampling_rate)
    # Get positive frequencies only
    positive_freq_idx = xf > 0
    xf = xf[positive_freq_idx]
    yf = np.abs(yf[positive_freq_idx])
    
    # Find dominant frequency
    dominant_freq_idx = np.argmax(yf)
    dominant_freq = xf[dominant_freq_idx]
    dominant_power = yf[dominant_freq_idx]
    
    # Calculate power in different frequency bands
    power_0_1 = np.mean(yf[(xf >= 0) & (xf <= 1)])
    power_1_3 = np.mean(yf[(xf > 1) & (xf <= 3)])
    power_3_10 = np.mean(yf[(xf > 3) & (xf <= 10)])
    power_10_plus = np.mean(yf[xf > 10])
    
    return {
        'dominant_freq': dominant_freq,
        'dominant_power': dominant_power,
        'power_0_1Hz': power_0_1,
        'power_1_3Hz': power_1_3,
        'power_3_10Hz': power_3_10,
        'power_10_plusHz': power_10_plus
    }

print("Frequency analysis:")
freq_x = analyze_frequency(x, sampling_rate)
freq_y = analyze_frequency(y, sampling_rate)
freq_z = analyze_frequency(z, sampling_rate)
freq_mag = analyze_frequency(magnitude, sampling_rate)

print(f"\nX axis - Dominant freq: {freq_x['dominant_freq']:.2f} Hz, Power in bands (0-1Hz): {freq_x['power_0_1Hz']:.2f}, (1-3Hz): {freq_x['power_1_3Hz']:.2f}, (3-10Hz): {freq_x['power_3_10Hz']:.2f}")
print(f"Y axis - Dominant freq: {freq_y['dominant_freq']:.2f} Hz, Power in bands (0-1Hz): {freq_y['power_0_1Hz']:.2f}, (1-3Hz): {freq_y['power_1_3Hz']:.2f}, (3-10Hz): {freq_y['power_3_10Hz']:.2f}")
print(f"Z axis - Dominant freq: {freq_z['dominant_freq']:.2f} Hz, Power in bands (0-1Hz): {freq_z['power_0_1Hz']:.2f}, (1-3Hz): {freq_z['power_1_3Hz']:.2f}, (3-10Hz): {freq_z['power_3_10Hz']:.2f}")
print(f"Magnitude - Dominant freq: {freq_mag['dominant_freq']:.2f} Hz, Power in bands (0-1Hz): {freq_mag['power_0_1Hz']:.2f}, (1-3Hz): {freq_mag['power_1_3Hz']:.2f}, (3-10Hz): {freq_mag['power_3_10Hz']:.2f}")

# Analyze signal segments to check for consistency
print("\n\nSegment analysis (dividing into 5 segments):")
n_segments = 5
segment_length = len(x) // n_segments

for i in range(n_segments):
    start = i * segment_length
    end = (i + 1) * segment_length
    seg_mag = magnitude[start:end]
    seg_var = np.var(seg_mag)
    seg_mean = np.mean(seg_mag)
    print(f"Segment {i+1}: Mean mag: {seg_mean:.4f}, Var: {seg_var:.4f}")
\end{lstlisting}
\end{agentbox}
\vspace{0.2cm}
\begin{agentbox}
\begin{outputbox}
Frequency analysis:

X axis - Dominant freq: 0.01 Hz, Power in bands (0-1Hz): 137.95, (1-3Hz): 67.98, (3-10Hz): 35.69
Y axis - Dominant freq: 1.86 Hz, Power in bands (0-1Hz): 83.19, (1-3Hz): 72.20, (3-10Hz): 26.14
Z axis - Dominant freq: 0.01 Hz, Power in bands (0-1Hz): 89.53, (1-3Hz): 57.88, (3-10Hz): 28.11
Magnitude - Dominant freq: 1.86 Hz, Power in bands (0-1Hz): 42.20, (1-3Hz): 81.41, (3-10Hz): 27.72

Segment analysis (dividing into 5 segments):
Segment 1: Mean mag: 0.9945, Var: 0.0294
Segment 2: Mean mag: 1.0044, Var: 0.0768
Segment 3: Mean mag: 0.9997, Var: 0.0758
Segment 4: Mean mag: 0.9930, Var: 0.0772
Segment 5: Mean mag: 0.9752, Var: 0.0133
\end{outputbox}
\end{agentbox}
\vspace{0.2cm}
\begin{agentbox}
\begin{lstlisting}[language=Python]
# Additional analysis to confirm walking pattern
# Calculate autocorrelation to detect periodicity

def autocorrelation(signal, max_lag=200):
    """Calculate autocorrelation of signal"""
    n = len(signal)
    mean = np.mean(signal)
    var = np.var(signal)
    if var == 0:
        return np.zeros(max_lag)
    
    # Normalize signal
    signal_normalized = (signal - mean) / np.sqrt(var)
    
    # Calculate autocorrelation
    autocorr = np.zeros(max_lag)
    for lag in range(max_lag):
        if lag < n:
            autocorr[lag] = np.corrcoef(signal_normalized[:-lag] if lag > 0 else signal_normalized, 
                                       signal_normalized[lag:])[0, 1]
    
    return autocorr

# Calculate autocorrelation of magnitude
autocorr_mag = autocorrelation(magnitude, max_lag=200)

# Find peaks in autocorrelation (indicates periodicity)
from scipy.signal import find_peaks
peaks, _ = find_peaks(autocorr_mag, height=0.1, distance=20)

print("Autocorrelation analysis:")
print(f"Number of peaks found: {len(peaks)}")
if len(peaks) > 0:
    peak_lags = peaks
    peak_periods = peak_lags / sampling_rate  # Convert to seconds
    peak_freqs = 1 / peak_periods  # Convert to Hz
    print(f"Peak lags (samples): {peak_lags[:5]}")  # Show first 5
    print(f"Peak periods (seconds): {peak_periods[:5]}")
    print(f"Peak frequencies (Hz): {peak_freqs[:5]}")
    print(f"Primary frequency: {peak_freqs[0]:.2f} Hz")

# Calculate step count estimate based on zero crossings in magnitude
mag_detrended = magnitude - np.mean(magnitude)
zero_crossings = np.sum(np.diff(np.sign(mag_detrended)) != 0)
estimated_steps = zero_crossings // 2  # Each step typically causes 2 zero crossings
print(f"\nEstimated steps from zero crossings: {estimated_steps}")

# Walking pace estimation
walking_pace = estimated_steps / duration
print(f"Walking pace: {walking_pace:.2f} steps/second ({walking_pace * 60:.1f} steps/minute)")

# Signal-to-noise ratio (measure of rhythmicity)
signal_power = np.var(magnitude)
residual_noise = np.var(mag_detrended - np.mean(mag_detrended))
snr = 10 * np.log10(signal_power / residual_noise) if residual_noise > 0 else float('inf')
print(f"Signal-to-noise ratio: {snr:.2f} dB")

# Classification based on features
print("\n\nClassification Analysis:")
print("=" * 50)

# Key features for classification
print(f"Average magnitude: {avg_mag:.4f} (g)")
print(f"Magnitude variance: {var_magnitude:.4f}")
print(f"Dominant frequency: {freq_mag['dominant_freq']:.2f} Hz")
print(f"Power in walking band (1-3Hz): {freq_mag['power_1_3Hz']:.2f}")
print(f"Power ratio (1-3Hz / 0-1Hz): {freq_mag['power_1_3Hz'] / freq_mag['power_0_1Hz']:.2f}")
print(f"Estimated walking pace: {walking_pace * 60:.1f} steps/minute")

# Compare with expected values for different activities
print("\n\nComparison with activity characteristics:")
print("- Walking: Magnitude ~1g, dominant freq 1.5-2.5Hz, consistent rhythm")
print("- Running: Magnitude >1.5g, dominant freq 2.5-4Hz, higher energy")
print("- Sitting/Standing: Magnitude ~1g, very low variance, no rhythm")
print("- Sleeping: Very low magnitude, very low variance")
print("- Manual work: High variance, irregular patterns")
print("- Sports/Gym: High magnitude, high energy, variable patterns")

# Final classification decision
if (0.8 < avg_mag < 1.2 and 
    1.5 < freq_mag['dominant_freq'] < 2.5 and 
    var_magnitude < 0.1 and
    freq_mag['power_1_3Hz'] > freq_mag['power_0_1Hz'] * 1.5):
    predicted_activity = "walking"
    print("\n\nPREDICTED ACTIVITY: walking")
    print("Confidence: HIGH - matches all walking characteristics")
else:
    print("\n\nActivity requires further analysis")
    predicted_activity = "walking"  # Still the best match based on the data
\end{lstlisting}
\end{agentbox}
\vspace{0.2cm}
\begin{agentbox}
\begin{outputbox}
Autocorrelation analysis:
Number of peaks found: 3
Peak lags (samples): [ 54 108 160]
Peak periods (seconds): [0.54 1.08 1.6 ]
Peak frequencies (Hz): [1.85185185 0.92592593 0.625     ]
Primary frequency: 1.85 Hz

Estimated steps from zero crossings: 645
Walking pace: 4.36 steps/second (261.5 steps/minute)
Signal-to-noise ratio: 0.00 dB

Classification Analysis:
==================================================
Average magnitude: 0.9934 (g)
Magnitude variance: 0.0546
Dominant frequency: 1.86 Hz
Power in walking band (1-3Hz): 81.41
Power ratio (1-3Hz / 0-1Hz): 1.93
Estimated walking pace: 261.5 steps/minute

Comparison with activity characteristics:
- Walking: Magnitude \textasciitilde{}1g, dominant freq 1.5-2.5Hz, consistent rhythm
- Running: Magnitude >1.5g, dominant freq 2.5-4Hz, higher energy
- Sitting/Standing: Magnitude \textasciitilde{}1g, very low variance, no rhythm
- Sleeping: Very low magnitude, very low variance
- Manual work: High variance, irregular patterns
- Sports/Gym: High magnitude, high energy, variable patterns

PREDICTED ACTIVITY: walking
Confidence: HIGH - matches all walking characteristics
\end{outputbox}
\end{agentbox}
\vspace{0.2cm}
\begin{agentbox}
\begin{lstlisting}[language=Python]
# Create the final JSON output
result = {
    "activity": "walking"
}

# Save to file and print
output_json = json.dumps(result, indent=4)
print(output_json)

# Also save to a file
with open('output/activity_classification.json', 'w') as f:
    json.dump(result, f, indent=4)

print("\nClassification complete!")
print("Activity identified as: walking")
print("Based on:")
print(f"  - Average magnitude (~1g): {avg_mag:.4f}")
print(f"  - Dominant frequency (1.86 Hz in walking band): {freq_mag['dominant_freq']:.2f} Hz")
print(f"  - Strong rhythmicity with autocorrelation peaks at 1.85 Hz")
print(f"  - Consistent periodic movement pattern")
\end{lstlisting}
\end{agentbox}
\vspace{0.2cm}
\begin{agentbox}
\begin{outputbox}
\{
    "activity": "walking"
\}

Classification complete!
Activity identified as: walking
Based on:
  - Average magnitude (\textasciitilde{}1g): 0.9934
  - Dominant frequency (1.86 Hz in walking band): 1.86 Hz
  - Strong rhythmicity with autocorrelation peaks at 1.85 Hz
  - Consistent periodic movement pattern
\end{outputbox}
\end{agentbox}
\vspace{0.2cm}
\begin{agentbox}
\begin{lstlisting}[language=Python]
{
    "activity": "walking"
}
\end{lstlisting}
\end{agentbox}
\vspace{0.2cm}
\begin{agentbox}
\begin{solutionbox}
\{
    "activity": "walking"
\}
\end{solutionbox}
\end{agentbox}
\vspace{0.2cm}

%% file: tables/table_Appendix_opentslm.tex
\begin{table}[h]
\centering
\caption{\textbf{Performance Comparison between LLMs and OpenTSLM}}
\begin{tabular}{lccc} 
\toprule[1.5pt]
\textbf{Task} & \textbf{Avg. LLM} & \textbf{Best LLM} & \textbf{OpenTSLM} \cite{langer2025opentslm} \\ 
\midrule\midrule
Sleep Stage Classification & 0.41 & 0.54 & 0.27 \\ 
Episode Level Ordering     & 0.51 & 0.55 & 0.07   \\ 
Activity Classification    & 0.11 & 0.17 & 0.00   \\ 
\bottomrule[1.5pt]
\end{tabular}
\label{tab:opentslm}
\end{table}

%% file: tables/table_Appendix_direct_reasoning.tex
\begin{table}[h]
\centering
\small
\caption{Comparison of Direct In-Context Reasoning vs. CodeAct performance on a subset of 20 representative tasks. Results are formatted as (In-Context / CodeAct).}
\label{tab:direct_reasoning}
\resizebox{0.8\linewidth}{!}{
\begin{tabular}{lccccc}
\toprule[1.5pt]
\textbf{Model} & \textbf{Perception} & \textbf{Inference} & \textbf{Generation} & \textbf{Deduction} & \textbf{Pearson $r$} \\
\midrule\midrule
\raisebox{-0.25em}{\includegraphics[height=1.2em]{icons/gemini-color.png}} \geminiFlash & 0.91 / 0.84 & 0.40 / 0.39 & 0.58 / 0.58 & 0.49 / 0.53 & 0.93 \\
\raisebox{-0.25em}{\includegraphics[height=1.2em]{icons/claude-color.png}} \claudeHaiku & 0.77 / 0.91 & 0.35 / 0.37 & 0.65 / 0.65 & 0.49 / 0.52 & 0.92 \\
\raisebox{-0.25em}{\includegraphics[height=1.2em]{icons/deepseek-color.png}} \deepseek    & 0.72 / 0.88 & 0.33 / 0.37 & 0.63 / 0.59 & 0.54 / 0.55 & 0.89 \\
\raisebox{-0.25em}{\includegraphics[height=1.2em]{icons/chatglm-color.png}} \glm          & 0.70 / 0.90 & 0.37 / 0.41 & 0.72 / 0.65 & 0.63 / 0.54 & 0.96 \\
\bottomrule[1.5pt]
\end{tabular}
}
\end{table}

%% file: tables/table_Appendix_lightening_data.tex
\begin{table}[ht]
    \small
    \centering
    \renewcommand{\arraystretch}{1.2}
    \caption{\textbf{Number of test samples per dataset in \lightname.}}
    \label{tab:lightening_data}
    
    \begin{tabular}{l l c}
        \toprule[1.5pt]
        \textbf{Dataset} & \textbf{Domain} & \textbf{\# Test Samples} \\
        \midrule\midrule
        Capture24 & Motion & 350 \\
        PAMAP2 & Motion & 50 \\
        Shanghai Diabetes & Metabolic & 82 \\
        CGMacros & Metabolic & 683 \\
        VitalDB & Surgery & 500 \\
        SHHS & Sleep & 1200 \\
        Harespod & Respiration & 200 \\
        PhyMER & Emotion & 480 \\
        PERG-IOBA & Ophthalmology & 250 \\
        GazeBase & Eye Movement & 250 \\
        GLOBEM & Behavior & 300 \\
        Bridge2AI-voice & Speech & 400 \\
        VCTK & Speech & 50 \\
        GrabMyo & Gesture & 100 \\
        CoughVID & COVID Cough & 272 \\
        Coswara & COVID Cough & 250 \\
        
        \bottomrule[1.5pt]
    \end{tabular}
\end{table}

%% file: ref.bib
@article{alvarez2010energy,
  title={Energy time series forecasting based on pattern sequence similarity},
  author={Alvarez, Francisco Martinez and Troncoso, Alicia and Riquelme, Jose C and Ruiz, Jesus S Aguilar},
  journal={IEEE Transactions on Knowledge and Data Engineering},
  volume={23},
  number={8},
  pages={1230--1243},
  year={2010},
  publisher={IEEE}
}

@article{xu2026sleeplm,
  title={SleepLM: Natural-Language Intelligence for Human Sleep},
  author={Xu, Zongzhe and Shuai, Zitao and Mozaffari, Eideen and Aysola, Ravi S and Kumar, Rajesh and Yang, Yuzhe},
  journal={arXiv preprint arXiv:2602.23605},
  year={2026}
}

@article{shuai2026osf,
  title={OSF: On Pre-training and Scaling of Sleep Foundation Models},
  author={Shuai, Zitao and Xu, Zongzhe and Yang, David and Wang, Wei and Yang, Yuzhe},
  journal={arXiv preprint arXiv:2603.00190},
  year={2026}
}

@article{lu2024trnn,
  title={TRNN: An efficient time-series recurrent neural network for stock price prediction},
  author={Lu, Minrong and Xu, Xuerong},
  journal={Information Sciences},
  volume={657},
  pages={119951},
  year={2024},
  publisher={Elsevier}
}

@article{mudelsee2019trend,
  title={Trend analysis of climate time series: A review of methods},
  author={Mudelsee, Manfred},
  journal={Earth-science reviews},
  volume={190},
  pages={310--322},
  year={2019},
  publisher={Elsevier}
}

@inproceedings{guan2023spatial,
  title={Spatial-temporal graph multi-gate mixture-of-expert model for traffic prediction},
  author={Guan, Tong and Peng, Jiaheng and Liang, Jun},
  booktitle={2023 IEEE 26th International Conference on Intelligent Transportation Systems (ITSC)},
  pages={36--41},
  year={2023},
  organization={IEEE}
}

@article{beck2011modeling,
  title={Modeling dynamics in time-series--cross-section political economy data},
  author={Beck, Nathaniel and Katz, Jonathan N},
  journal={Annual review of political science},
  volume={14},
  number={1},
  pages={331--352},
  year={2011},
  publisher={Annual Reviews}
}

@article{zhang2025sensorlm,
  title={SensorLM: Learning the Language of Wearable Sensors},
  author={Zhang, Yuwei and Ayush, Kumar and Qiao, Siyuan and Heydari, A Ali and Narayanswamy, Girish and Xu, Maxwell A and Metwally, Ahmed A and Xu, Shawn and Garrison, Jake and Xu, Xuhai and others},
  journal={arXiv preprint arXiv:2506.09108},
  year={2025}
}

@article{wei2022chain-text,
  title={Chain-of-thought prompting elicits reasoning in large language models},
  author={Wei, Jason and Wang, Xuezhi and Schuurmans, Dale and Bosma, Maarten and Xia, Fei and Chi, Ed and Le, Quoc V and Zhou, Denny and others},
  journal={Advances in neural information processing systems},
  volume={35},
  pages={24824--24837},
  year={2022}
}

@article{yang2025code,
  title={Code to think, think to code: A survey on code-enhanced reasoning and reasoning-driven code intelligence in llms},
  author={Yang, Dayu and Liu, Tianyang and Zhang, Daoan and Simoulin, Antoine and Liu, Xiaoyi and Cao, Yuwei and Teng, Zhaopu and Qian, Xin and Yang, Grey and Luo, Jiebo and others},
  journal={arXiv preprint arXiv:2502.19411},
  year={2025}
}

@article{shao2024deepseekmath,
  title={Deepseekmath: Pushing the limits of mathematical reasoning in open language models},
  author={Shao, Zhihong and Wang, Peiyi and Zhu, Qihao and Xu, Runxin and Song, Junxiao and Bi, Xiao and Zhang, Haowei and Zhang, Mingchuan and Li, YK and Wu, Yang and others},
  journal={arXiv preprint arXiv:2402.03300},
  year={2024}
}

@article{jin2023timellm,
  title={Time-llm: Time series forecasting by reprogramming large language models},
  author={Jin, Ming and Wang, Shiyu and Ma, Lintao and Chu, Zhixuan and Zhang, James Y and Shi, Xiaoming and Chen, Pin-Yu and Liang, Yuxuan and Li, Yuan-Fang and Pan, Shirui and others},
  journal={arXiv preprint arXiv:2310.01728},
  year={2023}
}

@article{gruver2023largellm,
  title={Large language models are zero-shot time series forecasters},
  author={Gruver, Nate and Finzi, Marc and Qiu, Shikai and Wilson, Andrew G},
  journal={Advances in Neural Information Processing Systems},
  volume={36},
  pages={19622--19635},
  year={2023}
}

@article{oh2023ecgqa,
  title={Ecg-qa: A comprehensive question answering dataset combined with electrocardiogram},
  author={Oh, Jungwoo and Lee, Gyubok and Bae, Seongsu and Kwon, Joon-myoung and Choi, Edward},
  journal={Advances in Neural Information Processing Systems},
  volume={36},
  pages={66277--66288},
  year={2023}
}

@article{ye2025tsaia,
  title={When LLM Meets Time Series: Can LLMs Perform Multi-Step Time Series Reasoning and Inference},
  author={Ye, Wen and Liu, Jinbo and Cao, Defu and Yang, Wei and Liu, Yan},
  journal={arXiv preprint arXiv:2509.01822},
  year={2025}
}

@article{kong2025time-mqa,
  title={Time-mqa: Time series multi-task question answering with context enhancement},
  author={Kong, Yaxuan and Yang, Yiyuan and Hwang, Yoontae and Du, Wenjie and Zohren, Stefan and Wang, Zhangyang and Jin, Ming and Wen, Qingsong},
  journal={arXiv preprint arXiv:2503.01875},
  year={2025}
}

@inproceedings{merrill2024tsandlanguage,
  title={Language models still struggle to zero-shot reason about time series},
  author={Merrill, Mike A and Tan, Mingtian and Gupta, Vinayak and Hartvigsen, Thomas and Althoff, Tim},
  booktitle={Findings of the Association for Computational Linguistics: EMNLP 2024},
  pages={3512--3533},
  year={2024}
}

@article{guan2025timeomni,
  title={TimeOmni-1: Incentivizing Complex Reasoning with Time Series in Large Language Models},
  author={Guan, Tong and Meng, Zijie and Li, Dianqi and Wang, Shiyu and Yang, Chao-Han Huck and Wen, Qingsong and Liu, Zuozhu and Siniscalchi, Sabato Marco and Jin, Ming and Pan, Shirui},
  journal={arXiv preprint arXiv:2509.24803},
  year={2025}
}

@article{cai2024timeseriesexam,
  title={Timeseriesexam: A time series understanding exam},
  author={Cai, Yifu and Choudhry, Arjun and Goswami, Mononito and Dubrawski, Artur},
  journal={arXiv preprint arXiv:2410.14752},
  year={2024}
}

@article{chen2025mtbench,
  title={Mtbench: A multimodal time series benchmark for temporal reasoning and question answering},
  author={Chen, Jialin and Feng, Aosong and Zhao, Ziyu and Garza, Juan and Nurbek, Gaukhar and Qin, Cheng and Maatouk, Ali and Tassiulas, Leandros and Gao, Yifeng and Ying, Rex},
  journal={arXiv preprint arXiv:2503.16858},
  year={2025}
}

@article{sen2025bedtime,
  title={BEDTime: A Unified Benchmark for Automatically Describing Time Series},
  author={Sen, Medhasweta and Gottesman, Zachary and Qiu, Jiaxing and Bruss, C Bayan and Nguyen, Nam and Hartvigsen, Tom},
  journal={arXiv preprint arXiv:2509.05215},
  year={2025}
}

@article{quan1997shhs,
  title={The sleep heart health study: design, rationale, and methods},
  author={Quan, Stuart F and Howard, Barbara V and Iber, Conrad and Kiley, James P and Nieto, F Javier and O'Connor, George T and Rapoport, David M and Redline, Susan and Robbins, John and Samet, Jonathan M and others},
  journal={Sleep},
  volume={20},
  number={12},
  pages={1077--1085},
  year={1997},
  publisher={Oxford University Press}
}

@article{chan2024capture24,
  title={CAPTURE-24: A large dataset of wrist-worn activity tracker data collected in the wild for human activity recognition},
  author={Chan, Shing and Hang, Yuan and Tong, Catherine and Acquah, Aidan and Schonfeldt, Abram and Gershuny, Jonathan and Doherty, Aiden},
  journal={Scientific Data},
  volume={11},
  number={1},
  pages={1135},
  year={2024},
  publisher={Nature Publishing Group UK London}
}

@incollection{bleser2015pamap2,
  title={Personalized physical activity monitoring using wearable sensors},
  author={Bleser, Gabriele and Steffen, Daniel and Reiss, Attila and Weber, Markus and Hendeby, Gustaf and Fradet, Laetitia},
  booktitle={Smart health: Open problems and future challenges},
  pages={99--124},
  year={2015},
  publisher={Springer}
}

@article{pant2023phymer,
  title={PhyMER: physiological dataset for multimodal emotion recognition with personality as a context},
  author={Pant, Sudarshan and Yang, Hyung-Jeong and Lim, Eunchae and Kim, Soo-Hyung and Yoo, Seok-Bong},
  journal={IEEE Access},
  volume={11},
  pages={107638--107656},
  year={2023},
  publisher={IEEE}
}

@article{lee2022vitaldb,
  title={VitalDB, a high-fidelity multi-parameter vital signs database in surgical patients},
  author={Lee, Hyung-Chul and Park, Yoonsang and Yoon, Soo Bin and Yang, Seong Mi and Park, Dongnyeok and Jung, Chul-Woo},
  journal={Scientific Data},
  volume={9},
  number={1},
  pages={279},
  year={2022},
  publisher={Nature Publishing Group UK London}
}

@article{xu2023globem,
  title={Globem: Cross-dataset generalization of longitudinal human behavior modeling},
  author={Xu, Xuhai and Liu, Xin and Zhang, Han and Wang, Weichen and Nepal, Subigya and Sefidgar, Yasaman and Seo, Woosuk and Kuehn, Kevin S and Huckins, Jeremy F and Morris, Margaret E and others},
  journal={Proceedings of the ACM on Interactive, Mobile, Wearable and Ubiquitous Technologies},
  volume={6},
  number={4},
  pages={1--34},
  year={2023},
  publisher={ACM New York, NY, USA}
}

@misc{bensoussan2025bridge2ai,
  title = {Bridge2AI-Voice: An ethically-sourced, diverse voice dataset linked to health information},
  author = {Bensoussan, Yael and others},
  year = {2025},
  publisher = {PhysioNet},
  version = {2.0.0},
  doi = {10.13026/3xt6-rf05},
  url = {https://doi.org/10.13026/3xt6-rf05},
  note = {Version 2.0.0}
}

@article{yamagishi2019vctk,
  title={CSTR VCTK Corpus: English multi-speaker corpus for CSTR voice cloning toolkit (version 0.92)},
  author={Yamagishi, Junichi and Veaux, Christophe and MacDonald, Kirsten},
  journal={The Rainbow Passage which the speakers read out can be found in the International Dialects of English Archive:(http://web. ku. edu/\~{} idea/readings/rainbow. htm).},
  year={2019}
}

@article{fernandez2024perg,
  title={A comprehensive dataset of pattern electroretinograms for ocular electrophysiology research},
  author={Fern{\'a}ndez, Itziar and Cuadrado-Asensio, Rub{\'e}n and Larriba, Yolanda and Rueda, Cristina and Coco-Mart{\'\i}n, Rosa M},
  journal={Scientific Data},
  volume={11},
  number={1},
  pages={1013},
  year={2024},
  publisher={Nature Publishing Group UK London}
}

@article{griffith2021gazebase,
  title={GazeBase, a large-scale, multi-stimulus, longitudinal eye movement dataset},
  author={Griffith, Henry and Lohr, Dillon and Abdulin, Evgeny and Komogortsev, Oleg},
  journal={Scientific Data},
  volume={8},
  number={1},
  pages={184},
  year={2021},
  publisher={Nature Publishing Group UK London}
}

@article{das2025cgmacros,
  title={CGMacros: a pilot scientific dataset for personalized nutrition and diet monitoring},
  author={Das, Anurag and Kerr, David and Glantz, Namino and Bevier, Wendy and Santiago, Rony and Gutierrez-Osuna, Ricardo and Mortazavi, Bobak J},
  journal={Scientific Data},
  volume={12},
  number={1},
  pages={1557},
  year={2025},
  publisher={Nature Publishing Group UK London}
}

@article{zhao2023shanghai,
  title={Chinese diabetes datasets for data-driven machine learning},
  author={Zhao, Qinpei and Zhu, Jinhao and Shen, Xuan and Lin, Chuwen and Zhang, Yinjia and Liang, Yuxiang and Cao, Baige and Li, Jiangfeng and Liu, Xiang and Rao, Weixiong and others},
  journal={Scientific Data},
  volume={10},
  number={1},
  pages={35},
  year={2023},
  publisher={Nature Publishing Group UK London}
}

@article{orlandic2021coughvid,
  title={The COUGHVID crowdsourcing dataset, a corpus for the study of large-scale cough analysis algorithms},
  author={Orlandic, Lara and Teijeiro, Tomas and Atienza, David},
  journal={Scientific Data},
  volume={8},
  number={1},
  pages={156},
  year={2021},
  publisher={Nature Publishing Group UK London}
}

@article{sharma2020coswara,
  title={Coswara--a database of breathing, cough, and voice sounds for COVID-19 diagnosis},
  author={Sharma, Neeraj and Krishnan, Prashant and Kumar, Rohit and Ramoji, Shreyas and Chetupalli, Srikanth Raj and Ghosh, Prasanta Kumar and Ganapathy, Sriram and others},
  journal={arXiv preprint arXiv:2005.10548},
  year={2020}
}

@article{zhang2024harespod,
  title={A high altitude respiration and spo2 dataset for assessing the human response to hypoxia},
  author={Zhang, Xi and Zhang, Yu and Si, Yingjun and Gao, Nan and Zhang, Honghao and Yang, Hui},
  journal={Scientific Data},
  volume={11},
  number={1},
  pages={248},
  year={2024},
  publisher={Nature Publishing Group UK London}
}

@article{pradhan2022grabmyo,
  title={Multi-day dataset of forearm and wrist electromyogram for hand gesture recognition and biometrics},
  author={Pradhan, Ashirbad and He, Jiayuan and Jiang, Ning},
  journal={Scientific data},
  volume={9},
  number={1},
  pages={733},
  year={2022},
  publisher={Nature Publishing Group UK London}
}

@article{singh2025openaigpt5,
  title={OpenAI GPT-5 System Card},
  author={Singh, Aaditya and Fry, Adam and Perelman, Adam and Tart, Adam and Ganesh, Adi and El-Kishky, Ahmed and McLaughlin, Aidan and Low, Aiden and Ostrow, AJ and Ananthram, Akhila and others},
  journal={arXiv preprint arXiv:2601.03267},
  year={2025}
}

@misc{openai2025gpt41mini,
  author={{OpenAI}},
  title={GPT-4.1 mini: Advancing Cost-Efficient Intelligence},
  year={2025},
  month=apr,
  howpublished={Model documentation, \url{https://platform.openai.com/docs/models/gpt-4.1-mini}},
  note={Accessed: 2026-01-27}
}

@misc{meta2025llama4maverick,
  author={{Meta AI}},
  title={Llama 4 Maverick Model Card},
  year={2025},
  month=apr,
  howpublished={Hugging Face repository, \url{https://huggingface.co/meta-llama/Llama-4-Maverick-17B-128E-Original}},
  note={Accessed: 2026-01-27}
}

@article{comanici2025gemini,
  title={Gemini 2.5: Pushing the frontier with advanced reasoning, multimodality, long context, and next generation agentic capabilities},
  author={Comanici, Gheorghe and Bieber, Eric and Schaekermann, Mike and Pasupat, Ice and Sachdeva, Noveen and Dhillon, Inderjit and Blistein, Marcel and Ram, Ori and Zhang, Dan and Rosen, Evan and others},
  journal={arXiv preprint arXiv:2507.06261},
  year={2025}
}

@article{yang2025qwen3,
  title={Qwen3 technical report},
  author={Yang, An and Li, Anfeng and Yang, Baosong and Zhang, Beichen and Hui, Binyuan and Zheng, Bo and Yu, Bowen and Gao, Chang and Huang, Chengen and Lv, Chenxu and others},
  journal={arXiv preprint arXiv:2505.09388},
  year={2025}
}

@article{liu2024deepseek,
  title={Deepseek-v3 technical report},
  author={Liu, Aixin and Feng, Bei and Xue, Bing and Wang, Bingxuan and Wu, Bochao and Lu, Chengda and Zhao, Chenggang and Deng, Chengqi and Zhang, Chenyu and Ruan, Chong and others},
  journal={arXiv preprint arXiv:2412.19437},
  year={2024}
}

@misc{minimax2025m2,
  author={{MiniMax AI}},
  title={MiniMax-M2: MoE Model for Agentic and Coding Capabilities},
  year={2025},
  howpublished={Blog post, \url{https://www.minimax.io/news/minimax-m2}},
  note={Accessed: 2026-01-27}
}

@misc{zhipuai2025glm47,
  author={{Zhipu AI}},
  title={GLM-4.7: Advancing the Coding Capability},
  year={2025},
  month=dec,
  howpublished={Blog post, \url{https://z.ai/blog/glm-4.7}},
  note={Accessed: 2026-01-27}
}

@misc{xai2025grok,
  author={{xAI}},
  title={Grok 4.1 Fast and Agent Tools API},
  year={2025},
  month=nov,
  howpublished={Blog post, \url{https://x.ai/news/grok-4-1-fast}},
  note={Accessed: 2026-01-27}
}

@misc{googledeepmind2026gemini3.1,
  author={{Google DeepMind}},
  title={Gemini 3.1 Pro Best for complex tasks and bringing creative concepts to life},
  year={2026},
  month=feb,
  howpublished={Blog post, \url{https://deepmind.google/models/gemini/pro/}},
  note={Accessed: 2026-05-27}
}

@misc{glm5team2026glm5vibecodingagentic,
      title={GLM-5: from Vibe Coding to Agentic Engineering}, 
      author={GLM-5-Team and : and Aohan Zeng and Xin Lv and Zhenyu Hou and Zhengxiao Du and Qinkai Zheng and Bin Chen and Da Yin and Chendi Ge and Chenghua Huang and Chengxing Xie and Chenzheng Zhu and Congfeng Yin and Cunxiang Wang and Gengzheng Pan and Hao Zeng and Haoke Zhang and Haoran Wang and Huilong Chen and Jiajie Zhang and Jian Jiao and Jiaqi Guo and Jingsen Wang and Jingzhao Du and Jinzhu Wu and Kedong Wang and Lei Li and Lin Fan and Lucen Zhong and Mingdao Liu and Mingming Zhao and Pengfan Du and Qian Dong and Rui Lu and Shuang-Li and Shulin Cao and Song Liu and Ting Jiang and Xiaodong Chen and Xiaohan Zhang and Xuancheng Huang and Xuezhen Dong and Yabo Xu and Yao Wei and Yifan An and Yilin Niu and Yitong Zhu and Yuanhao Wen and Yukuo Cen and Yushi Bai and Zhongpei Qiao and Zihan Wang and Zikang Wang and Zilin Zhu and Ziqiang Liu and Zixuan Li and Bojie Wang and Bosi Wen and Can Huang and Changpeng Cai and Chao Yu and Chen Li and Chengwei Hu and Chenhui Zhang and Dan Zhang and Daoyan Lin and Dayong Yang and Di Wang and Ding Ai and Erle Zhu and Fangzhou Yi and Feiyu Chen and Guohong Wen and Hailong Sun and Haisha Zhao and Haiyi Hu and Hanchen Zhang and Hanrui Liu and Hanyu Zhang and Hao Peng and Hao Tai and Haobo Zhang and He Liu and Hongwei Wang and Hongxi Yan and Hongyu Ge and Huan Liu and Huanpeng Chu and Jia'ni Zhao and Jiachen Wang and Jiajing Zhao and Jiamin Ren and Jiapeng Wang and Jiaxin Zhang and Jiayi Gui and Jiayue Zhao and Jijie Li and Jing An and Jing Li and Jingwei Yuan and Jinhua Du and Jinxin Liu and Junkai Zhi and Junwen Duan and Kaiyue Zhou and Kangjian Wei and Ke Wang and Keyun Luo and Laiqiang Zhang and Leigang Sha and Liang Xu and Lindong Wu and Lintao Ding and Lu Chen and Minghao Li and Nianyi Lin and Pan Ta and Qiang Zou and Rongjun Song and Ruiqi Yang and Shangqing Tu and Shangtong Yang and Shaoxiang Wu and Shengyan Zhang and Shijie Li and Shuang Li and Shuyi Fan and Wei Qin and Wei Tian and Weining Zhang and Wenbo Yu and Wenjie Liang and Xiang Kuang and Xiangmeng Cheng and Xiangyang Li and Xiaoquan Yan and Xiaowei Hu and Xiaoying Ling and Xing Fan and Xingye Xia and Xinyuan Zhang and Xinze Zhang and Xirui Pan and Xu Zou and Xunkai Zhang and Yadi Liu and Yandong Wu and Yanfu Li and Yidong Wang and Yifan Zhu and Yijun Tan and Yilin Zhou and Yiming Pan and Ying Zhang and Yinpei Su and Yipeng Geng and Yong Yan and Yonglin Tan and Yuean Bi and Yuhan Shen and Yuhao Yang and Yujiang Li and Yunan Liu and Yunqing Wang and Yuntao Li and Yurong Wu and Yutao Zhang and Yuxi Duan and Yuxuan Zhang and Zezhen Liu and Zhengtao Jiang and Zhenhe Yan and Zheyu Zhang and Zhixiang Wei and Zhuo Chen and Zhuoer Feng and Zijun Yao and Ziwei Chai and Ziyuan Wang and Zuzhou Zhang and Bin Xu and Minlie Huang and Hongning Wang and Juanzi Li and Yuxiao Dong and Jie Tang},
      year={2026},
      eprint={2602.15763},
      archivePrefix={arXiv},
      primaryClass={cs.LG},
      url={https://arxiv.org/abs/2602.15763}, 
}

@article{basant2025nvidianemotron,
  title={Nvidia nemotron nano 2: An accurate and efficient hybrid mamba-transformer reasoning model},
  author={Basant, Aarti and Khairnar, Abhijit and Paithankar, Abhijit and Khattar, Abhinav and Renduchintala, Adithya and Malte, Aditya and Bercovich, Akhiad and Hazare, Akshay and Rico, Alejandra and Ficek, Aleksander and others},
  journal={arXiv preprint arXiv:2508.14444},
  year={2025}
}

@article{team2025kimi,
  title={Kimi k2: Open agentic intelligence},
  author={Team, Kimi and Bai, Yifan and Bao, Yiping and Chen, Guanduo and Chen, Jiahao and Chen, Ningxin and Chen, Ruijue and Chen, Yanru and Chen, Yuankun and Chen, Yutian and others},
  journal={arXiv preprint arXiv:2507.20534},
  year={2025}
}

@misc{anthropic2025haiku,
  author = {{Anthropic Team}},
  title  = {Claude Haiku 4.5 System Card},
  year   = {2025},
  month  = oct,
  howpublished = {\url{https://assets.anthropic.com/m/99128ddd009bdcb/original/Claude-Haiku-4-5-System-Card.pdf}},
  note   = {System card},
}

@article{kim2025towards,
  title={Towards a science of scaling agent systems},
  author={Kim, Yubin and Gu, Ken and Park, Chanwoo and Park, Chunjong and Schmidgall, Samuel and Heydari, A Ali and Yan, Yao and Zhang, Zhihan and Zhuang, Yuchen and Malhotra, Mark and others},
  journal={arXiv preprint arXiv:2512.08296},
  year={2025}
}

@article{huang2025biomni,
  title={Biomni: A general-purpose biomedical ai agent},
  author={Huang, Kexin and Zhang, Serena and Wang, Hanchen and Qu, Yuanhao and Lu, Yingzhou and Roohani, Yusuf and Li, Ryan and Qiu, Lin and Li, Gavin and Zhang, Junze and others},
  journal={biorxiv},
  year={2025}
}

@article{liu2024autotimes,
  title={Autotimes: Autoregressive time series forecasters via large language models},
  author={Liu, Yong and Qin, Guo and Huang, Xiangdong and Wang, Jianmin and Long, Mingsheng},
  journal={Advances in Neural Information Processing Systems},
  volume={37},
  pages={122154--122184},
  year={2024}
}

@inproceedings{li2024urbangpt,
  title={Urbangpt: Spatio-temporal large language models},
  author={Li, Zhonghang and Xia, Lianghao and Tang, Jiabin and Xu, Yong and Shi, Lei and Xia, Long and Yin, Dawei and Huang, Chao},
  booktitle={Proceedings of the 30th ACM SIGKDD Conference on Knowledge Discovery and Data Mining},
  pages={5351--5362},
  year={2024}
}

@article{lu2024incontext,
  title={In-context time series predictor},
  author={Lu, Jiecheng and Sun, Yan and Yang, Shihao},
  journal={arXiv preprint arXiv:2405.14982},
  year={2024}
}

@inproceedings{lee2025timecap,
  title={Timecap: Learning to contextualize, augment, and predict time series events with large language model agents},
  author={Lee, Geon and Yu, Wenchao and Shin, Kijung and Cheng, Wei and Chen, Haifeng},
  booktitle={Proceedings of the AAAI Conference on Artificial Intelligence},
  volume={39},
  pages={18082--18090},
  year={2025}
}

@inproceedings{zhou2025merit,
  title={MERIT: Multi-Agent Collaboration for Unsupervised Time Series Representation Learning},
  author={Zhou, Shu and Xuan, Yunyang and Ao, Yuxuan and Wang, Xin and Fan, Tao and Wang, Hao},
  booktitle={Findings of the Association for Computational Linguistics: ACL 2025},
  pages={24011--24028},
  year={2025}
}

@misc{liu2025ts,
      title={TS-Agent: Understanding and Reasoning Over Raw Time Series via Iterative Insight Gathering}, 
      author={Penghang Liu and Elizabeth Fons and Annita Vapsi and Mohsen Ghassemi and Svitlana Vyetrenko and Daniel Borrajo and Vamsi K. Potluru and Manuela Veloso},
      year={2026},
      eprint={2510.07432},
      archivePrefix={arXiv},
      primaryClass={cs.AI},
      url={https://arxiv.org/abs/2510.07432}, 
}

@article{zhou2024can,
  title={Can LLMs understand time series anomalies?},
  author={Zhou, Zihao and Yu, Rose},
  journal={arXiv preprint arXiv:2410.05440},
  year={2024}
}

@inproceedings{jin2024position,
  title={Position: What can large language models tell us about time series analysis},
  author={Jin, Ming and Zhang, Yifan and Chen, Wei and Zhang, Kexin and Liang, Yuxuan and Yang, Bin and Wang, Jindong and Pan, Shirui and Wen, Qingsong},
  booktitle={41st International Conference on Machine Learning},
  year={2024},
  organization={MLResearchPress}
}

@article{feli2025llm,
  title={An LLM-Powered Agent for Physiological Data Analysis: A Case Study on PPG-based Heart Rate Estimation},
  author={Feli, Mohammad and Azimi, Iman and Liljeberg, Pasi and Rahmani, Amir M},
  journal={arXiv preprint arXiv:2502.12836},
  year={2025}
}

@article{phan2025hle,
  title={Humanity's last exam},
  author={Phan, Long and Gatti, Alice and Han, Ziwen and Li, Nathaniel and Hu, Josephina and Zhang, Hugh and Zhang, Chen Bo Calvin and Shaaban, Mohamed and Ling, John and Shi, Sean and others},
  journal={arXiv preprint arXiv:2501.14249},
  year={2025}
}

@misc{artificialanalysis2025aalcr,
  author={{Artificial Analysis}},
  title={AA-LCR: Large Context Reasoning Dataset},
  year={2025},
  publisher={Hugging Face},
  howpublished={Hugging Face dataset, \url{https://huggingface.co/datasets/ArtificialAnalysis/AA-LCR}},
  note={Accessed: 2026-01-27}
}

@article{wang2024mmlupro,
  title={Mmlu-pro: A more robust and challenging multi-task language understanding benchmark},
  author={Wang, Yubo and Ma, Xueguang and Zhang, Ge and Ni, Yuansheng and Chandra, Abhranil and Guo, Shiguang and Ren, Weiming and Arulraj, Aaran and He, Xuan and Jiang, Ziyan and others},
  journal={Advances in Neural Information Processing Systems},
  volume={37},
  pages={95266--95290},
  year={2024}
}

@article{rein2024gpqa,
  title={Gpqa: A graduate-level google-proof q\&a benchmark},
  author={Rein, David and Hou, Betty Li and Stickland, Asa Cooper and Petty, Jackson and Pang, Richard Yuanzhe and Dirani, Julien and Michael, Julian and Bowman, Samuel R},
  journal={arXiv preprint arXiv:2311.12022},
  year={2023}
}

@article{jain2024livecodebench,
  title={Livecodebench: Holistic and contamination free evaluation of large language models for code},
  author={Jain, Naman and Han, King and Gu, Alex and Li, Wen-Ding and Yan, Fanjia and Zhang, Tianjun and Wang, Sida and Solar-Lezama, Armando and Sen, Koushik and Stoica, Ion},
  journal={arXiv preprint arXiv:2403.07974},
  year={2024}
}

@article{tian2024scicode,
  title={Scicode: A research coding benchmark curated by scientists},
  author={Tian, Minyang and Gao, Luyu and Zhang, Shizhuo and Chen, Xinan and Fan, Cunwei and Guo, Xuefei and Haas, Roland and Ji, Pan and Krongchon, Kittithat and Li, Yao and others},
  journal={Advances in Neural Information Processing Systems},
  volume={37},
  pages={30624--30650},
  year={2024}
}

@article{pyatkin2025ifbench,
  title={Generalizing Verifiable Instruction Following},
  author={Pyatkin, Valentina and Malik, Saumya and Graf, Victoria and Ivison, Hamish and Huang, Shengyi and Dasigi, Pradeep and Lambert, Nathan and Hajishirzi, Hannaneh},
  journal={arXiv preprint arXiv:2507.02833},
  year={2025}
}

@article{yang2022artificial,
  title={Artificial intelligence-enabled detection and assessment of Parkinson's disease using nocturnal breathing signals},
  author={Yang, Yuzhe and Yuan, Yuan and Zhang, Guo and Wang, Hao and Chen, Ying-Cong and Liu, Yingcheng and Tarolli, Christopher G and Crepeau, Daniel and Bukartyk, Jan and Junna, Mithri R and others},
  journal={Nature Medicine},
  volume={28},
  number={10},
  pages={2207-2215},
  year={2022},
  publisher={Nature Publishing Group}
}

@inproceedings{wangExecutableCodeActions2024,
  title={Executable code actions elicit better llm agents},
  author={Wang, Xingyao and Chen, Yangyi and Yuan, Lifan and Zhang, Yizhe and Li, Yunzhu and Peng, Hao and Ji, Heng},
  booktitle={Forty-first International Conference on Machine Learning},
  year={2024}
}

@article{jeong2024exploring,
  title={Exploring spectrogram-based audio classification for Parkinson’s disease: A study on speech classification and qualitative reliability verification},
  author={Jeong, Seung-Min and Kim, Seunghyun and Lee, Eui Chul and Kim, Han Joon},
  journal={Sensors},
  volume={24},
  number={14},
  pages={4625},
  year={2024},
  publisher={MDPI}
}

@article{burq2023human,
  title={Human Activity Recognition Using Self-Supervised Representations of Wearable Data},
  author={Burq, Maximilien and Sridhar, Niranjan},
  journal={arXiv preprint arXiv:2304.14912},
  year={2023}
}

@article{ayat2024novel,
  title={Novel diabetes classification approach based on CNN-LSTM: enhanced performance and accuracy},
  author={Ayat, Yassine and Benzekri, Wiame and El Moussati, Ali and Mir, Ismail and Benzaouia, Mohammed and El Aouni, Abdelaziz},
  journal={Diagnostyka},
  volume={25},
  number={1},
  year={2024},
  publisher={Polska Akademia Nauk. Polskie Towarzystwo Diagnostyki Technicznej PAN}
}

@article{herrero2022identifying,
  title={Identifying continuous glucose monitoring data using machine learning},
  author={Herrero, Pau and Reddy, Monika and Georgiou, Pantelis and Oliver, Nick S},
  journal={Diabetes Technology \& Therapeutics},
  volume={24},
  number={6},
  pages={403--408},
  year={2022},
  publisher={Mary Ann Liebert, Inc., publishers 140 Huguenot Street, 3rd Floor New~…}
}

@article{pahar2022covid,
  title={COVID-19 detection in cough, breath and speech using deep transfer learning and bottleneck features},
  author={Pahar, Madhurananda and Klopper, Marisa and Warren, Robin and Niesler, Thomas},
  journal={Computers in biology and medicine},
  volume={141},
  pages={105153},
  year={2022},
  publisher={Elsevier}
}

@article{kumar2025cough,
  title={Cough Classification using Few-Shot Learning},
  author={Kumar, Yoga Disha Sendhil and Shetty, Manas V and Vhaduri, Sudip},
  journal={arXiv preprint arXiv:2509.09515},
  year={2025}
}

@article{svaricek2025insight,
  title={INSIGHT: Combining Fixation Visualisations and Residual Neural Networks for Dyslexia Classification From Eye-Tracking Data},
  author={Svaricek, Roman and Dostalova, Nicol and Sedmidubsky, Jan and Cernek, Andrej},
  journal={Dyslexia},
  volume={31},
  number={1},
  pages={e1801},
  year={2025},
  publisher={Wiley Online Library}
}

@article{sadhu2025task,
  title={Task Decoding based on Eye Movements using Synthetic Data Augmentation},
  author={Sadhu, Shanmuka and Baran, Arca and Pandey, Preeti and Kumar, Ayush},
  journal={arXiv preprint arXiv:2509.11547},
  year={2025}
}

@inproceedings{bhattacharya2024imputation,
  title={Imputation Strategies for Longitudinal Behavioral Studies: Predicting Depression Using GLOBEM Datasets},
  author={Bhattacharya, Sohini and Majethia, Rahul and Choube, Akshat and Mishra, Varun},
  booktitle={Companion of the 2024 on ACM International Joint Conference on Pervasive and Ubiquitous Computing},
  pages={736--742},
  year={2024}
}

@article{booth2022toward,
  title={Toward robust stress prediction in the age of wearables: Modeling perceived stress in a longitudinal study with information workers},
  author={Booth, Brandon M and Vrzakova, Hana and Mattingly, Stephen M and Martinez, Gonzalo J and Faust, Louis and D’Mello, Sidney K},
  journal={IEEE Transactions on Affective Computing},
  volume={13},
  number={4},
  pages={2201--2217},
  year={2022},
  publisher={IEEE}
}

@article{cansiz2025hierarchical,
  title={Hierarchical Cascade Deep Learning for EMG-Based Behavioral Biometrics: Gesture and Subject Classification},
  author={Cansiz, Berke and Dudukcu, Hatice Vildan and Taskiran, Murat and Kahraman, Nihan},
  journal={IEEE Access},
  year={2025},
  publisher={IEEE}
}

@article{kim2025optimization,
  title={Optimization of Deep Learning Models for Enhanced Respiratory Signal Estimation Using Wearable Sensors},
  author={Kim, Jiseon and Kim, Jooyong},
  journal={Processes},
  volume={13},
  number={3},
  pages={747},
  year={2025},
  publisher={MDPI}
}

@article{bollampally2024optimizing,
  title={Optimizing Edge Computing for Activity Recognition: A Bidirectional LSTM Approach on the PAMAP2 Dataset},
  author={Bollampally, Anupama and Kavitha, J and Sumanya, P and Rajesh, D and Jaffar, Amar Y and Eid, Wesam N and Albarakati, Hussain M and Aldosari, Fahd M and Alharbi, Ayman A},
  journal={Engineering, Technology \& Applied Science Research},
  volume={14},
  number={6},
  pages={18086--18093},
  year={2024}
}

@article{bowd2009diagnostic,
  title={Diagnostic accuracy of pattern electroretinogram optimized for glaucoma detection},
  author={Bowd, Christopher and Vizzeri, Gianmarco and Tafreshi, Ali and Zangwill, Linda M and Sample, Pamela A and Weinreb, Robert N},
  journal={Ophthalmology},
  volume={116},
  number={3},
  pages={437--443},
  year={2009},
  publisher={Elsevier}
}

@inproceedings{singh2025feel,
  title={FEEL: Quantifying Heterogeneity in Physiological Signals for Generalizable Emotion Recognition},
  author={Singh, Pragya and Gupta, Ankush and Jalan, Somay and Kumar, Mohan and Singh, Pushpendra},
  booktitle={The Thirty-ninth Annual Conference on Neural Information Processing Systems Datasets and Benchmarks Track},
  year={2025}
}

@article{cai2020accurate,
  title={Accurate detection of atrial fibrillation from 12-lead ECG using deep neural network},
  author={Cai, Wenjuan and Chen, Yundai and Guo, Jun and Han, Baoshi and Shi, Yajun and Ji, Lei and Wang, Jinliang and Zhang, Guanglei and Luo, Jianwen},
  journal={Computers in biology and medicine},
  volume={116},
  pages={103378},
  year={2020},
  publisher={Elsevier}
}

@article{li2021neural,
  title={Neural networks applied to 12-lead electrocardiograms predict body mass index, visceral adiposity and concurrent cardiometabolic ill-health},
  author={Li, Xinyang and Patel, Kiran Haresh Kumar and Sun, Lin and Peters, Nicholas S and Ng, Fu Siong},
  journal={Cardiovascular Digital Health Journal},
  volume={2},
  number={6},
  pages={S1--S10},
  year={2021},
  publisher={Elsevier}
}

@article{thapa2026multimodal,
  title={A multimodal sleep foundation model for disease prediction},
  author={Thapa, Rahul and Kjaer, Magnus Ruud and He, Bryan and Covert, Ian and Moore IV, Hyatt and Hanif, Umaer and Ganjoo, Gauri and Westover, M Brandon and Jennum, Poul and Brink-Kjaer, Andreas and others},
  journal={Nature Medicine},
  pages={1--11},
  year={2026},
  publisher={Nature Publishing Group US New York}
}

@article{yetton2016automatic,
  title={Automatic detection of rapid eye movements (REMs): A machine learning approach},
  author={Yetton, Benjamin D and Niknazar, Mohammad and Duggan, Katherine A and McDevitt, Elizabeth A and Whitehurst, Lauren N and Sattari, Negin and Mednick, Sara C},
  journal={Journal of neuroscience methods},
  volume={259},
  pages={72--82},
  year={2016},
  publisher={Elsevier}
}

@article{ccay2024eeg,
  title={EEG Based Cigarette Addiction Detection with Deep Learning},
  author={{\c{C}}ay, Talip and {\"O}lmez, Emre and Tan{\i}k, Nermin and Alt{\i}n, Cemil},
  journal={Traitement du Signal},
  volume={41},
  number={3},
  year={2024}
}

@inproceedings{gunnarsdottir2018novel,
  title={A novel sleep stage scoring system: combining expert-based rules with a decision tree classifier},
  author={Gunnarsdottir, Kristin M and Gamaldo, Charlene E and Salas, Rachel ME and Ewen, Joshua B and Allen, Richard P and Sarma, Sridevi V},
  booktitle={2018 40th annual international conference of the IEEE engineering in medicine and biology society (EMBC)},
  pages={3240--3243},
  year={2018},
  organization={IEEE}
}

@article{oh2023prediction,
  title={Prediction model for myocardial injury after non-cardiac surgery using machine learning},
  author={Oh, Ah Ran and Park, Jungchan and Shin, Seo Jeong and Choi, Byungjin and Lee, Jong-Hwan and Lee, Seung-Hwa and Yang, Kwangmo},
  journal={Scientific reports},
  volume={13},
  number={1},
  pages={1475},
  year={2023},
  publisher={Nature Publishing Group UK London}
}

@article{shim2025machine,
  title={Machine Learning Methods for the Prediction of Intraoperative Hypotension with Biosignal Waveforms},
  author={Shim, Jae-Geum and Yoon, Wonhyuck and Lee, Sang Jun and Chang, Se-Hyun and Jung, So-Ra and Chung, Jun Young},
  journal={Medicina},
  volume={61},
  number={11},
  pages={2039},
  year={2025},
  publisher={MDPI}
}

@article{langer2025opentslm,
  title={Opentslm: Time-series language models for reasoning over multivariate medical text-and time-series data},
  author={Langer, Patrick and Kaar, Thomas and Rosenblattl, Max and Xu, Maxwell A and Chow, Winnie and Maritsch, Martin and Verma, Aradhana and Han, Brian and Kim, Daniel Seung and Chubb, Henry and others},
  journal={arXiv preprint arXiv:2510.02410},
  year={2025}
}
